\documentclass[english]{MyArticle}
\makeatletter

\providecommand{\tabularnewline}{\\}
\usepackage{tablefootnote,soul}
\usepackage{adjustbox}
\usepackage{soul}
\usepackage{listings}
\usepackage{babel,xcolor}
\usepackage{algorithm}
\usepackage{courier}
\usepackage{lineno}

\usepackage{algpseudocode}

\makeatother

\begin{document}
%\linenumbers

\title{A Unifying Perspective on Non-Stationary Kernels for Deeper Gaussian Processes}
\author[1,*]{Marcus M. Noack}
\author[1,2]{Hengrui Luo}
\author[3]{Mark D. Risser}
\affil[1]{Applied Mathematics and Computational Research Division, Lawrence Berkeley National Laboratory, Berkeley, CA 94720}
\affil[2]{Department of Statistics, Rice University, TX 77005}
\affil[3]{Climate and Ecosystem Sciences Division, Lawrence Berkeley National Laboratory, Berkeley, CA 94720}
\affil[*]{MarcusNoack@lbl.gov}

\maketitle
\begin{abstract}
The Gaussian process (GP) is a popular statistical technique for stochastic
function approximation and uncertainty quantification from data. GPs
have been adopted into the realm of machine learning in the last two
decades because of their superior prediction abilities, especially
in data-sparse scenarios, and their inherent ability to provide robust uncertainty
estimates. Even so, their performance highly depends on intricate customizations of the core methodology, 
which often leads to dissatisfaction among practitioners when standard setups and off-the-shelf software tools are being deployed. 
Arguably the most important building block of a GP is the
kernel function which assumes the role of a covariance operator. Stationary kernels of the Mat\'ern class
are used in the vast majority of applied studies; poor prediction performance and unrealistic uncertainty quantification
are often the consequences. Non-stationary kernels show improved performance
but are rarely used due to their more complicated functional form and the associated
effort and expertise needed to define and tune them optimally. In this perspective,
we want to help ML practitioners make sense of some of the
most common forms of non-stationarity for Gaussian processes. We show
a variety of kernels in action using representative datasets, 
carefully study their properties, and compare their performances. 
Based on our findings, we propose a new
kernel that combines some of the identified advantages of existing kernels.
\end{abstract}

%%%%%%%%%%%%%%%%%%%%%%%%%%%%%%
%%%%%%%%%%%%%%%%%%%%%%%%%%%%%%
%%%%%%%%%%%%%%%%%%%%%%%%%%%%%%
%%%%%%%%%%%%%%%%%%%%%%%%%%%%%%
\section{Introduction}
%%%%%%%%%%%%%%%%%%%%%%%%%%%%%%
%%%%%%%%%%%%%%%%%%%%%%%%%%%%%%
%%%%%%%%%%%%%%%%%%%%%%%%%%%%%%
%%%%%%%%%%%%%%%%%%%%%%%%%%%%%%
The Gaussian process (GP) is arguably the most popular member of the large family of stochastic processes and provides a powerful and flexible framework for stochastic function approximation in the form of Gaussian process regression (GPR). 
A GP is defined as a collection of random variables, such that any 
finite subset has a joint normal (Gaussian) distribution \cite{williams2006gaussian}. In the scope of GPR, the random variables are considered (model or latent) function values $\{f(\mathbf{x}_1),f(\mathbf{x}_2),f(\mathbf{x}_3),...\}$. 
Equivalently, we can understand a GP as a description of distributions 
over function spaces, in which case $\mathcal{GP}(m(\mathbf{x}),\mathbf{K})$ is 
entirely specified by a mean function 
$m(\mathbf{x})=\mathbb{E}[f(\mathbf{x})]$ 
and a covariance $\mathbf{K}=k(\mathbf{x}_j,\mathbf{x}_i;h)=\mathbb{E}[(f(\mathbf{x}_i)-m(\mathbf{x}_i))(f(\mathbf{x}_j)-m(\mathbf{x}_j))]$,
%A GP is characterized as a Gaussian probability distribution over (model or latent) function values $\{f(\mathbf{x}_1),f(\mathbf{x}_2),f(\mathbf{x}_3),...\}$ and therefore over the subspace $\{f:f(\mathbf{x})=\sum_i^m~\alpha_i k(\mathbf{x},\mathbf{x}_i;h)~\forall ~ \mathbf{x}\in \mathcal{X},~m \in \mathbb{N},~\alpha_i \in \mathbb{R}\}$, called a reproducing kernel Hilbert space (RKHS)
where $k(\mathbf{x},\mathbf{x}_i;h)$ is the kernel or covariance function, $h \in \Theta$ is a vector of hyperparameters, and $\mathcal{X}$ is the input space.
This role of the kernel as a covariance operator makes it arguably the most important building block when it comes to optimizing the flexibility,  prediction accuracy, and expressiveness of a GP.
%The RKHS --- as the name would suggest --- is directly influenced by the choice of the kernel function, which assumes the role of the covariance function in the GP framework. This double role makes the kernel an important building block when it comes to optimizing the flexibility, accuracy, and expressiveness of a GP. 

\vspace{2mm}
\noindent
%stat
In a recent review \cite{pilario2020review}, it was pointed out that the vast majority of applied studies using GPs employ the radial basis function (RBF) kernel (also referred to as the squared exponential or Gaussian kernel). The fraction is even higher when we include other stationary kernels. Stationary kernels are characterized by $k(\mathbf{x}_i,\mathbf{x}_j)=k(|\mathbf{x}_i-\mathbf{x}_j|)$, i.e., covariance matrix entries only depend on some distance between data points in the input domain, not on their respective location. Stationary kernels are popular because they carry little risk --- in terms of model misspecification --- and come with only a few hyperparameters that are easy to optimize or train. However, it has been shown that the stationarity assumption can lead to poor prediction performance and unrealistic uncertainty quantification that is affected mostly by data-point geometry \cite{noack2023mathematical}. In other words, uncertainty will increase when moving away from data points at a constant rate across the input space. To overcome these limitations, significant research attention has been drawn to non-stationary Gaussian process regression \citep{sampson1992nonparametric,paciorek2003nonstationary,paciorek2006spatial}. Non-stationary kernels depend on the respective locations in the input space explicitly, i.e., $k(\mathbf{x}_i,\mathbf{x}_j) \neq k(|\mathbf{x}_i-\mathbf{x}_j|)$. This makes them significantly more flexible and expressive leading to higher-quality estimates of uncertainty across the domain. Even so, non-stationary kernels are rarely used in applied machine learning (ML) due to the inherent difficulty of customization, optimizing hyperparameters, and the associated risks of model misspecification (wrong hyperparameters), overfitting, and computational costs due to the need to find many hyperparameters \citep{snoek2012practical, noack2023exact}. When applied correctly, non-stationary GPs have been shown to provide significant advantages over their stationary counterparts, especially in scenarios where the data exhibit non-stationary behavior \citep{schabenberger2017statistical}.

\vspace{2mm}
\noindent
This paper aims to bring some structure to the use of non-stationary kernels and related techniques to make it more feasible for the practitioner to use these kernels effectively. Throughout this paper --- in the hope of covering the most practical set of available options --- we focus on and compare four ways to handle non-stationarity in a dataset:
\begin{enumerate}
    \item Ignore it: Most datasets and underlying processes exhibit some level of non-stationarity which is often simply ignored. This leads to the use of stationary kernels of the form $k_{stat} = k(|\mathbf{x}_i-\mathbf{x}_j|)$; $|\cdot|$ is a norm appropriate to the properties of the input space. Given the key takeaways in \cite{pilario2020review}, this option is chosen by many which serves as the main motivation for the authors to write this perspective.
    \item Parametric non-stationarity: Kernels of the form $k(\mathbf{x}_i,\mathbf{x}_j)=\sum_{d=1}^N g_d(\mathbf{x}_i) g_d(\mathbf{x}_j)  k_{stat}(|\mathbf{x}_i-\mathbf{x}_j|)$, where $g_d(\mathbf{x})$ can be any parametric function over the input space and $N$ is some positive integer.
    \item Deep kernels: Kernels of the form $k_{stat}(|\boldsymbol{\phi}(\mathbf{x}_i)-\boldsymbol{\phi}(\mathbf{x}_j)|)$, where $\boldsymbol{\phi}$ is a neural network (NN), and $k_{stat}$ is any stationary kernel. This kernel was introduced by Wilson et al. \cite{wilson2016deep} and was quickly established as a powerful technique, even though potential pitfalls related to overfitting were also discovered \cite{ober2021promises}.
    \item Deep GPs: Achieving non-stationarity not by using a non-stationary kernel, but by stacking stationary GPs --- meaning the output of one GP becomes the input to the next, similar to how neural networks are made up of stacked layers --- which introduces a non-linear transformation of the input space and thereby non-stationarity \cite{damianou2013deep}. 
\end{enumerate}
Non-stationarity can also be modeled through the prior-mean function of a Gaussian process but, for the purpose of this paper, we use a constant prior mean and focus on the four methods described above. From the brief explanations of the three types of non-stationarity, it is immediately obvious that deriving, implementing, and deploying a non-stationary kernel and deep GPs (DGPs) can be an intimidating endeavor. The function $g_d$ (in \#2 above), for instance, can be chosen to be any sum of local basis functions --- linear, quadratic, cubic, piecewise constant, etc. --- any polynomial or sums thereof, and splines, just to name a few. Given the vast variety of neural networks (in \#3), choosing $\boldsymbol{\phi}$ is no easier. DGPs involve stacking GPs which increases flexibility but also leads to a very large number of hyperparameters, or latent variables, making optimization and inference a big challenge. In addition, a more complex kernel generally results in reduced interpretability --- a highly valued property of GPs. 

\vspace{2mm}
\noindent
This paper is concerned with shining a light on the characteristics and performance of different kernels and experiencing them in action. In a very pragmatic approach, we study the performance of the three methodologies above to tackle non-stationarity and compare them to a stationary Mat\'ern kernel to create a baseline. Our contributions can be summarized as follows: 
\begin{enumerate}
    \item We strategically choose a representative set of test kernels and investigate their properties.
    \item We choose test datasets to study the performance of the chosen kernels.
    \item We derive performance and (non)-stationarity metrics that allow fair comparisons of the non-stationary kernels and the selected deep GP implementation.
    \item We compare the performance of the candidate kernels tasked with modeling the test datasets and present the uncensored results.
    \item We use the gathered insights to draw some conclusions and suggest a new kernel design for further investigation.
\end{enumerate}
We identify and exploit two primary pathways to achieve non-stationary properties: through the application of deep architectures --- deep kernels or deep GPs --- and the utilization of parametric non-stationary kernels. Our work delves into the role of parametric non-stationary kernels arguing that these kernels offer a similar behavior to a deep-kernel architecture without any direct notion of deepness, hinting at the possibility that deepness and flexibility are practically synonymous. This provides a valuable computational perspective to the broader dialogue surrounding deep architectures and non-stationarity, offering insights that may be instrumental in future research.

\vspace{2mm}
\noindent
The remainder of this paper is organized as follows. The next section introduces some minimal but necessary theory. There, we also introduce some measures that will help us later when it comes to investigating the performance of different kernels. Then we give an overview of the relevant literature and some background. Subsequently, we introduce a range of different kernels and some test datasets, which positions the reader to see each kernel in action in different situations. We then discuss the key takeaways; we give some pointers toward improved kernel designs, make remarks about the connection between non-stationary kernels and multi-task learning, and conclude with a summary of the findings.
\begin{figure}
    \centering
    \includegraphics[width = \linewidth]{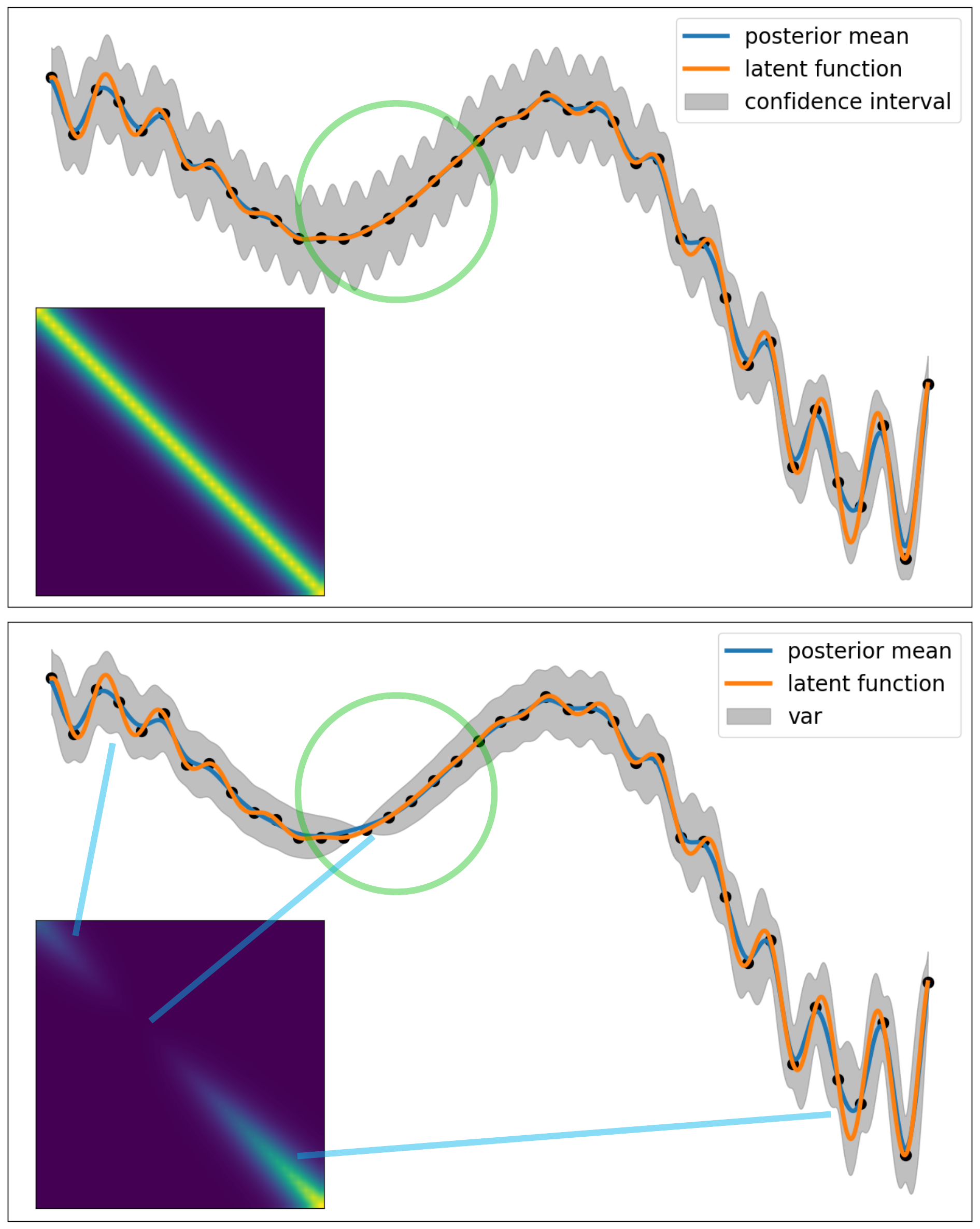}
    \put(-420,520){a)}
    \put(-420,250){b)}
    \caption{The key concept and essence of non-stationary kernels. A synthetic function --- that is also later used for our computational experiments --- was sampled at 40 equidistant points. The function is comprised of high-frequency regions (far left and right), and near-constant-gradient regions (center, green circle). A Gaussian process (GP) is tasked with interpolating the data using a stationary (top, a) and a non-stationary (bottom, b) kernel. For each case, the function approximation and the prior covariance matrix are presented. While the posterior mean is similar in both cases, the posterior variance differs substantially. Focusing on the central region, the uncertainty increases between data points, even though the function is very well-behaved there. The covariance matrix can deliver clues as to why this might happen. The matrix is constant along diagonals, which translates into uncertainties that depend on the distance from surrounding data points only, independent of where in the domain they are located. The non-stationary kernel has no such restriction and provides more realistic estimates of the uncertainty. The covariance entries are not constant along diagonals but correspond to different regions of the function (blue line connections).}
    \label{fig:intro}
\end{figure}

\section{Preliminaries} \label{sec:prelim}
%%%%%%%%%%%%%%%%%%%%%%%%%%%%%%
%%%%%%%%%%%%%%%%%%%%%%%%%%%%%%
%%%%%%%%%%%%%%%%%%%%%%%%%%%%%%
%%%%%%%%%%%%%%%%%%%%%%%%%%%%%%
The purpose of this section is to give the reader the necessary tools to follow along with our computational experiments. It is not intended to be a complete or comprehensive overview of the GP theory or related methodologies. 
For the remainder of this paper, we consider some input space $\mathcal{X} \subset \mathbb{R}^{n}$ with elements $\mathbf{x}$. While we often think of the input space as a subspace of the Euclidean space, this is not a necessary assumption of the framework in general. We assume data $\mathcal{D}$ as a set of tuples $\{\mathbf{x}_i,\mathbf{y}_i\}~\forall~i=\{1,2,3,...,|\mathcal{D}|\}$. In this work, except for a remark on multivariate GPs at the end, we will assume scalar $y_i$. 

\subsection{GPs in a Nutshell}
The GP is a versatile statistical learning tool that can be applied to any black-box function. The ability of GPs to estimate both the mean and the variance of the latent function makes it an ideal ML tool for quantifying uncertainties in the function approximation. GPs assume a model of the form $y(\mathbf{x})~=~f(\mathbf{x})+\epsilon(\mathbf{x})$, where $f(\mathbf{x})$ is the unknown latent function, $y(\mathbf{x})$ is the noisy function evaluation (the measurement), $\epsilon(\mathbf{x})$ is the noise term, and $\mathbf{x}$ is an element of the input space $\mathcal{X}$. For a fixed, finite set of inputs $\mathcal{D} = \{\mathbf{x}_1, \dots, \mathbf{x}_{|\mathcal{D}|}\}$, a Gaussian process implies that the vector $\mathbf{f}=(f(\mathbf{x}_1), \dots, f(\mathbf{x}_{|\mathcal{D}|}) )$ has a multivariate Gaussian distribution
\begin{equation}
    p(\mathbf{f})=\frac{1}{\sqrt{(2\pi)^{|\mathcal{D}|}|\mathbf{K}|}}
    \exp \left[ -\frac{1}{2}(\mathbf{f}-\mathbf{m})^T \mathbf{K}^{-1}(\mathbf{f}-\mathbf{m}) \right]
    \label{eq:priorGP}
\end{equation}
(here taken to be the prior distribution over function values), where $\mathbf{K}$ is the covariance matrix defined by the kernel function $K_{ij}~=~k(\mathbf{x}_i,\mathbf{x}_j)$, and $\mathbf{m}$ is the prior mean 
obtained by evaluating the prior-mean function at points $\mathbf{x}_i$. Conditioning on $\mathbf{f}$, we can define a likelihood for the corresponding vector of data, $\mathbf{y}=(y(\mathbf{x}_1), \dots, y(\mathbf{x}_{|\mathcal{D}|}) )$, denoted $p(\mathbf{y}|\mathbf{f})$, which allows us to perform Bayesian inference, typically (but not necessarily) using a multivariate Gaussian distribution
\begin{equation} \label{eq:likelihood}
    p(\mathbf{y}|\mathbf{f}) = \frac{1}{\sqrt{(2\pi)^{|\mathcal{D}|}|\mathbf{V}|}}
    \exp \left[ -\frac{1}{2}(\mathbf{y}-\mathbf{f})^T \mathbf{V}^{-1}(\mathbf{y}-\mathbf{f}) \right],
\end{equation}
where the covariance matrix $\mathbf{V}$ is used to describe measurement errors arising from $\epsilon(\mathbf{x})$. Training the hyperparameters, extending the prior over latent function values at points of interest, marginalizing over $\mathbf{f}$, and conditioning on the data $\mathbf{y}$ yield the posterior probability density function (PDF) for the requested points \cite{williams2006gaussian}. For example, for a Gaussian process with Gaussian likelihood, marginalization over $\mathbf{f}$ can be obtained in closed form
\begin{equation} \label{eq:margLik}
    p(\mathbf{y}|h) = \int_\mathcal{\mathbb{R}^{|\mathcal{D}|}} p(\mathbf{y}|\mathbf{f}) p(\mathbf{f}) d\mathbf{f} = \frac{1}{\sqrt{(2\pi)^{|\mathcal{D}|}|\mathbf{K} + \mathbf{V}|}}
    \exp \left[ -\frac{1}{2}(\mathbf{y}-\mathbf{m})^T (\mathbf{K} + \mathbf{V})^{-1}(\mathbf{y}-\mathbf{m}) \right],
\end{equation}
where we suppress the implicit conditioning on the hyperparameters. Training the GP can be done by maximizing the log marginal likelihood --- i.e., the log of Equation~\eqref{eq:margLik} taken as a function of the hyperparameters $h$ when the data $\mathbf{y}$ is given ---
\begin{equation}
    \ln(p(\mathbf{y}|h)) \propto -\frac{1}{2} (\mathbf{y}-\mathbf{m}(h))^T (\mathbf{K}(h)+\mathbf{V}(h))^{-1} (\mathbf{y}-\mathbf{m}(h)) - \frac{1}{2} \ln(|\mathbf{K}(h)+\mathbf{V}(h)|) 
    \label{eqn:log_likelihood}
\end{equation}
with respect to the hyperparameters $h$. 
After the hyperparameters are found, the posterior is defined as
\begin{align}\label{eq:pred_distr}
    p(\mathbf{f}_0|\mathbf{y},h)~&=~\int_{\mathbb{R}^{|\mathcal{D}|}} p(\mathbf{f}_0|\mathbf{f},\mathbf{y},h)~p(\mathbf{f}|\mathbf{y},h)~d\mathbf{f} \nonumber \\
    &~\propto \mathcal{N}(\mathbf{m}_0 +\pmb{\kappa}^T~
 (\mathbf{K}+\mathbf{V})^{-1}~(\mathbf{y}-\mathbf{m} ), \pmb{\mathcal{K}} -
    \pmb{\kappa}^T~(\mathbf{K}+\mathbf{V})^{-1}~\pmb{\kappa}),
\end{align}
where $\pmb{\kappa}=k(\mathbf{x}_0,\mathbf{x}_j)$, $\pmb{\mathcal{K}}=k(\mathbf{x}_0,\mathbf{x}_0)$, and $\mathbf{m}_0$ is the prior mean function evaluated at the prediction points $\mathbf{x}_0$. If, in a Bayesian setting, we want to further incorporate uncertainties in the hyperparameters $h \in \Theta$, Equation~\eqref{eq:pred_distr} becomes
\begin{align}\label{eq:pred_distr_bayes}
    p(\mathbf{f}_0|\mathbf{y})~&=~ \int_{\Theta} p(\mathbf{f}_0| \mathbf{y},h)~p(h|\mathbf{y})~dh,
\end{align}
where $p(\mathbf{f}_0| \mathbf{y},h)$ is available in closed form \eqref{eq:pred_distr}.
This basic framework can be extended by more flexible mean, noise, and kernel functions \cite{picheny2013nonstationary,pilario2020review, noack2022advanced, noack2022advanced, noack2021gaussian}.
Much of the GP framework depends on the ability to extend the original prior $p(\mathbf{f})$ to the joint distribution $p(\mathbf{f},\mathbf{f}_0)$, which is made possible through the definition of the kernel as a covariance operator that can be evaluated on pairs of arbitrary inputs $(\mathbf{x}_i,\mathbf{x}_j)$. 

\subsection{The Covariance Operator of a GP: The Kernel or Covariance Function}
%Marcus
In this work, the focus is on the kernel function --- also called covariance function --- of a GP, denoted $k: \mathcal{X} \times \mathcal{X} \rightarrow \mathbb{R}$. In the GP framework, the kernel function assumes the role of a covariance operator, i.e., the elements of the covariance matrix $\mathbf{K}$ in Equation~\eqref{eq:priorGP} are defined by  ${K}_{ij}=k(\mathbf{x}_i,\mathbf{x}_j)$. Because of that, kernel functions have to be symmetric and positive semi-definite  (psd); a complete characterization of the class of valid kernel functions is given by Bochner's theorem \citep{bochner1959lectures,adler1981geometry}. 
%At the same time, the kernel uniquely defines the underlying reproducing kernel Hilbert space (RKHS).
%, which is the function or hypothesis space of the Gaussian process posterior mean.

\vspace{2mm}
\noindent
Kernels are called ``stationary'' if the function only depends on the distance between the two inputs, not their respective locations, i.e., $k(\mathbf{x}_i,\mathbf{x}_j)=k(|\mathbf{x}_i-\mathbf{x}_j|)$. The arguably most prominent members are kernels of the Mat\'ern class (see for instance \cite{stein1999interpolation}), which includes the exponential
\begin{equation} \label{eq:expK}
    k(\mathbf{x}_i,\mathbf{x}_j) = \sigma_s e^{-0.5 \frac{||\mathbf{x}_i - \mathbf{x}_j||_2}{l}}
\end{equation}
and the squared exponential (RBF) kernel
\begin{equation} \label{eq:rbfK}
    k(\mathbf{x}_i,\mathbf{x}_j) = \sigma_s e^{-0.5 \frac{||\mathbf{x}_i - \mathbf{x}_j||_2^2}{l^2}},
\end{equation}
the by-far most used kernel by practitioners \cite{pilario2020review}; here, $||\cdot||_2$ is the Euclidean norm, which makes the kernel function both stationary and isotropic. These two special cases of the Mat\'ern kernel class have two hyperparameters: $\sigma_s$, the signal variance, and $l$, the length scale, both of which are constant scalars applied to the entire domain $\mathcal{X}$.
Stationary and isotropic kernels can be advanced by allowing anisotropy in the norm, i.e., $((\mathbf{x}_i - \mathbf{x}_j)^T \mathbf{M}(\mathbf{x}_i - \mathbf{x}_j))^{1/2}$
where $\mathbf{M}$ is some symmetric positive definite matrix whose entries are typically included in the vector of hyperparameters that need to be found. Incorporating anisotropy allows the kernel function to stretch distances such that the implied covariances have ellipsoidal patterns (as opposed to spherical patterns for isotropic kernels). When $\mathbf{M} = \frac{1}{l} \mathbf{I}_n$ the isotropic kernels in Equations~\eqref{eq:expK} and \eqref{eq:rbfK} are recovered; more generally, when $\mathbf{M}$ is diagonal with different elements along the diagonal, automatic relevance detection is recovered \cite{wipf2007new}.
Other stationary kernel designs include the spectral kernel, the periodic kernel, and the rational quadratic kernel. 

\vspace{2mm}
\noindent
Non-stationary kernels do not have the restriction of only depending on the distance between the input points, but depend on their explicit positions $\mathbf{x}_i$ and $\mathbf{x}_j$, i.e., $k(\mathbf{x}_i,\mathbf{x}_j) \neq k(|\mathbf{x}_i-\mathbf{x}_j|)$. Formulating new non-stationary kernels comes with the difficulty of proving positive semi-definiteness which is often challenging. However, the statistics and machine learning literature has a variety of general approaches for applying valid non-stationary kernels; this is the primary topic of this paper which is discussed in Section~\ref{sec:nonstatkern}. 
For now, it is important to keep in mind that the essence of stationary versus non-stationary kernels --- due to the way they deal with the locations in the input space --- manifests itself in the covariance matrix which, in the stationary case, has to be constant along all bands parallel to the diagonal for sorted and equidistant input points, a property that is not observed in the non-stationary case (see Figure \ref{fig:intro}). Therefore, non-stationary kernels lead to more descriptive, expressive, and flexible encodings of covariances and therefore uncertainties. Of course, in addition, the function space contains a much broader class of functions as well when non-stationary kernels are used.

\subsection{A Note on Scalability}
Computational complexity is a significant challenge for GPs, for both
stationary and non-stationary kernels. The need to calculate the log-determinant and invert 
the covariance matrix --- or solve a linear system instead --- results in computational complexity of $\mathcal{O}(|\mathcal{D}|^{3})$,
where $|\mathcal{D}|$ is the number of data points \citep{williams2006gaussian,luo2022sparse}.
This complexity limits the application of GPs for large-scale datasets.
Several methods have been proposed to overcome this issue. Sparse
methods \citep{snelson2005sparse,luo2022sparse} and scalable GPR approaches
\citep{hensman2015scalable} have been developed for stationary GPs.
For non-stationary GPs, methods such as local GPs \citep{gramacy2015local}
and the Bayesian treed GP \citep{gramacy2008bayesian} have been proposed
to tackle this issue. These methods have an origin in divide-and-conquer
methods attempting to break down the problem into smaller, manageable
pieces that can be solved independently \citep{hrluo_2022e}, thereby
reducing the computational complexity for each piece. However, these
approaches can lead to a loss of global information, so finding the
right balance between computational efficiency and model accuracy
remains a key challenge. A recent approach can scale GPs to millions 
of data points without using approximations by allowing a compactly supported, non-stationary 
kernel to discover naturally occurring sparsity \cite{noack2023exact}.

%%%%%%%%%%%%%%%%%%%%%%%%%%%%%%
%%%%%%%%%%%%%%%%%%%%%%%%%%%%%%
%%%%%%%%%%%%%%%%%%%%%%%%%%%%%%
%%%%%%%%%%%%%%%%%%%%%%%%%%%%%%
\section{Non-Stationary Kernels} \label{sec:nonstatkern}
%%%%%%%%%%%%%%%%%%%%%%%%%%%%%%
%%%%%%%%%%%%%%%%%%%%%%%%%%%%%%
%%%%%%%%%%%%%%%%%%%%%%%%%%%%%%
%%%%%%%%%%%%%%%%%%%%%%%%%%%%%%
Stationary kernels are widely used primarily because they are easy and convenient to implement, even though the implied assumption of translation-invariant covariances is almost never exactly true for real-world data sets. As mentioned in Section~\ref{sec:prelim}, there are serious challenges associated with both deriving non-stationary kernels and choosing an appropriate and practical non-stationary kernel from the valid options for any given implementation of a Gaussian process. We now provide a brief overview of the literature on non-stationary kernels, including both a historical perspective of early developments followed by greater detail on three modern frameworks for non-stationary kernels as well as metrics for quantifying non-stationarity in data sets.

\subsection{Historical Perspective}

It has now been over three decades since the first paper on non-stationary kernels via ``deformations'' or warping of the input space appeared \citep{sampson1992nonparametric}. Since then, the statistics literature has developed a number of approaches for non-stationary kernels, mostly in the context of modeling spatially-referenced data. These methods can broadly be categorized as basis function expansions and kernel convolutions, in addition to the aforementioned deformation approach. We now briefly summarize each method, focusing on aspects that apply directly to kernel functions for Gaussian processes.

\vspace{2mm}
\noindent
The fundamental idea underpinning the deformation or warping approach \citep{sampson1992nonparametric} is that instead of deriving new classes of non-stationary kernels one can keep isotropic kernels but obtain non-stationarity implicitly by rescaling interpoint distances in a systematic way over the input space. In other words, this approach transforms $\mathcal{X}$ to a new domain, say $\mathcal{X}^*$, wherein stationarity holds. The transformation, say $\boldsymbol{\phi}: \mathbb{R}^n \rightarrow \mathbb{R}^{n^*}$, is a (possibly nonlinear) mapping applied to elements of $\mathcal{X}$ to yield a non-stationary kernel via
\begin{equation}
    k(\mathbf{x}_i,\mathbf{x}_j)=k_{stat}(||\boldsymbol{\phi}(\mathbf{x}_i) - \boldsymbol{\phi}(\mathbf{x}_j)||),
\end{equation}
where $k_{stat}$ is an arbitrary stationary kernel function. Two extensions were later proposed to this approach \citep{damian2001bayesian,schmidt2003bayesian} that supposed the mapping $\boldsymbol{\phi}(\cdot)$ was itself a stochastic process. For example, \cite{schmidt2003bayesian} placed a Gaussian process prior on $\boldsymbol{\phi}(\cdot)$ --- essentially coming up with the idea of deep kernels more than a decade before related ideas appeared in the machine learning literature. In some cases $n^*>n$, i.e., the mapping involves dimension expansion \cite{bornn2012modeling}. Ultimately, early approaches to warping the input space were largely unused due to a lack of computational tools for optimizing the mapping function $\boldsymbol{\phi}(\cdot)$ in a reliable and robust manner.

\vspace{2mm}
\noindent
In contrast to deformations, basis function expansion methods provide constructive approaches for developing non-stationary kernel functions. The main idea for this approach arises from the Karhunen-Lo\`{e}ve Expansion \citep{karhunen1946spektraltheorie,loeve1955probability} of a (mean-zero) stochastic process in terms of orthogonal eigenfunctions $E_m(\cdot)$ and weights $w_m$:
\begin{equation} \label{KarLoev}
f(\mathbf{x}) = \sum_{m=1}^\infty w_m \hskip0.35ex E_m(\mathbf{x}).
\end{equation}
This framework defines a Gaussian process if the weights have a Gaussian distribution; the implied kernel function is
\[
k(\mathbf{x}_i,\mathbf{x}_j) = \sum_{m=1}^\infty {v_m} E_m(\mathbf{x}_i) E_m(\mathbf{x}_j),
\]
where the eigenfunctions and weight variances $v_m$ come from the Fredholm integral equation 
\begin{equation} \label{fredholm}
\int_\mathcal{X} k(\mathbf{x}_i,\mathbf{x}_j) E_m(\mathbf{x}_i) d\mathbf{x}_i =  v_m E_m(\mathbf{x}_j).
\end{equation}
If the infinite series in Equation~\eqref{KarLoev} is truncated to the leading $M$ terms, the finite sum approximation to the kernel can be used instead and is optimal in the sense that it minimizes the variance of the truncation error for all sets of $M$ basis functions when the $E_m(\cdot)$ are the exact solutions to the Fredholm equation \citep{wikleChapter}. The main task is then to model the weight-eigenfunction pairs  $\{ w_m, E_m(\cdot) \}$, which can be done empirically using singular value decomposition \citep{holland} or parametrically using, e.g., wavelets \citep{nychka2002}.

\vspace{2mm}
\noindent
Like basis function expansions, the kernel convolution approach is useful in that it provides a constructive approach to specifying both stochastic models and their covariance functions. The main result is that a stochastic process can be defined by the kernel convolution
\begin{equation} \label{kernelconvolution}
f(\mathbf{x}) = \int_{\mathcal{X}} \kappa_\mathbf{x}(\mathbf{u}) dW({\bf u})
\end{equation}
\citep{thiebaux76,thiebaux_pedder}, where $W(\cdot)$ is a $n$-dimensional stochastic process and $\kappa_\mathbf{x}(\cdot)$ is a kernel function that depends on input location $\mathbf{x}$. \cite{higdon2} summarizes the extremely flexible class of stochastic models defined using Equation~\eqref{kernelconvolution}: see, for example, \cite{Barry1996},  \cite{Wolpert1999}, and \cite{VerHoef2004}. The popularity of this approach is due largely to the fact that it is much easier to specify (possibly parametric) kernel functions than a covariance function directly since the kernel functions only require $\int_{\mathbb{R}^d} \kappa_\mathbf{x}({\bf u}) d{\bf u} <\infty$ and $\int_{\mathbb{R}^d} \kappa^2_\mathbf{x}({\bf u}) d{\bf u} <\infty$. The process $f(\cdot)$ in Equation~\eqref{kernelconvolution} is a Gaussian process when $W(\cdot)$ is chosen to be Gaussian, and the associated covariance function is
\begin{equation} \label{eq:cov_kc}
k(\mathbf{x}_i,\mathbf{x}_j) = \int_{\mathcal{X}}  \kappa_{\mathbf{x}_i}(\mathbf{u}) \kappa_{\mathbf{x}_j}(\mathbf{u}) d{\bf u},    
\end{equation}
which cannot be written in terms of $||\mathbf{x}_i-\mathbf{x}_j||$ and is hence non-stationary. Various choices can be made for using this general framework in practice: replace the integral in Equation~\eqref{kernelconvolution} with a discrete sum approximation \citep{Higdon98} or choose specific $\kappa_\mathbf{x}(\cdot)$ such that the integral in Equation~\eqref{eq:cov_kc} can be evaluated in closed form \citep{Higdon99}. The latter choice can be generalized to yield a closed-form kernel function that allows all aspects of the resulting covariances to be input-dependent: the length-scale \citep{paciorek2003nonstationary,paciorek2006spatial}, the signal variance \citep{risser2015regression}, and even the differentiability of realizations from the resulting stochastic process when the Mat\'ern kernel is leveraged \citep{stein2005}. This approach is often referred to as ``parametric'' non-stationarity since a non-stationary kernel function is obtained by allowing its hyperparameters to depend on input location. In practice, some care needs to be taken to ensure that the kernel function is not too flexible and can be accurately optimized \citep{paciorek2006spatial,anderes2008}. We return to a version of this approach in the next section.

\vspace{2mm}
\noindent
In conclusion of this section, the statistics literature contains a broad set of techniques (only some of which are summarized here) for developing non-stationary kernel functions. However, historically speaking, these techniques were not widely adopted because, in most of the cases described here, the number of hyperparameters is on the same order as the number of data points. This property makes it very difficult to apply the kernels to real-world data sets due to the complex algorithms required to fit or optimize such models.

\vspace{2mm}
\noindent
Nonetheless, the potential benefits of applying non-stationary kernels far outweigh the risks in our opinion, and this perspective is all about managing this trade-off. To do so, we now introduce three modern approaches to handle non-stationarity in datasets in order to later test them and compare their performance (Section~\ref{sec:compExp}). 

\subsection{Non-Stationarity via a Parametric Signal Variance}
This class of non-stationary kernels uses a parametric function as the signal variance.
The term $g(\mathbf{x}_1) g(\mathbf{x}_2)$ is always symmetric and psd \cite{noack2022advanced} and is, therefore, a valid kernel function. Also, any product of kernels is a valid kernel, which gives rise to kernels of the form
\begin{equation}\label{eq:nonpara1}
    k(\mathbf{x}_i, \mathbf{x}_j) = g(\mathbf{x}_i) g(\mathbf{x}_j) k_{stat}(\mathbf{x}_i, \mathbf{x}_j).
\end{equation}
This can be seen as a special case of the non-stationary kernel derived in \cite{paciorek2006spatial} and \cite{risser2015regression} wherein the length-scale is taken to be a constant. \cite{risser2015regression}, in particular, consider parametric signal variance and (anisotropic) length scale. In an extension of \eqref{eq:nonpara1}, any sum of kernels is a valid kernel which allows us to write
\begin{equation}\label{eq:nonpara2}
    k(\mathbf{x}_i, \mathbf{x}_j) = \sum_{l=1}^N g_l(\mathbf{x}_i) g_l(\mathbf{x}_j) k_{stat}(\mathbf{x}_i, \mathbf{x}_j).
\end{equation}
The function $g$ can be any function defined on the input domain, but we will restrict ourselves to functions of the form
\begin{equation}\label{eq:basis}
    g(\mathbf{x}) = \sum_{k=1}^{N_2} c_k \beta(\mathbf{x}_k,\mathbf{x}),
\end{equation}
where $c_k$ are some coefficients (or parameters), and $\beta(\mathbf{x}_k,\mathbf{x})$ are basis functions centered at $\mathbf{x}_k$.
For our computational experiments, we use radial basis functions of the form
\begin{equation}
    \beta(\mathbf{x}_k,\mathbf{x}) = e^{-\frac{||\mathbf{x}_k - \mathbf{x}||_2^2}{w}},
\end{equation}
where $w$ is the width parameter.

\subsection{Deep Kernels}

\begin{algorithm}[!t]
%\SetAlgoNlRelativeSize{0}
%\SetNlRelativeSize{-1}
%\SetNlRelativeSize{0}
%\SetAlgoNlRelativeSize{-2}
\caption{Deep Kernel with Neural Network}

\textbf{Input} Set of points as arrays $x_1$, $x_2$, Hyperparameters $h \in \Theta$

\textbf{Output} Deep kernel covariance matrix of shape ($len(x_1) \times len(x_2)$)

\textbf{Step 1:} Define the neural network architecture with input dimension $n$, hidden layers, and appropriate activation functions

\textbf{Step 2:} Initialize or set the weights and biases of the neural network using $h$

\textbf{Step 3:} Transform $x_1$ and $x_2$ through the neural network
  \begin{itemize}
  \item Apply the forward pass of the neural network to all points in sets $x_1$ and $x_2$
  \item Store the transformed values as $x_{1nn}$ and $x_{2nn}$
  \end{itemize}

\textbf{Step 4:} Calculate the pairwise distance matrix between $x_{1nn}$ and $x_{2nn}$

\textbf{Step 5:} Compute the deep kernel value using the distance matrix
  \begin{itemize}
  \item Apply a kernel function (e.g., exponential) to the distance matrix
  \item Scale and combine with other kernel functions if needed
  \end{itemize}

\textbf{Step 6:} Return the deep kernel value\;
\label{alg:deepkernel}
\end{algorithm}

Our version of parametric non-stationarity operates on the signal variance only. This is by design so that we can separate the effects of the different kernels later in our tests. This next approach uses a constant signal variance but warps the input space to yield flexible non-constant length scales. The set of valid kernels is closed under non-linear transformation of the input space as long as this transformation is constant across the domain and the resulting space is considered a linear space. This motivates the definition of kernels of the form
\begin{equation} \label{eq:deepkernel}
    k(\mathbf{x}_i, \mathbf{x}_j)=k_{stat}(||\boldsymbol{\phi}(\mathbf{x}_i) - \boldsymbol{\phi}(\mathbf{x}_j)||_2),
\end{equation}
where $\boldsymbol{\phi}: \mathbb{R}^n \rightarrow \mathbb{R}^{n^*}$ can again be any scalar or vector function on the input space. Deep neural networks have been established as a preferred choice for $\boldsymbol{\phi}$ due to their flexible approximation properties which gives rise to, so-called, deep kernels (see Algorithm \ref{alg:deepkernel} for an implementation example). For our tests, we define a 2-layer-deep network with varying layer widths. While it is possible through deep kernels to perform dimensionality reduction, in this work we map the original input space into a linear space of the same dimensionality, i.e., $n=n^*$. Care must be taken not to use neural networks whose weights and biases, given the dataset set size, are underdetermined. That is why comparatively small networks are commonly preferred. We use ReLu as an activation function. The neural network weights and biases are treated as hyperparameters and are trained accordingly. We set $k_{stat}$ in \eqref{eq:deepkernel} to be the Mat\'ern kernel with $\nu=3/2$. 
\vspace{2mm}
\noindent

Our deep kernel construction shares the perspective of using a neural network to estimate a warping function \citep{sampson1992nonparametric}. In \cite{zammit2022deep}, the authors propose an approach of deep compositional spatial models that differs from traditional warping methods in that it models an injective warping function through a composition of multiple elemental injective functions in a deep-learning framework. This allows for greater flexibility in capturing non-stationary and anisotropic spatial data and is able to provide better predictions and uncertainty quantification than other deep stochastic models of similar complexity. This uncertainty quantification is point-wise, similar to the deep GPs we introduce next.

\subsection{Deep GPs}
%Hengrui
Deep Gaussian process (DGP) models are hierarchical extensions of Gaussian processes where GP layers are stacked --- similar to a neural network --- enhancing modeling flexibility and accuracy \citep{damianou2013deep,dunlop2018deep,jones2023alignment} (more details can be found in the Appendix \ref{app:deepGP}). The DGP model is one of the deep hierarchical models \cite{ranganath2014deep,salakhutdinov2010efficient,teh2004sharing} and consists of a number of variational Gaussian process layers 
defined by a mean function and a stationary covariance (kernel) function. 

Let us first consider a simple 2-layer DGP taking $n$-dimensional inputs $\mathbf{x}$. The first layer uses a constant zero-mean $GP^{(1)}$ for a lower-dimensional representation of the input data having its coordinate components distributed as $f_i^{(1)}\mid \mathbf{x} \sim N_n(0, \Sigma_1(\mathbf{x}))$ and $f^{(1)}=(f_1^{(1)},\cdots,f_{n_1}^{(1)})$ where $n_1$ is the number of nodes in the hidden layer where each component $f_i^{(1)}$ is mutually independent.

The second layer uses a constant zero mean and takes the first layer's output. Its output $f^{(2)}$ is distributed as  $f^{(2)}\mid f^{(1)} \sim N_n(0, \Sigma_2(f^{(1)}))$ and is fed into an independent GP $GP^{(2)}$ as the second layer, generating the model output.
We can impose more than two hidden layers for a single DGP according to our needs, where $f^{(1)}\in \mathbb{R}^{n\times n_1},f^{(2)}\in \mathbb{R}^{n_1\times n_2}, \cdots f^{(L)}\in \mathbb{R}^{n_{L-1}\times n}$ with layouts as random matrices. This is analogous to the connecting matrices in the architecture of a neural network. However, we match the 2-layer structure used in our deep kernel and point out that the complexity of the neural architecture may not always lead to better performance.

For $L>2$ layers in a DGP, we iterate this procedure of using the previous layer's output as the next layer's GP input until reaching the last layer. For a $L$-layer DGP architecture, this creates a sequence of $L$ independent GPs $GP^{(1)},GP^{(2)},\cdots,GP^{(L)}$ corresponding to each layer. Each layer's forward method applies the mean and covariance functions to input data and returns a multivariate normal distribution as output, which serves as the next layer's input.

\vspace{2mm}
\noindent
Although each layer of a DGP is equipped with stationary kernels, the output of one
GP layer becomes the input to the next GP layer, hence the final output will not be stationary. For a DGP with $L$
layers, we can represent the model as follows using unique mean functions $m^{(1)},\cdots,m^{(L)}$ and covariance kernels $k^{(1)},\cdots,k^{(L)}$ that generate $\Sigma_1,\cdots, \Sigma_L$ used in the probabilistic distributions: 
\begin{align}
f^{(1)}_i(\mathbf{x})\sim GP_{n}^{(1)}(m^{(1)}(\mathbf{x}),k^{(1)}(\mathbf{x},\mathbf{x}')),i=1,\cdots,n_1\\
f^{(2)}_i(\mathbf{x})\mid f^{(1)}(\mathbf{x})\sim GP_{n_1}^{(2)}(m^{(2)}(f^{(1)}(\mathbf{x})),k^{(2)}(f^{(1)}(\mathbf{x}),f^{(1)}(\mathbf{x}'))),i=1,\cdots,n_2\\
\cdots\nonumber\\
f^{(L)}_i(\mathbf{x}) \mid f^{(L-1)}(\mathbf{x})\sim GP_{n_{L-1}}^{(L)}(m^{(L)}(f^{(L-1)}(\mathbf{x})),k^{(L)}(f^{(L-1)}(\mathbf{x}),f^{(L-1)}(\mathbf{x}'))),i=1,\cdots,n
\end{align}
Then, an optimizer and the variational Evidence Lower Bound (ELBO) are used for training the DGP  model. Using a variational approximation in ELBO, 
instead of exact inference, leads to manageable computational complexity of deeper GP models \citep{titsias2010bayesian}. The deep Gaussian Processes (GPs) can be perceived as hierarchical models whose kernel does not admit a closed form. Crucially, this ``hierarchy of means'' is constructed via the means of the layer-distributions in the deep GP, but not higher moments like \cite{allenby2006hierarchical,daniels1999nonconjugate, mohamed2015statistical}. Only the mean functions at each layer of the deep GP are contingent upon computations from preceding layers, signifying hierarchies that rely on the first-order structure at every layer of the model.

\vspace{2mm}
\noindent
Using a slightly different framework based on the Vecchia approximation, \cite{jimenez2023vecchia} introduced a ``deep Vecchia ensemble,'' a hybrid method combining Gaussian processes (GPs) and deep neural networks (DNNs) for regression. This model builds an ensemble of GPs on DNN hidden layers, utilizing Vecchia approximations for GP scalability. Mathematically, the joint distribution of variables is decomposed using Vecchia's approximation, and predictions are combined using a generalized product of experts. This approach offers both representation learning and uncertainty quantification.
As described in the last section, a GP model can utilize a deep kernel, constructively combining the neural network's and the GP's strength, leading to a model that benefits from the GP's interpretability and the NN's flexibility \citep{wilson2016deep}. 
\vspace{2mm}
\noindent

Our Bayesian DGP (BDGP) architecture follows \cite{sauer2022vecchia} and includes a two-layer neural network, applied as a transformation to the input data. The first layer uses a rectified linear unit (ReLU) activation function and the second employs a sigmoid activation. This non-linear feature mapping expresses complex input space patterns (See Appendix \ref{sec:appB}). 

\vspace{2mm}
\noindent
Contrasting DGPs with deep-kernel GPs, DGPs use multiple GP layers to capture intricate dependencies, whereas deep-kernel GPs employ a NN for input data transformation before GP application. Essentially, while DGPs exploit GP layering to manage complex dependencies, deep kernel learning leverages NNs for non-linear input data transformation, enhancing the GP's high-dimensional function representation ability.

\subsection{Measuring Non-Stationarity of Datasets}
When it comes to characterizing non-stationarity, some
methods focus on non-stationarity in the mean function
(e.g., polynomial regression), while others concentrate on the non-stationarity
in the variance (e.g., geographically weighted regression \citep{fotheringham2003geographically}).
Non-stationarity is typically characterized by a change in statistical
properties over the input space,  e.g., changes in the  
dataset's mean,
variance, or other higher moments.
Quantifying non-stationarity is an active area of research and in  
this paper, we introduce a particular kind of non-stationarity measure for the purpose of judging our test kernels when applied to our test datasets
Overall,
measuring a given dataset's non-stationarity properties is an important ingredient in understanding the performance of a particular kernel. 
For the reader's convenience, we offer our non-stationarity measure as a pseudocode (see Algorithm \ref{alg:NonStationarityCheck}). For detailed theoretical motivation of the non-stationarity measure we use in this work, please refer to Appendix \ref{sec:AppNonStat}.

\vspace{2mm}
\noindent
To avoid bias through user-based subset selection we draw the location of a subdomain from a uniform distribution over the input domain (in this case $[0,1]^{n}$). We then select a set of data points randomly from within this subdomain 100 times and use MLE (maximum likelihood estimation, maximizing Equation \eqref{eqn:log_likelihood}) to get a stationary length scale and signal variance in each iteration. In the case of a synthetic function, the locations of the data points are drawn from a uniform distribution over the subdomain; in the case of a fixed dataset, the points are randomly selected. The final distribution of all signal variances and length scales is then assessed to measure non-stationarity (See Algorithm \ref{alg:NonStationarityCheck}).

\begin{algorithm}
\caption{Measuring non-stationarity via local, stationary-GP hyperparameter distributions.}\label{alg:NonStationarityCheck}
\begin{algorithmic}[1]
\Procedure{MeasureNonStationarity}{$\mathbf{X}$, $\mathbf{y}, m, size, c$}\Comment{$\mathbf{X}$ is the set of data points, $\mathbf{y}$ is the collected data, $m$ is the number of iterations, $size$ is the size of the subdomain, $c$ is the number of selected points in each iteration}
\State length scale list = []
\State signal variance list = []
\For{$\textit{i}$ in $0$ to $m$}
\State $a \sim U([0,1-size]^n)$ 
\State $b = a + size$
\State select $c$ test data points $\mathbf{X}_t$ from $[a,b]^n$: 
\State select associated $\mathbf{y}_t$ in dataset or evaluate synthetic function at $\mathbf{X}_t$
\State initialize a stationary GP
\State signal variance, length scale = run\_MLE($\mathbf{X}_t,\mathbf{y}_t$)
\State append new signal variance to signal variance list
\State append new length scale to length scale list
\EndFor
\State \textbf{return} signal variance list, length scale list
\EndProcedure
\end{algorithmic}
\end{algorithm}

\begin{figure}[ht!]
    \centering
     \begin{subfigure}[b]{0.3 \textwidth}
         \centering
         \includegraphics[height = 5cm, width = 5cm]{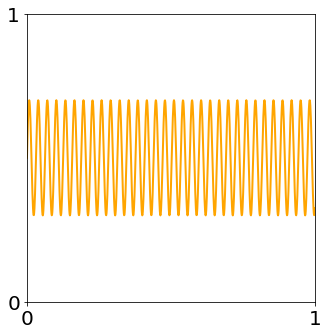}
     \end{subfigure}
     \hfill
    \begin{subfigure}[b]{0.3 \textwidth}
         \centering
         \includegraphics[height = 5cm, width = 5cm]{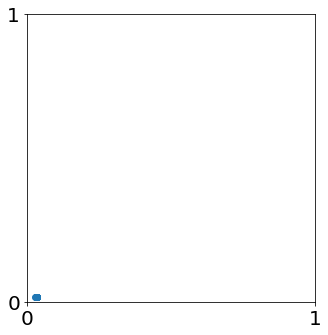}
     \end{subfigure}
     \hfill
     \begin{subfigure}[b]{0.15 \textwidth}
         \centering
         \includegraphics[height = 5cm, width = 2cm]{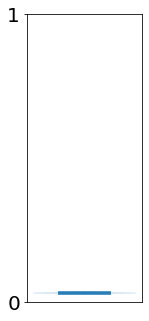}
     \end{subfigure}
     \hfill
     \begin{subfigure}[b]{0.15 \textwidth}
         \centering
         \includegraphics[height = 5cm, width = 2cm]{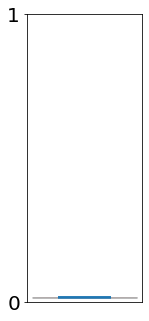}
     \end{subfigure}
     \put(-380,140){test function}
     \put(-230,140){scatter plot}
     \put(-90,140){violin plots}
     \put(-250,5){signal variance}
     \put(-125,20){\rotatebox{90}{signal variance, var=2e-6}}
     \put(-45,20){\rotatebox{90}{length scale, var=6e-8}}
     \put(-285,20){\rotatebox{90}{length scale}}
     
     \begin{subfigure}[b]{0.3 \textwidth}
         \centering
         \includegraphics[height = 5cm, width = 5cm]{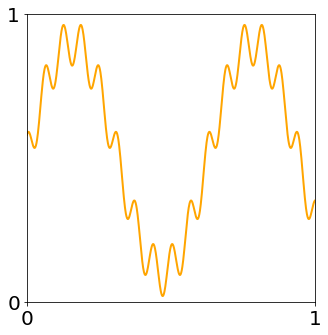}
     \end{subfigure}
     \hfill
    \begin{subfigure}[b]{0.3 \textwidth}
         \centering
         \includegraphics[height = 5cm, width = 5cm]{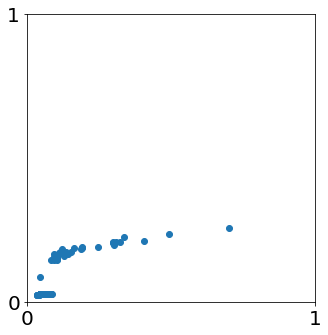}
     \end{subfigure}
     \hfill
     \begin{subfigure}[b]{0.15 \textwidth}
         \centering
         \includegraphics[height = 5cm, width = 2cm]{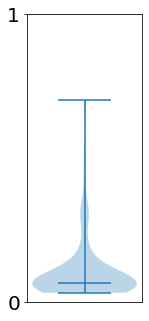}
     \end{subfigure}
     \hfill
     \begin{subfigure}[b]{0.15 \textwidth}
         \centering
         \includegraphics[height = 5cm, width = 2cm]{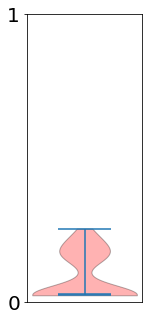}
     \end{subfigure}
     \put(-250,5){signal variance}
     \put(-125,20){\rotatebox{90}{signal variance, var=0.02}}
     \put(-45,20){\rotatebox{90}{length scale, var=0.006}}
     \put(-285,20){\rotatebox{90}{length scale}}

     \begin{subfigure}[b]{0.3 \textwidth}
         \centering
         \includegraphics[height = 5cm, width = 5cm]{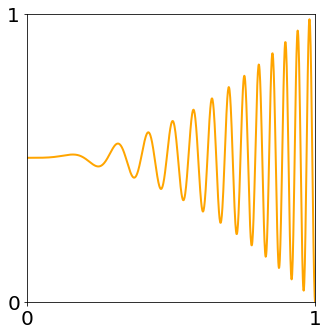}
     \end{subfigure}
     \hfill
    \begin{subfigure}[b]{0.3 \textwidth}
         \centering
         \includegraphics[height = 5cm, width = 5cm]{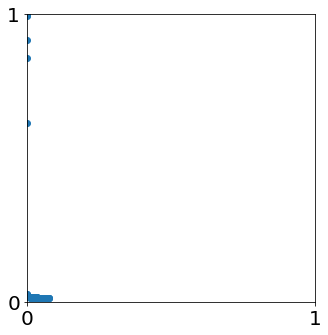}
     \end{subfigure}
     \hfill
     \begin{subfigure}[b]{0.15 \textwidth}
         \centering
         \includegraphics[height = 5cm, width = 2cm]{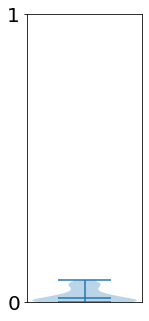}
     \end{subfigure}
     \hfill
     \begin{subfigure}[b]{0.15 \textwidth}
         \centering
         \includegraphics[height = 5cm, width = 2cm]{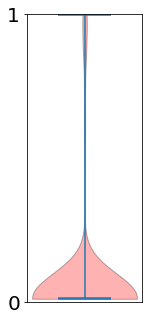}
     \end{subfigure}
     \put(-250,5){signal variance}
     \put(-125,20){\rotatebox{90}{signal variance, var=5e-4}}
     \put(-45,20){\rotatebox{90}{length scale, var=0.04}}
     \put(-285,20){\rotatebox{90}{length scale}}
     \caption{Our way of measuring the non-stationarity of a dataset or a synthetic function. When a set of data points is drawn randomly from within the input domain and a GP, using a stationary kernel, is trained via log marginal likelihood maximization after each draw (MLE), the final distribution of the hyperparameters --- here a signal variance and an isotropic length scale --- can be used to measure non-stationarity. This distribution can be visualized via scatter (middle) and violin (right) plots. We assign the variance of these distributions as a numerical measure of stationarity. The stationary function (top) leads to a very narrow distribution of the length scale and the signal variance. For the non-stationary functions (second and third row), the distributions for both hyperparameters are broader and the associated variances are larger. In this figure, the axis scales are kept constant to allow easy comparison of the stationarity properties of the test functions.}
     \label{fig:3_cases}
\end{figure}
 
\vspace{2mm}
\noindent
To test our non-stationary measure, we applied it to three synthetic functions (see Figure \ref{fig:3_cases}). 
In each case, we indicate the final variance of the length scale and signal variance distributions.
In the first scenario (first row in Figure \ref{fig:3_cases}), where the signal is a high-frequency sine function, the algorithm's behavior leads to a high concentration of points in a single compact cluster when plotting the estimated length scale versus the signal variance. This concentration reflects the inherent stationarity of a signal, where the statistical properties do not change over the input space. The unimodal and concentrated distributions of both estimated parameters in the violin plots further corroborate this observation. The consistency in the length scale and signal variance across multiple iterations of MLE indicates that the underlying data structure is stationary.

\vspace{2mm}
\noindent
In the second case (second row in Figure \ref{fig:3_cases}), the signal is a trigonometric curve with local oscillations. The algorithm's response to this function results in a wider spread of both hyperparameters compared to the first test function. This behavior is reflected in the violin plots on the right.

\vspace{2mm}
\noindent
The third scenario (third row in Figure \ref{fig:3_cases}) presents a more complex signal structure, a trigonometric curve with varying amplitude and frequency. The algorithm's reaction to this non-stationary signal is manifested in a broad distribution of length scales and some spread in the signal variance. These distributions are reflected in the violin plots.

\vspace{2mm}
\noindent
The three cases in Figure \ref{fig:3_cases} demonstrate the algorithm's capability to measure non-stationarity through the local, stationary GP-hyperparameter distributions. The varying behaviors in clustering and distribution of the estimated parameters across the three cases provide insights into the underlying stationarity or non-stationarity of the signals. The algorithm's sensitivity to these variations underscores its potential as a valuable tool for understanding and characterizing non-stationarity in different contexts.

\subsection{Performance Measures}
Throughout our computational experiments, we will measure the performance via three different error metrics as a function of training time. We argue that this allows us to compare methodologies across different implementations as long as all tests are run on the same computing architecture with similar hardware utilization. As for error metrics, we utilize the log marginal likelihood (LML, Equation \ref{eqn:log_likelihood}), the root mean square error (RMSE), and the Continuous Ranked Probability Score (CRPS). The RMSE and CRPS are evaluated on test data points that are not part of the training data set. The RMSE is defined as
\begin{equation}\label{eq:rmse}
    RMSE= \sqrt{\frac{\sum_i^N (y_i - f^i_0)^2}{N}},
\end{equation}
where $y_i$ are the data values of the test dataset and $f^i_0$ are the posterior mean predictions. The RMSE metric provides a measure of how closely the model's predictions align with the actual values --- approaching zero as fit quality improves --- while the log marginal likelihood evaluates the fit of the Gaussian Process model given the observed data. The log marginal likelihood will increase as the model fits the data more accurately. 
The CRPS is defined as
\begin{equation}\label{eq:crps}
CRPS(f_0,y_i) = \sigma  \big( \frac{1}{\sqrt{\pi}} -2 \psi(\frac{y_i-\mu}{\sigma}) -\frac{y_i-\mu}{\sigma}  (2 \Psi(\frac{y_i-\mu}{\sigma})-1) \big),
\end{equation}
where $\psi$ is the probability density function of a standard normal distribution and $\Psi$ is the associated cumulative distribution function. For a GP, $f_0$ is Gaussian with mean $\mu$ and variance $\sigma^2$. The CRPS is negative and approaches zero as fit quality improves. 
In our computational experiments, we use the CRPS as a performance metric on a set of points by averaging \eqref{eq:crps} over all predictions. The CRPS is arguably the more important score compared to the RMSE because it is \emph{uncertainty aware}. In other words, if the prediction accuracy is low, but uncertainty in those regions is high --- the algorithm is aware of its inaccuracy --- the score improves.

\vspace{2mm}
\noindent
Computational cost is becoming a main research topic in recent studies in large-scale non-parametric models \citep{hrluo_2022e,Scott2016cmcmc,hensman2015scalable}, especially GPs \citep{litvinenko2019likelihood,geoga2020scalable,lin2023sampling,luo2022sparse}. 
In our analysis, we sought to examine the progression of the optimization process. To achieve this, we established a callback function during the optimization phase, tracking the RMSE, the log marginal likelihood, and the CRPS as a function of compute time. In terms of interpretation, ideally, we expect the RMSE and the CRPS to increasingly approach zero over time, suggesting that the model's predictive accuracy and estimation of uncertainties are improving. On the other hand, the log marginal likelihood should increase, indicating a better fit of the model to the observed data. This analysis gives us a summary of the model's learning process and helps us understand the progression of the optimization, thus providing valuable insights into the efficacy of our model and the optimization strategy employed.

%%%%%%%%%%%%%%%%%%%%%%%%%%%%%%
%%%%%%%%%%%%%%%%%%%%%%%%%%%%%%
%%%%%%%%%%%%%%%%%%%%%%%%%%%%%%
%%%%%%%%%%%%%%%%%%%%%%%%%%%%%%
\section{Computational Experiments} \label{sec:compExp}
%%%%%%%%%%%%%%%%%%%%%%%%%%%%%%
%%%%%%%%%%%%%%%%%%%%%%%%%%%%%%
%%%%%%%%%%%%%%%%%%%%%%%%%%%%%%
%%%%%%%%%%%%%%%%%%%%%%%%%%%%%%
The purpose of this section is to see how different kernels and a deep GP deal with non-stationarity in several datasets and to compare the characteristics and properties of the solutions. To make this comparison fair and easier, we ran all tests on the same Intel i9 CPU (Intel Core i9-9900KF CPU $@$ 3.60GHz $\times$ 8) and used the total compute time as the cost. As a performance metric, we calculate and observe the RMSE (Root Mean Squared Error, Equation \eqref{eq:rmse}), the CRPS (continuous rank probability score, Equation \eqref{eq:crps}) of the prediction, both applied to a test dataset, and the log marginal likelihood of the observational data (Equation \eqref{eqn:log_likelihood}). We attempted to run fair tests in good faith; this means, the effort spent to set up each kernel or methodology was roughly proportional to a method's perceived complexity, within reasonable bounds, similar to the effort expended by an ML practitioner --- this meant minutes of effort for stationary kernels and hours to days for non-stationary kernels and DGPs. The optimizer to reach the final model was \emph{scipy}'s differential evolution. We used an in-house MCMC (Markov Chain Monte Carlo) algorithm to create the plots showing the evolution of the performance metrics over time. In cases when our efforts did not lead to satisfactory performance, we chose to present the result ``as is'' to give the reader the ability to judge for themselves. 
We always start our tests with a standard stationary kernel to create a baseline --- stationary kernels with few hyperparameters are currently the most widely used kernel class.
To further the hands-on aspect of this section, we also included the used algorithms in the Appendix and on a specifically designed website together with links to download the data. The performance-measure-over-time plots were created without considering deep GPs due to incompatible differences in the implementations (see Appendix \ref{app:deepGP}). All computational experiments were run multiple times to make sure we showed representative results. 

\vspace{2mm}
\noindent
This section, first, introduces three datasets we use later to evaluate the performance of the test methodologies. Second, we present the unredacted, uncensored results of the test runs. The purpose is not to judge some kernels or methodologies as better or worse universally, but to evaluate how these techniques perform when tested under certain well-defined conditions and under the described constraints. We encourage the reader to follow our tests, to rerun them if desired, and to judge the performance of the methods for themselves.

%%%%%%%%%%%%%%%%%%%%%%%%%%%%%%%%%%%%%%%%%%%%%%%%%%%%%%%%%%%%%%%%%%%%%%%%
\subsection{Introducing the Test Datasets}\label{sec:datasets}
%%%%%%%%%%%%%%%%%%%%%%%%%%%%%%%%%%%%%%%%%%%%%%%%%%%%%%%%%%%%%%%%%%%%%%%%
We will consider three test data sets. 
All datasets are normalized such that the range and the image --- the set of all measured function values --- are in $[0,1]$.

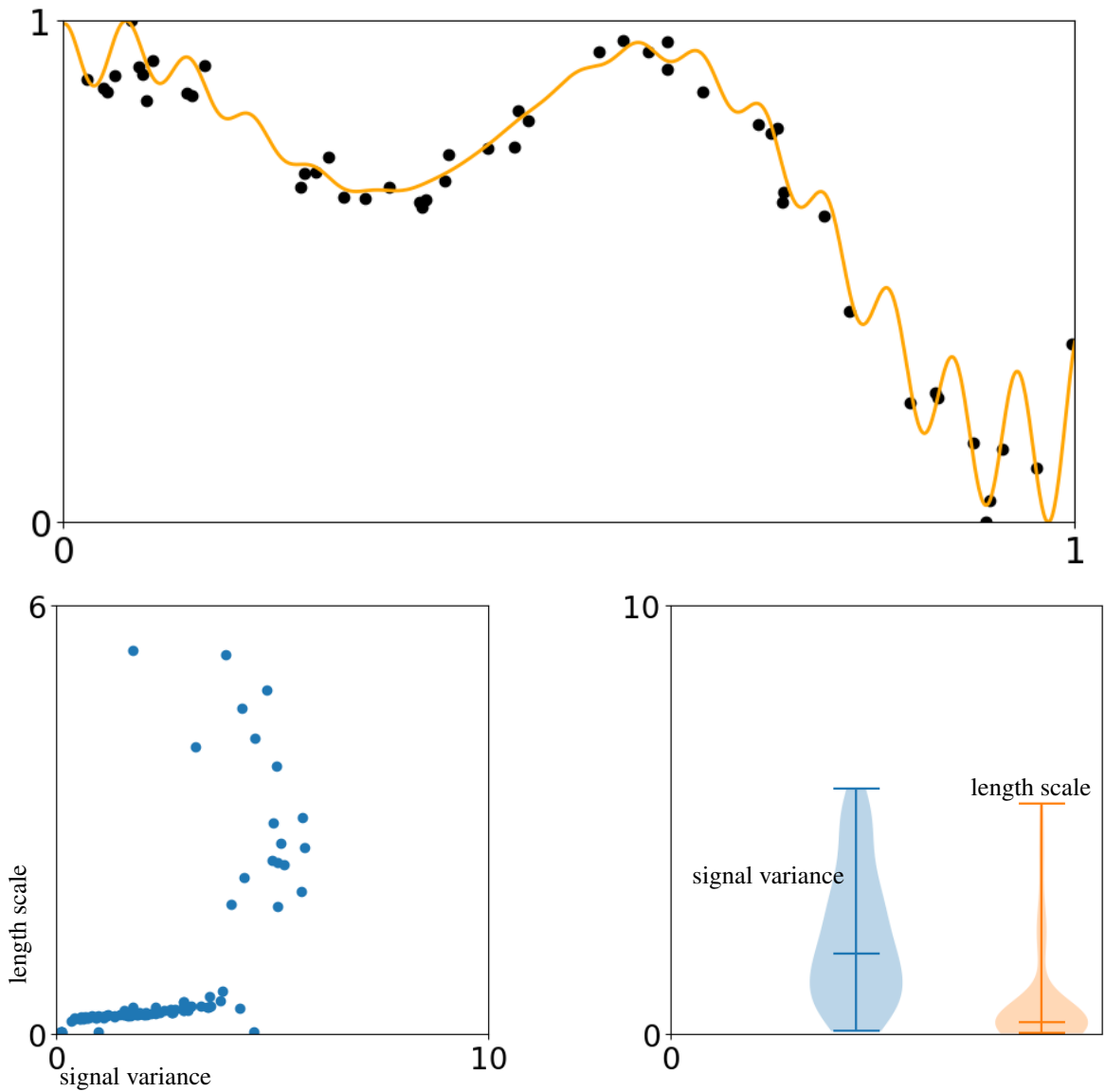
\begin{figure}[ht!]
    \centering
     \begin{subfigure}[b]{\textwidth}
         \centering
         \includegraphics[width=\textwidth]{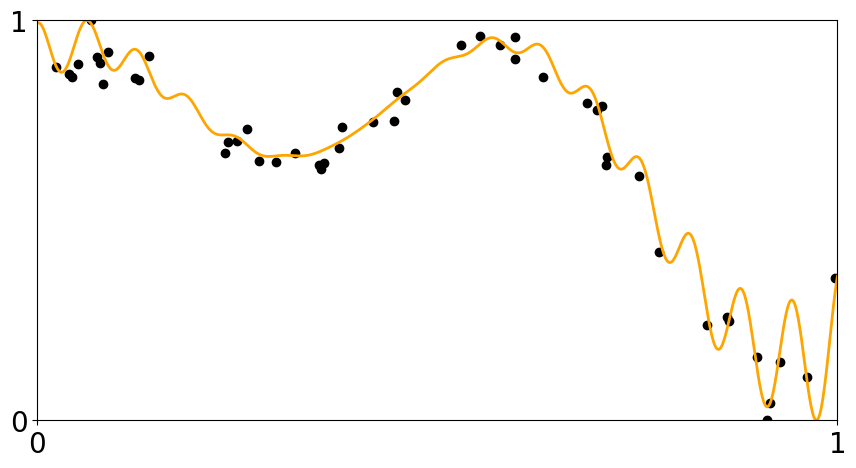}
     \end{subfigure}
     \put (-410,230){a)}

     \begin{subfigure}[b]{0.45\textwidth}
         \centering
         \includegraphics[height = 7cm]{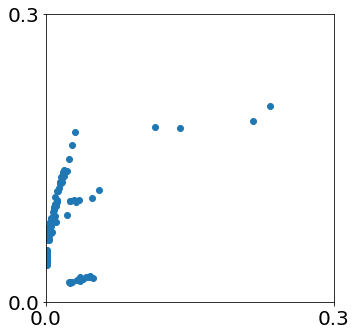}
     \end{subfigure}
     \hfill
     \begin{subfigure}[b]{0.2\textwidth}
         \centering
         \includegraphics[height = 7cm]{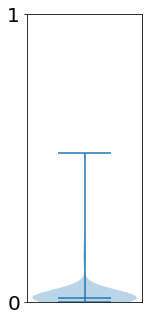}
     \end{subfigure}
     \hfill
     \begin{subfigure}[b]{0.2\textwidth}
         \centering
         \includegraphics[height = 7cm]{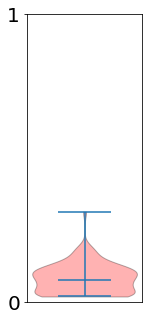}
     \end{subfigure}
     \put (-400,195){b)}
     \put (-177,195){c)}
     \put (-60,195){d)}
     \put (-420,40){\rotatebox{90}{length scale} }
     \put (-380,5){signal variance}
     \put (-180,40){\rotatebox{90}{signal variance, var=0.003}}
     \put (-60,40){\rotatebox{90}{length scale, var=0.002}}
     \caption{Test dataset 1 (a) and its non-stationarity measures visualized as distributions (b, c, d) in the hyperparameters of a stationary kernel trained on local subsets of that data. The dataset is derived from a one-dimensional synthetic function (see Equation \eqref{eq:synthfunc}). Non-stationarity appears to be clearly present in the length scale and the signal variance.}
     \label{fig:syntheticData}
\end{figure}

\vspace{2mm}
\noindent
For the first dataset, we define a one-dimensional synthetic function 
\begin{equation}
    f(x) = (\sin(5x) + \cos(10x) + (2 (x-0.4)^2)  \cos(100 x) + 2.597)/3.94
    \label{eq:synthfunc}
\end{equation}
data is drawn from. 50 data points are drawn randomly and  the noise $\epsilon \sim \mathcal{N}(0,0.001)$ is added. Figure \ref{fig:syntheticData} presents the function and its non-stationary measures.

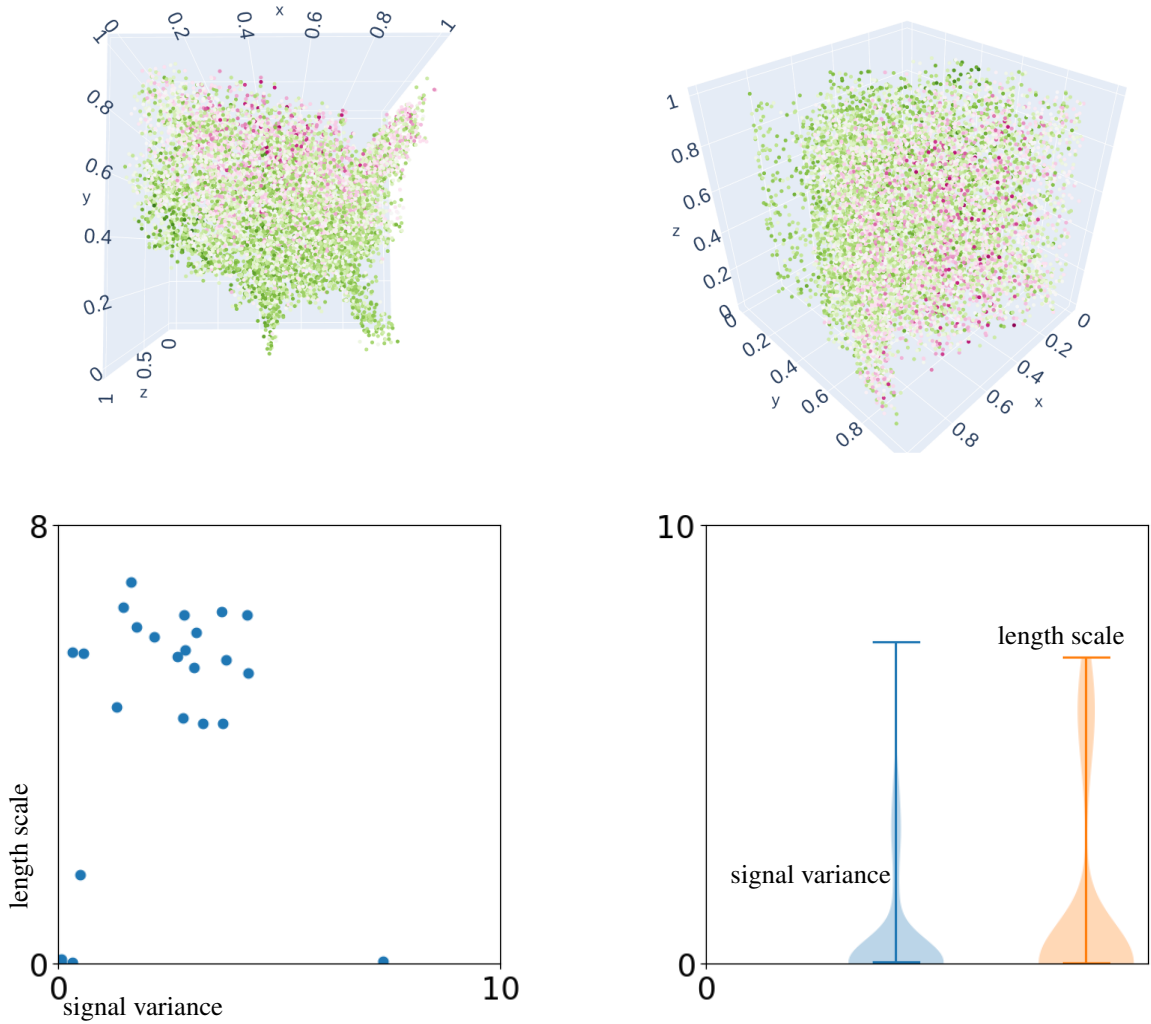
\begin{figure}
    \centering
     \begin{subfigure}[b]{0.45 \textwidth}
         \centering
         \includegraphics[trim={2cm 0 2cm 2cm},clip,width=\textwidth]{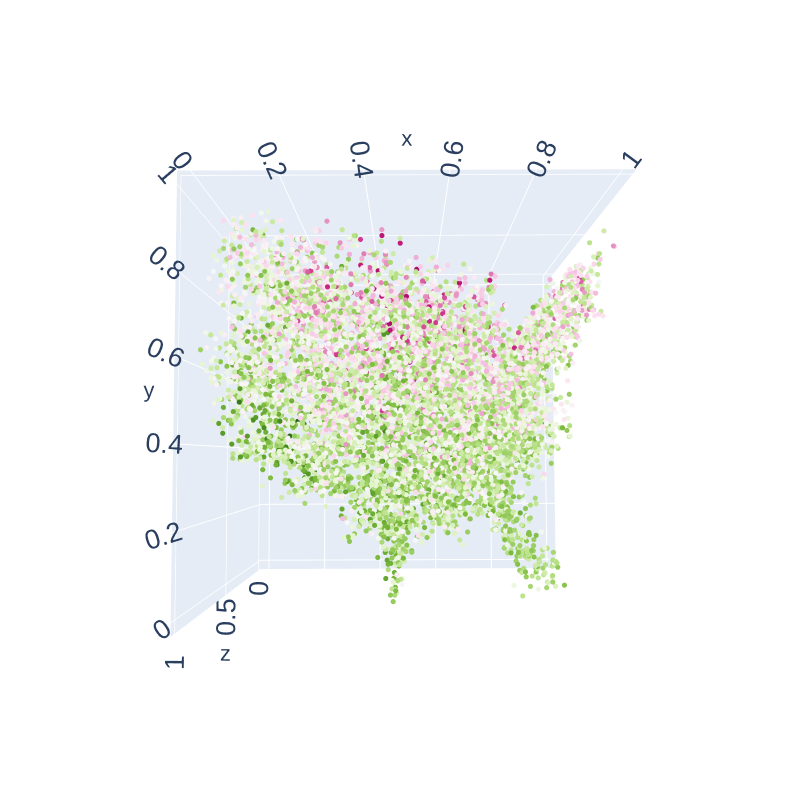}
     \end{subfigure}
     \hfill
     \begin{subfigure}[b]{0.45\textwidth}
         \centering
         \includegraphics[trim={2cm 0 2cm 2cm},clip,width=\textwidth]{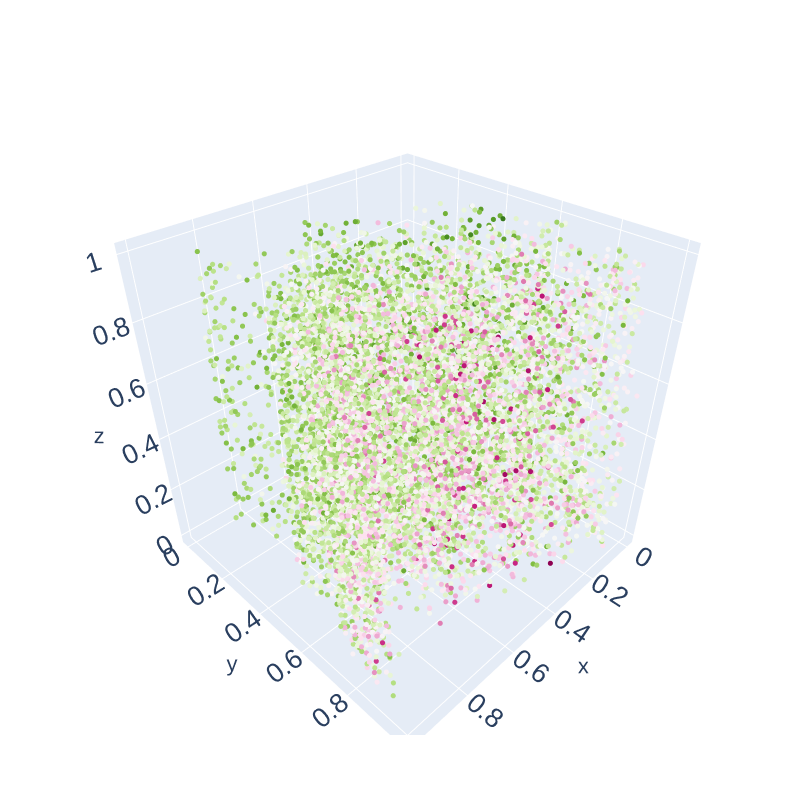}
     \end{subfigure}
     \put (-410,200){a)}
     \put (-190,200){b)}
 
     \begin{subfigure}[b]{0.45\textwidth}
         \centering
         \includegraphics[height = 7cm, width=\textwidth]{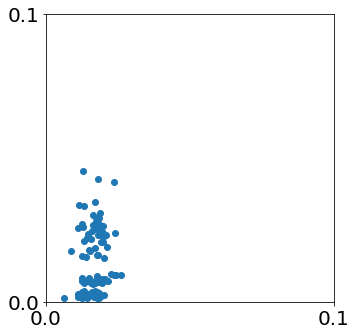}
     \end{subfigure}
     \hfill     
     \begin{subfigure}[b]{0.2\textwidth}
         \centering
         \includegraphics[height = 7cm]{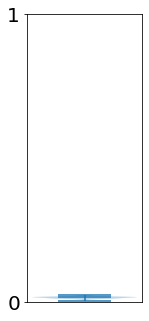}
     \end{subfigure}
     \hfill
     \begin{subfigure}[b]{0.2\textwidth}
         \centering
         \includegraphics[height = 7cm]{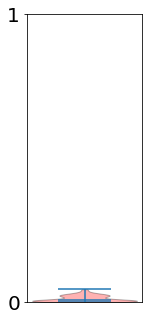}
     \end{subfigure}
     \put (-410,200){c)}
     \put (-190,200){d)}
     \put (-70,200){e)}
     \put (-420,40){\rotatebox{90}{length scale}}
     \put (-380,5){signal variance}
     \put (-180,20){\rotatebox{90}{signal variance, var=1.1e-05}}
     \put (-60,20){\rotatebox{90}{length scale, var=0.0001}}
     \caption{Test dataset 2 (a, b) and its non-stationarity measures visualized as distributions in the hyperparameters (c, d, e) of a stationary kernel trained on local subsets of that data. The dataset consists of recorded temperatures across the United States and a period of time. Weak non-stationarity appears to be present in the length scale and the signal variance.}
     \label{fig:climateData}
\end{figure}

\vspace{2mm}
\noindent
Second, we consider a three-dimensional climate dataset that is available online (\url{https://www.ncei.noaa.gov/data/global-historical-climatology-network-daily/}), 

consisting of in situ measurements of daily maximum surface air temperature ($^\circ$C) collected from weather stations across the contiguous United States (geospatial locations defined by longitude and latitude) over time (the third dimension). 
The data and its non-stationarity measures are presented in Figure \ref{fig:climateData}.

\begin{figure}[ht!]
    \centering
     \begin{subfigure}[t]{0.8 \textwidth}
         \centering
         \includegraphics[trim={2cm 0 2cm 7cm},clip,width=0.8\textwidth]{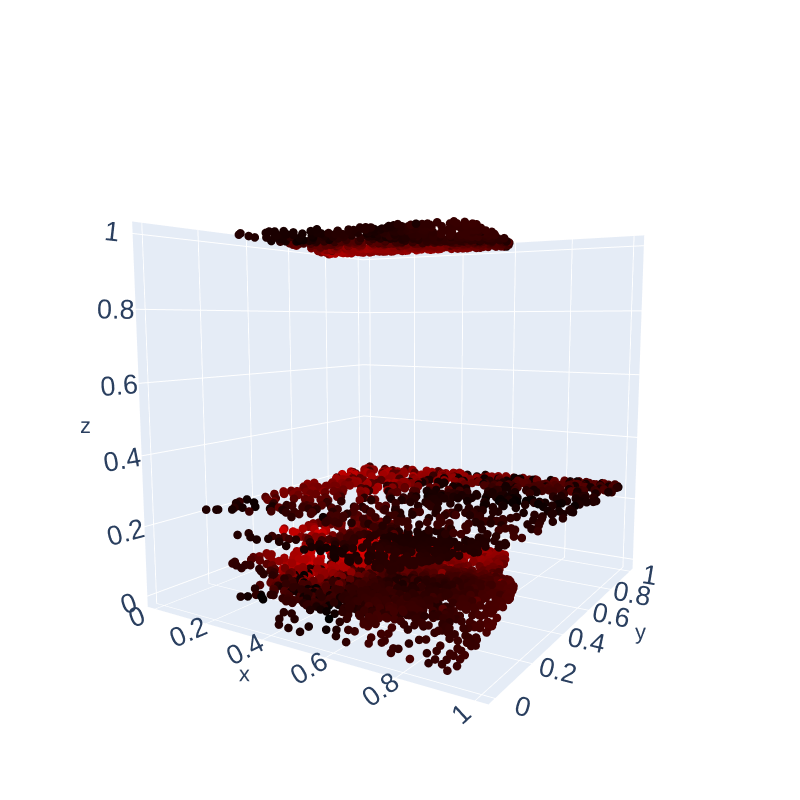}
     \end{subfigure}
     \put (-310,230){a)}
 
     \begin{subfigure}[b]{0.45\textwidth}
         \centering
         \includegraphics[height = 7cm, width=\textwidth]{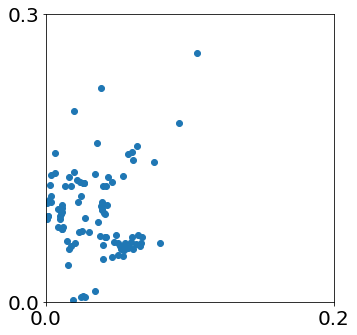}
     \end{subfigure}
     \hfill     
     \begin{subfigure}[b]{0.2\textwidth}
         \centering
         \includegraphics[height = 7cm]{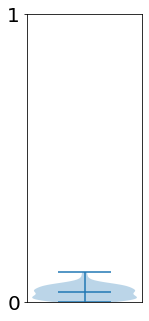}
     \end{subfigure}
     \hfill
     \begin{subfigure}[b]{0.2\textwidth}
         \centering
         \includegraphics[height = 7cm]{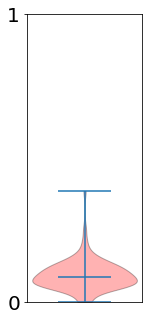}
     \end{subfigure}
     \put (-410,200){b)}
     \put (-190,200){c)}
     \put (-70,200){d)}
     \put (-420,40){\rotatebox{90}{length scale}}
     \put (-380,5){signal variance}
     \put (-180,20){\rotatebox{90}{signal variance, var=5e-5}}
     \put (-60,20){\rotatebox{90}{length scale, var=0.0027}}
     \caption{Test dataset 3 (a) and its non-stationarity measures visualized as distributions in the hyperparameters (b, c, d) of a stationary kernel trained on local subsets of that data. The dataset consists of analyzed X-ray scattering signals over $[0,1]^3$. Non-stationarity is apparent in both length scale and signal variance.}
     \label{fig:xrayData}
\end{figure}
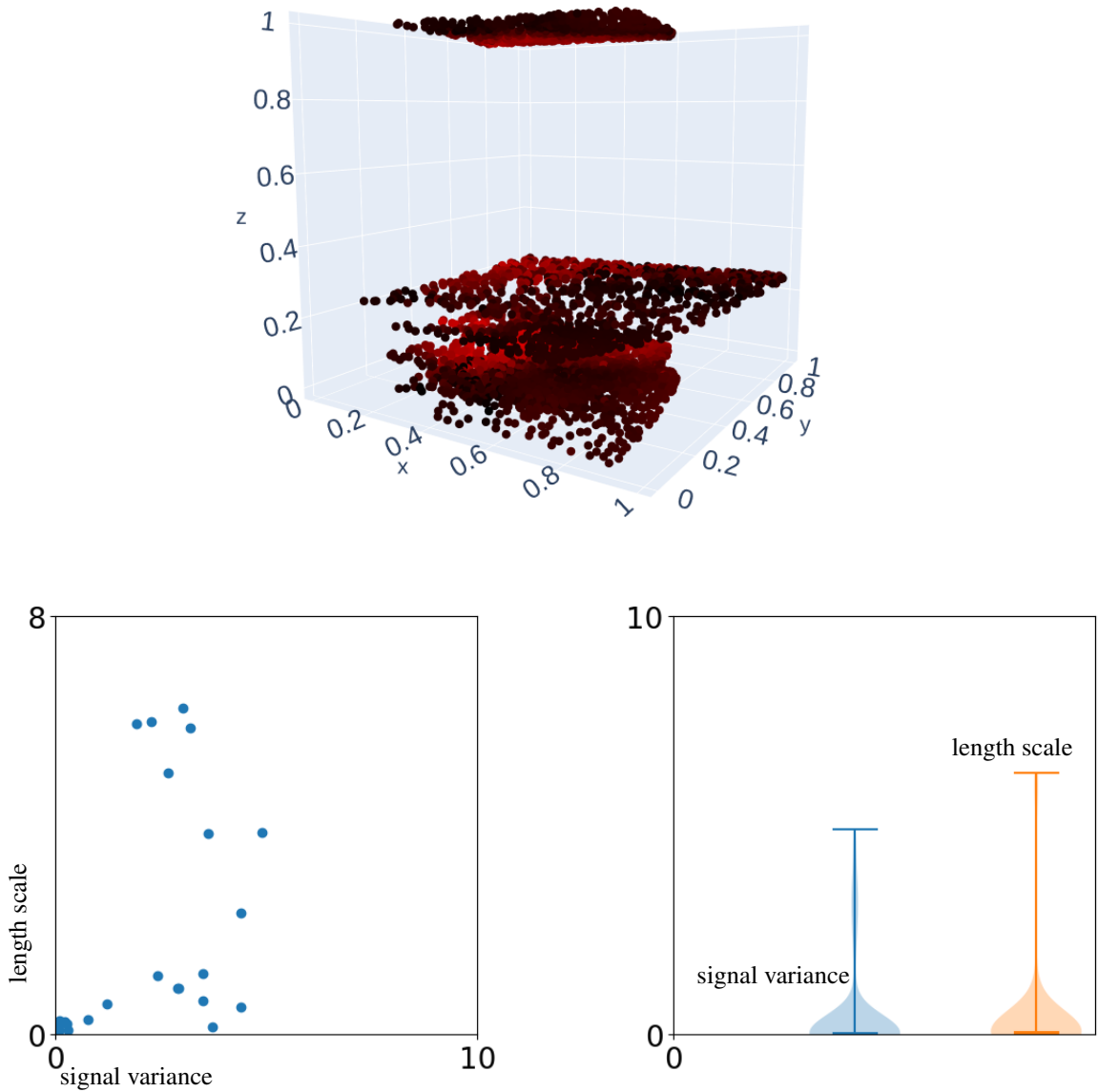

\vspace{2mm}
\noindent
Third, we consider a dataset that was collected during an X-ray scattering experiment at the CMS beamline at NSLSII, Brookhaven National Laboratory. The dataset originated from an autonomous exploration of multidimensional material state-spaces underlying the self-assembly of copolymer mixtures. Because the scientific outcome of this experiment has not been published yet, all scientific insights have been obscured by normalization and the removal of units.
The dataset is presented in Figure \ref{fig:xrayData}.

\vspace{2mm}
\noindent
To evaluate the predictive performance metrics, we divided our datasets into training datasets and test datasets by random selection and applied the metrics only to the latter.

\subsection{Results}
In this section, we present quantitative results of how the different kernels and deep GPs performed when tasked to learn the underlying data-generating latent functions that produced the test datasets introduced in the previous section. For each test, we show the model and its uncertainties across the domain or a subdomain after convergence of a global evolutionary optimization of the log marginal likelihood, and the performance measures as a function of the compute time of an MCMC algorithm. Code snippets can be found in the Appendix and on our website (see the Code Availability paragraph at the end). To reiterate, for all tests, we put ourselves in the position of an ML practitioner setting up each algorithm under a \emph{reasonable-effort} constraint --- we did not optimize each method to its full extent because this would not lead to a fair comparison. However, we used our best judgment and followed online documentation closely. The same thought process was put into the exact design of the kernels; of course, one could always argue that a particular model would have performed better using more hyperparameters. For the fairness of the comparison, we kept the number of hyperparameters similar across the kernels for a particular computational experiment and increased the number of kernel parameters in a near-proportional fashion for the non-stationary kernels as we moved to higher dimensions. See Table \ref{Tbl:1} for the number of hyperparameters for the different experiments.
All non-stationary kernels were implemented in the open-source GP package gpCAM (\url{https://github.com/lbl-camera/gpCAM}).
We tried two different deep GPs, the gpflux package (\url{https://github.com/secondmind-labs/GPflux}) and the Bayesian deep GP (BDGP) by \cite{sauer2022vecchia,sauer2023non} (\url{https://cran.r-project.org/web/packages/deepgp/index.html}). We selected the latter in its two-layer version for our final comparisons because of performance issues with the gpflux package (see Figure \ref{fig:gpfluxdgp}).

\begin{table}[!t]
    \caption{Number of hyperparameters for our computational experiments.}
% \vskip0.5ex
\begin{center}
\begin{tabular}{lccc}
% \hline\noalign{\smallskip}
 & \multicolumn{3}{c}{\textbf{Number of hyperparameters per experiment}}  \\
{\textbf{Kernel functions}} & \textit{1D Synthetic} & \textit{3D Climate} & \textit{3D X-ray}    \\
\hline\noalign{\smallskip} 
Stationary,  $k_{stat}$    & 2         & 3      & 3\\ 
Parametric non-stationary,  $k_{para}$   & 15     & 58         & 58\\
Deep kernel,  $k_{nn}$    & 48      & 186      & 186\\ 
\noalign{\smallskip}\hline
\end{tabular}
\end{center}
\label{Tbl:1}
\end{table}

%%%%%%%%%%%%%%%%%%%%%%%%%%%%%%%%%
\subsubsection{One-Dimensional Synthetic Function}\label{sec:1dSynth}
\begin{figure}[H]
    \centering
     \begin{subfigure}[b]{0.45 \textwidth}
         \centering
         \includegraphics[width = \textwidth]{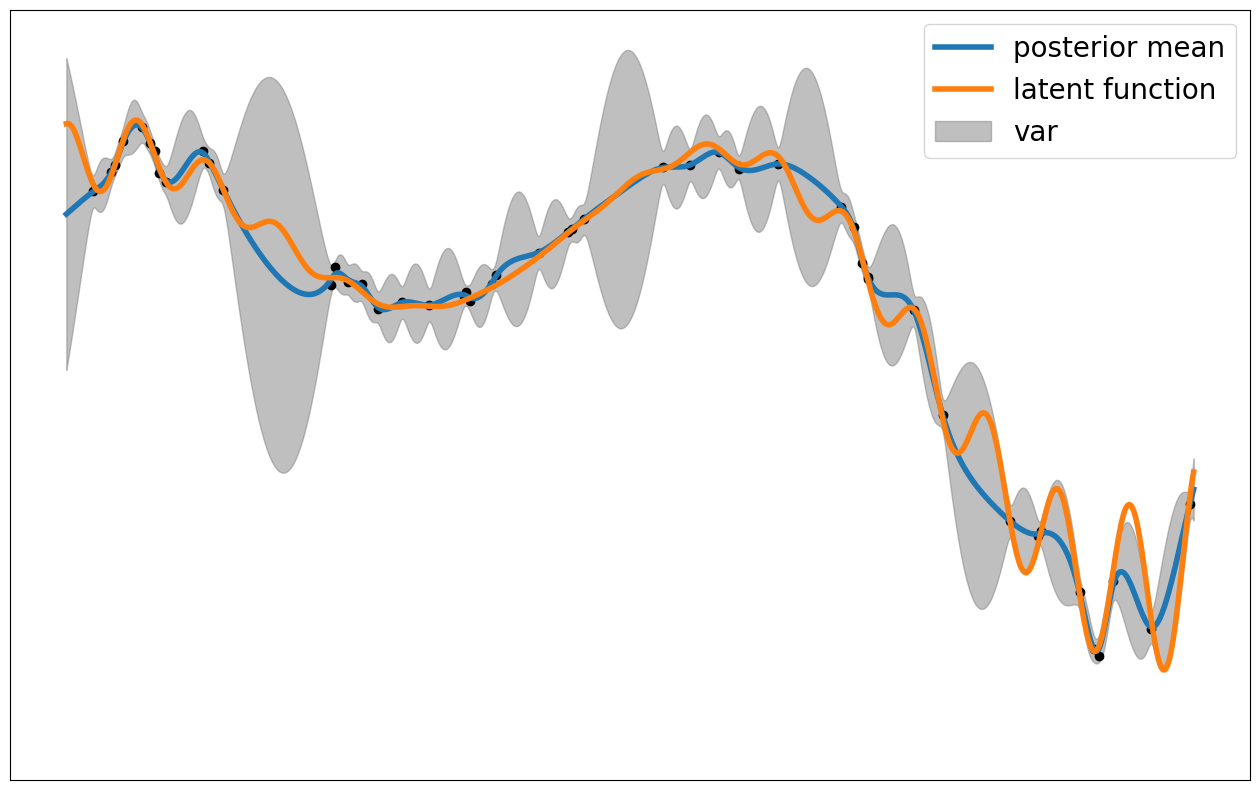}
     \end{subfigure}
     \hfill
     \begin{subfigure}[b]{0.45\textwidth}
         \centering
         \includegraphics[width=\textwidth]{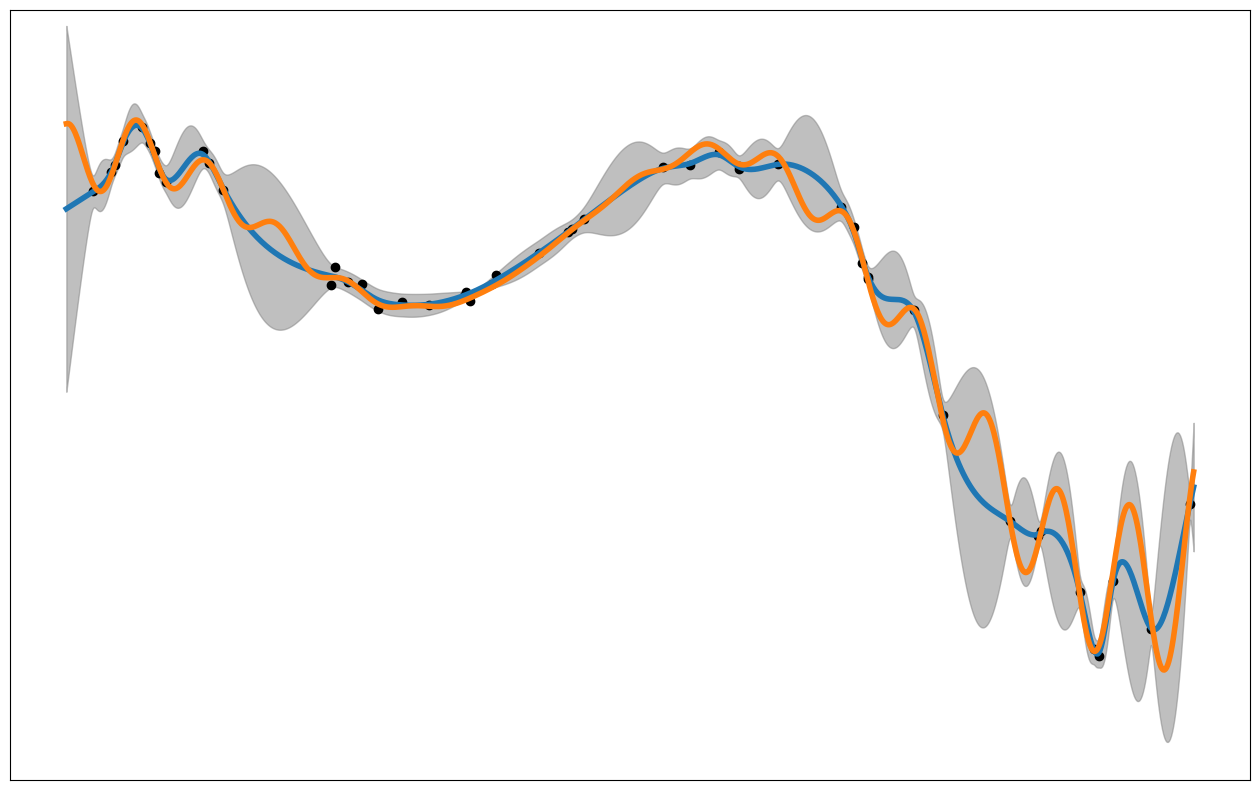}
     \end{subfigure}
     \put(-420,50){$RMSE=0.046$}
     \put(-420,40){$L=72.42$}
     \put(-420,30){$CRPS=-0.022$}
     \put(-420,20){$Time=0.6~s$}
     \put(-180,50){$RMSE=0.038$}
     \put(-180,40){$L=90.10$}
     \put(-180,30){$CRPS=-0.015$}
     \put(-180,20){$Time=15.32~s$}
     \put(-420,110){$k_{stat}$}
     \put(-180,110){$k_{para}$}
     
     \begin{subfigure}[b]{0.45\textwidth}
         \centering
         \includegraphics[width=\textwidth]{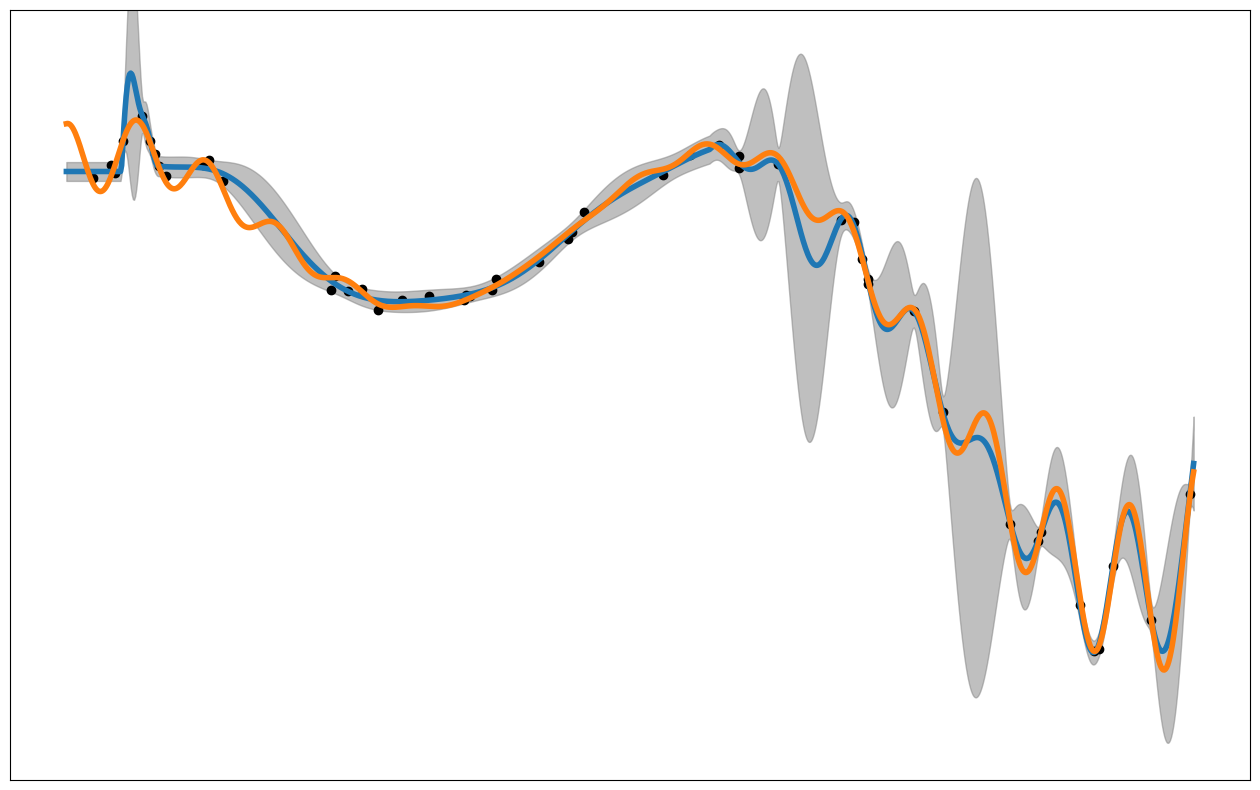}
     \end{subfigure}
     \hfill     
     \begin{subfigure}[b]{0.45\textwidth}
         \centering
         \includegraphics[width=\textwidth]{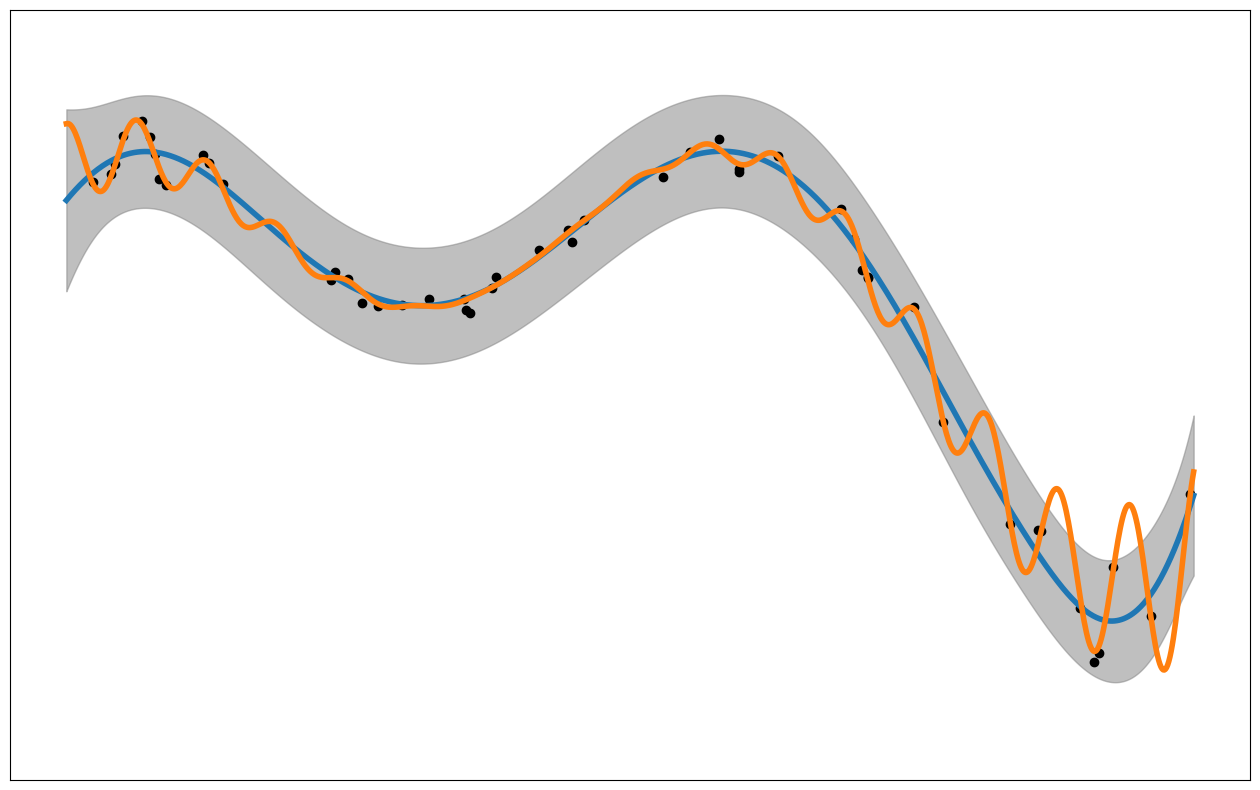}
     \end{subfigure}
     \put(-420,50){$RMSE=0.039$}
     \put(-420,40){$L=85.62$}
     \put(-420,30){$CRPS=-0.016$}
     \put(-420,20){$Time=61.5~s$}
     \put(-180,50){$RMSE=0.053$}
     \put(-180,30){$CRPS=-0.026$}
     \put(-180,20){$Time^*=46~s$}
     \put(-420,110){$k_{nn}$}
     \put(-180,110){deep GP}
     
     \begin{subfigure}[b]{0.8\textwidth}
         \centering
         \includegraphics[width=\textwidth]{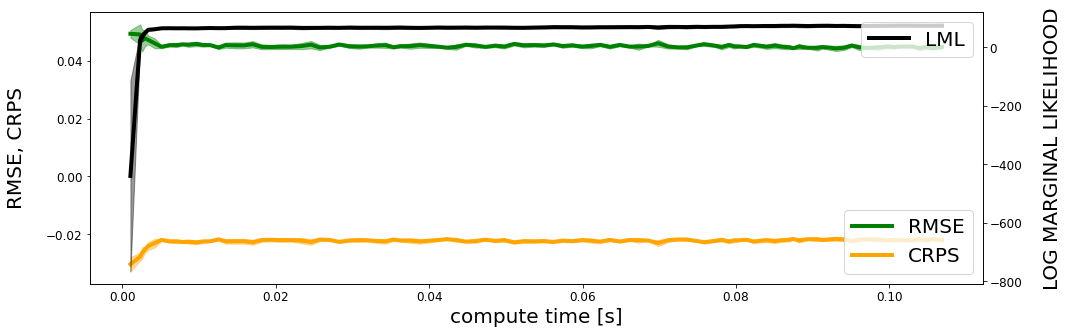}
     \end{subfigure}
     \put(-280,60){$k_{stat}$}

     \begin{subfigure}[b]{0.8\textwidth}
         \centering
         \includegraphics[width=\textwidth]{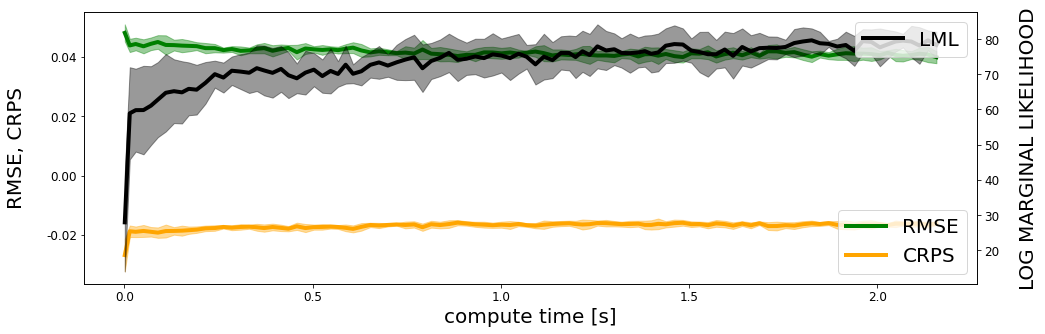}
     \end{subfigure}
     \put(-280,50){$k_{para}$}

    \begin{subfigure}[b]{0.8\textwidth}
         \centering
         \includegraphics[width=\textwidth]{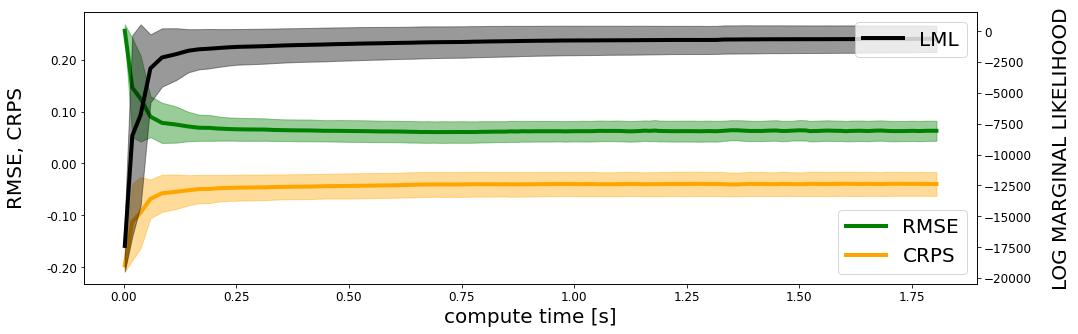}
     \end{subfigure}
     \put(-280,70){$k_{nn}$}
     \caption{Performance overview for dataset 1. From the top-left: stationary reference kernel ($k_{stat}$), parametric non-stationary kernel ($k_{para}$), deep kernel ($k_{nn}$), and deep GP. In this test, the parametric non-stationary kernel and the deep kernel reached similar approximation performances, while the former used significantly fewer hyperparameters, which is why it can be trained significantly faster. The deep GP (see Appendix \ref{app:deepGP} for the algorithm) over-smoothed the model and took a long time to train. We note that the deep-GP result highly depended on the specified nugget --- too small, and the algorithms produced NaNs, just a little larger, and the presented smoothing was observed. The result was the same when we allowed the algorithm to choose its own nugget. Note the asterisk next to the deep-GP compute time, denoting that this is a separate software written in a different language (R), and compute times can therefore not be directly compared. The calculations of the performance measures as a function of MCMC training time were run five times and displayed are the mean and confidence bounds (one standard deviation).}
     \label{fig:1dsynth}
\end{figure}

\begin{figure}
    \centering
    \includegraphics[width = 0.5 \linewidth]{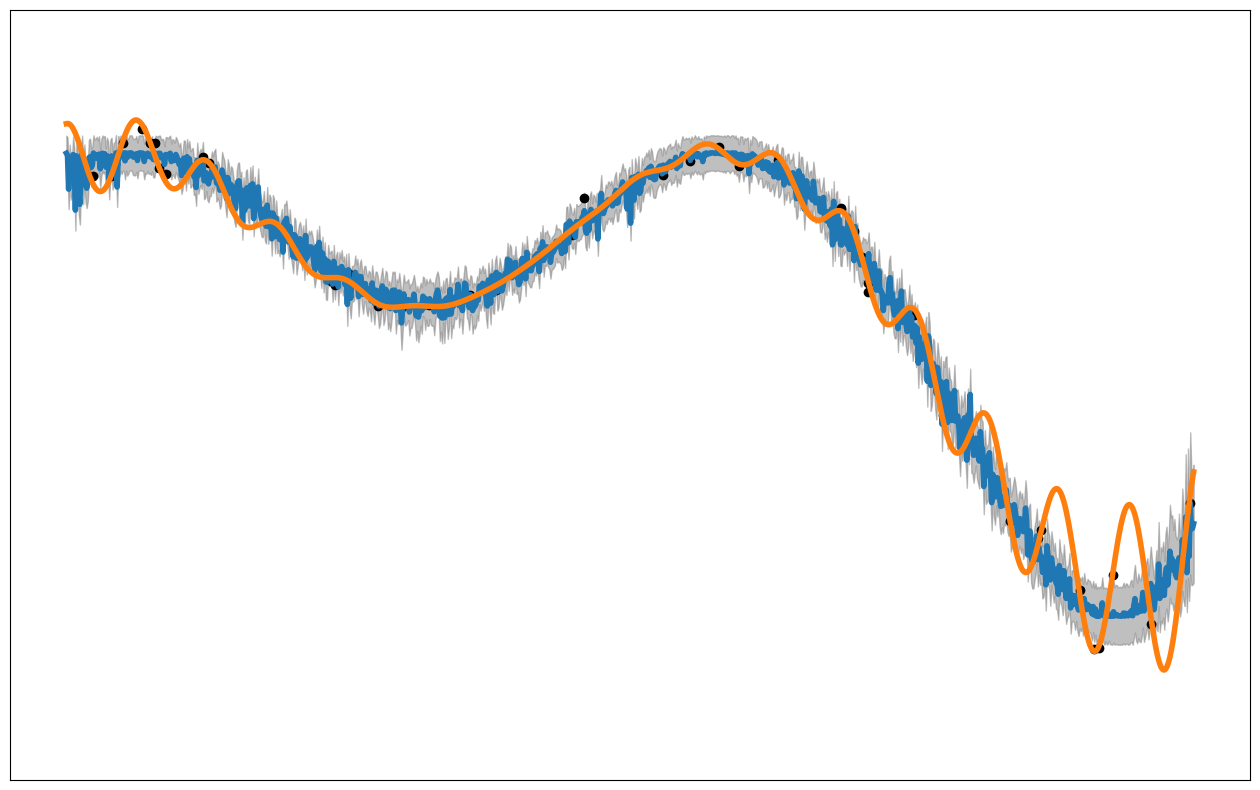}
    \caption{Approximation result for the one-dimensional synthetic dataset using the gpflux DGP. Despite following the online documentation closely, the model appears to show artifacts. This is the reason why we did not use the gpflux DGP for our higher-dimensional test cases. For reproducibility purposes, we publish our exact run script in the Appendix (see \ref{sec:appA}).}
    \label{fig:gpfluxdgp}
\end{figure}

%%%%%%%%%%%%%%%%%%%%%%%%%%%%%%%%%
Our one-dimensional synthetic test function was introduced in Section \ref{sec:datasets}. The stationary reference kernel ($k_{stat}$) is a Mat\'ern kernel with $\nu = 3/2$
\begin{equation}\label{eq:kstat1d}
    k_{stat}(x_i,x_j)=\sigma^2\left(1+\frac{\sqrt{3}d}{l}\right)\exp\left(-\frac{\sqrt{3}d}{l}\right),
\end{equation}
where $l$ is the length scale, and $d = ||x_i - x_j||_2 = |x_i - x_j|$. 
The parametric non-stationary kernel in this experiment was defined as
\begin{equation}\label{eq:kpara1d}
    k_{para}(x_i,x_j) = \big( g_1(x_i)g_1(x_j) + g_2(x_i)g_2(x_j)\big) k_{stat},
\end{equation}
where
\begin{equation}\label{eq:rbf}
    g_a(x) = \sum_{b=1}^6 c^b_a \exp{[-0.5 (||\Tilde{x}_b-x||_2^2)/w_a]},
\end{equation}
$\Tilde{x}=\{0,0.2,0.4,0.6,0.8,1\}$, leading to a total of 15 hyperparameters --- counting two $g_a$ functions in the sum, a constant width of the radial basis functions for each $g_a$, and a constant length scale. The deep kernel is
\begin{equation}\label{eq:knn1d}
    k_{nn}(x_i,x_j)=k_{stat}(\phi(x_i),\phi(x_j)),
\end{equation}
where $\phi$ is a fully connected neural network mapping $\mathbb{R} \rightarrow \mathbb{R}$, with ReLu activation functions and two hidden layers of width five, which led to a total of 48 hyperparameters (weights, biases, and one constant signal variance). The results, presented in Figure \ref{fig:1dsynth}, show a gradual improvement in approximation performance as more flexible kernels are used. The stationary kernel ($k_{stat}$) stands out through its fast computation time. However, the keen observer notices similar uncertainties independent of local properties of the latent function; only point spacing is considered in the uncertainty estimate. That is in stark contrast to the parametric non-stationary kernel ($k_{para}$) and the deep kernel ($k_{nn}$) which both predict lower uncertainties in the well-behaved center region of the domain. This is a very desirable characteristic of non-stationary kernels. The deep kernel and parametric non-stationary kernel reached very similar approximation performance but the deep kernel was by far the most costly to train. The BDGP predicts a very smooth model with subpar accuracy compared to the other methods. We repeated the experiment with different values for the nugget, and let the algorithm choose the nugget, without further success. This is not to say the method cannot perform better, but we remind the reader that we are working under the assumption of reasonable effort, which, in this case, was insufficient to reach a better performance. We share the code with the reader in the Appendix (see \ref{sec:appB}) for reproducibility purposes. We also tested another DGP implementation without success (see Figure \ref{fig:gpfluxdgp}). The MCMC sampling runs revealed what was expected, the stationary kernel converges most robustly; however, all kernels led to a stable convergence within a reasonable compute time.

%%%%%%%%%%%%%%%%%%%%%%%%%%%%%%%%%
\subsubsection{The Climate Model}\label{sec:climate}
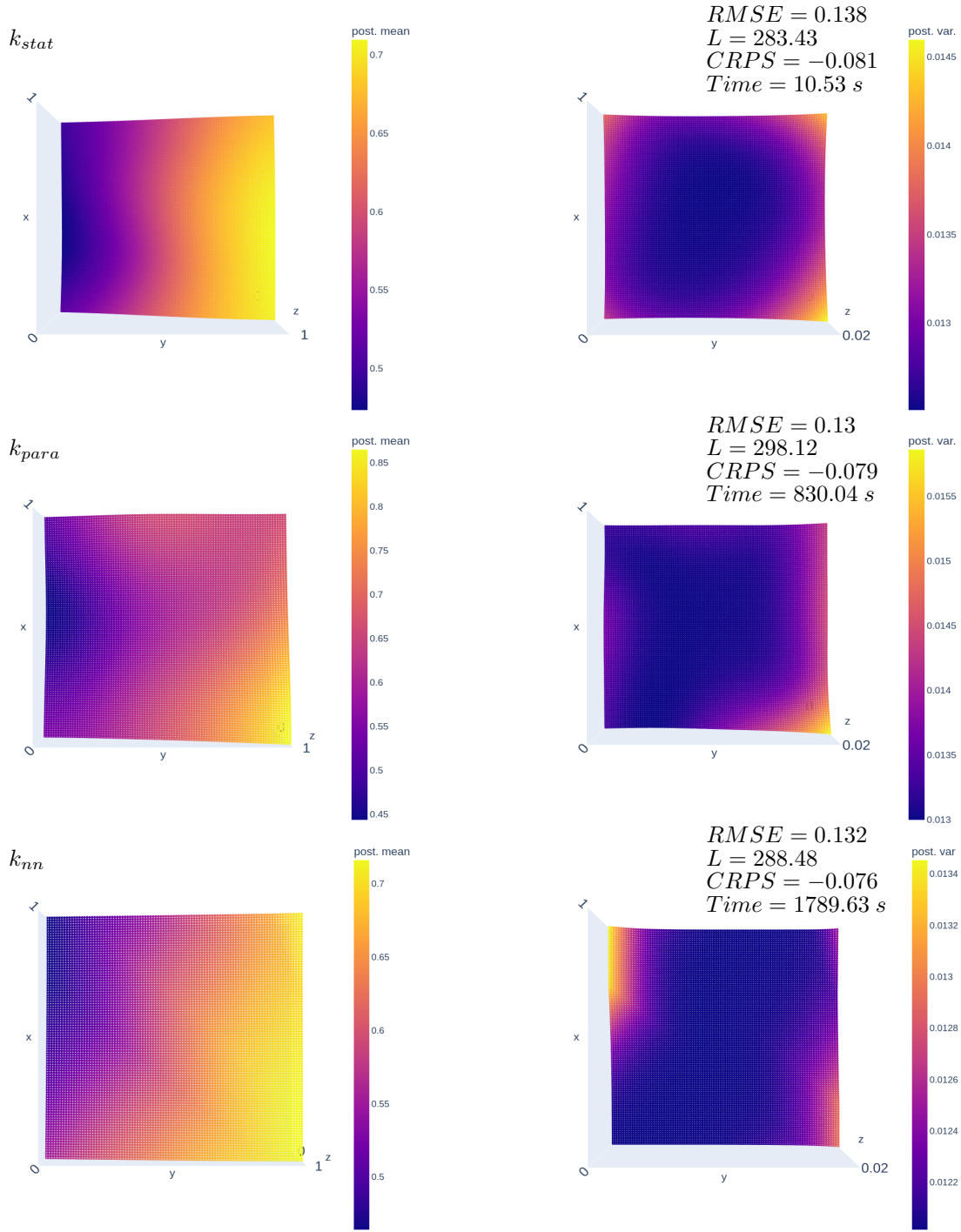
\begin{figure}[H]
    \centering
     \begin{subfigure}[b]{0.45 \textwidth}
         \centering
         \includegraphics[trim={2cm 0 0cm 0cm},clip,width=6cm, height = 6cm]{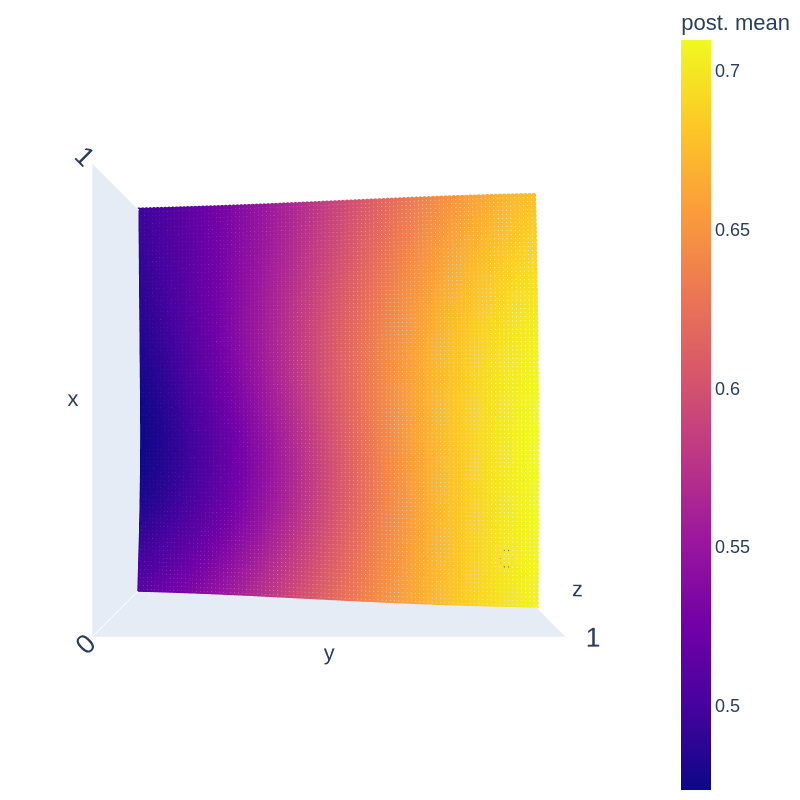}
     \end{subfigure}
     \hfill
     \begin{subfigure}[b]{0.45\textwidth}
         \centering
         \includegraphics[trim={2cm 0 0cm 0cm},clip,width=6cm, height = 6cm]{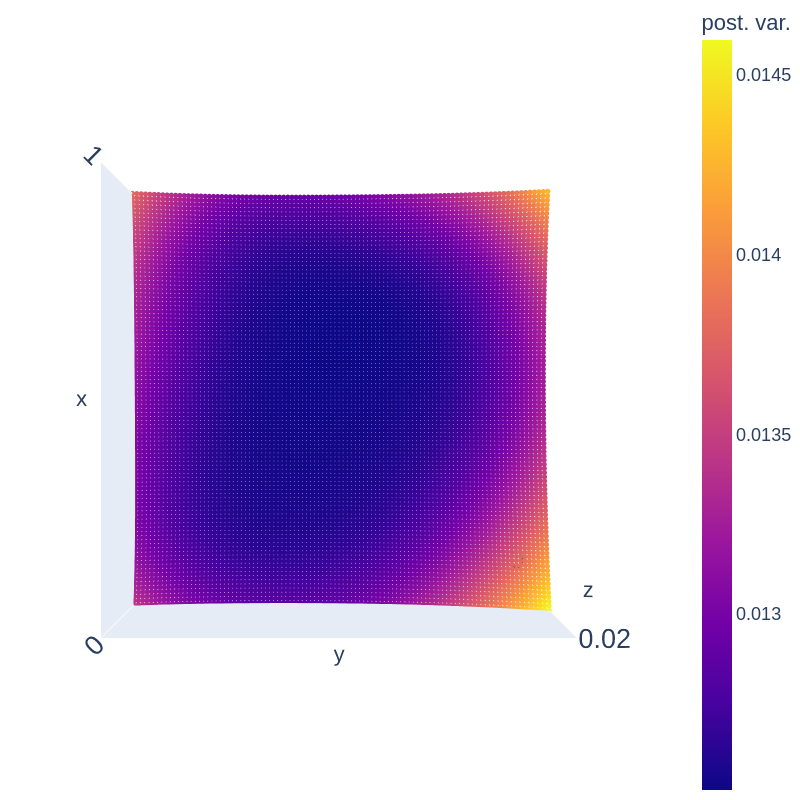}
     \end{subfigure}
     \put(-420,160){$k_{stat}$}
     \put(-120,170){$RMSE=0.138$}
     \put(-120,160){$L=283.43$}
     \put(-120,150){$CRPS=-0.081$}
     \put(-120,140){$Time=10.53~s$}    

     \begin{subfigure}[b]{0.45 \textwidth}
         \centering
         \includegraphics[trim={2cm 0 0cm 0cm},clip,width=6cm, height = 6cm]{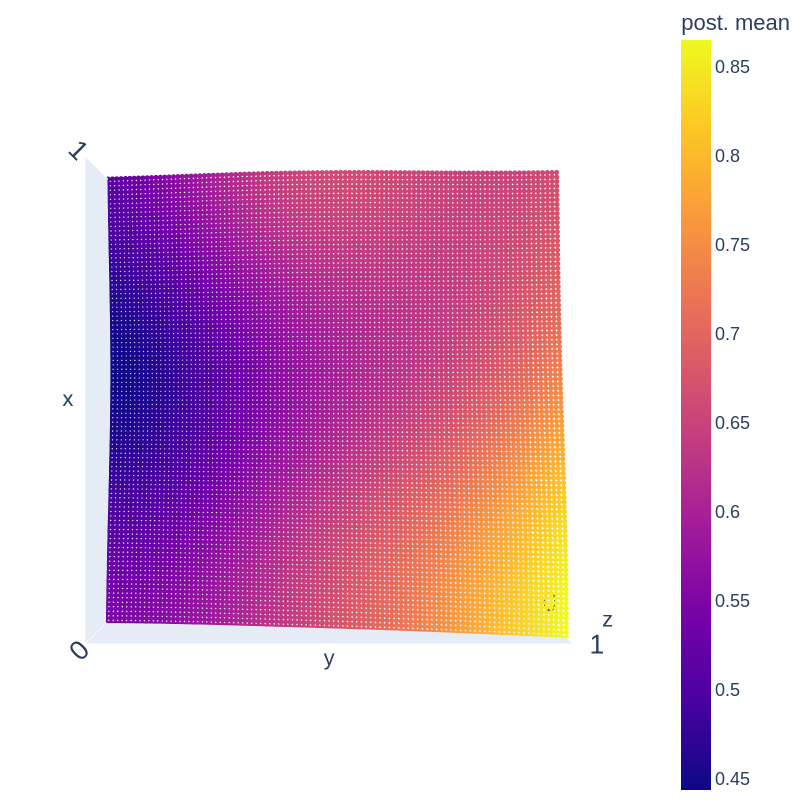}
     \end{subfigure}
     \hfill
     \begin{subfigure}[b]{0.45\textwidth}
         \centering
         \includegraphics[trim={2cm 0 0cm 0cm},clip,width=6cm, height = 6cm]{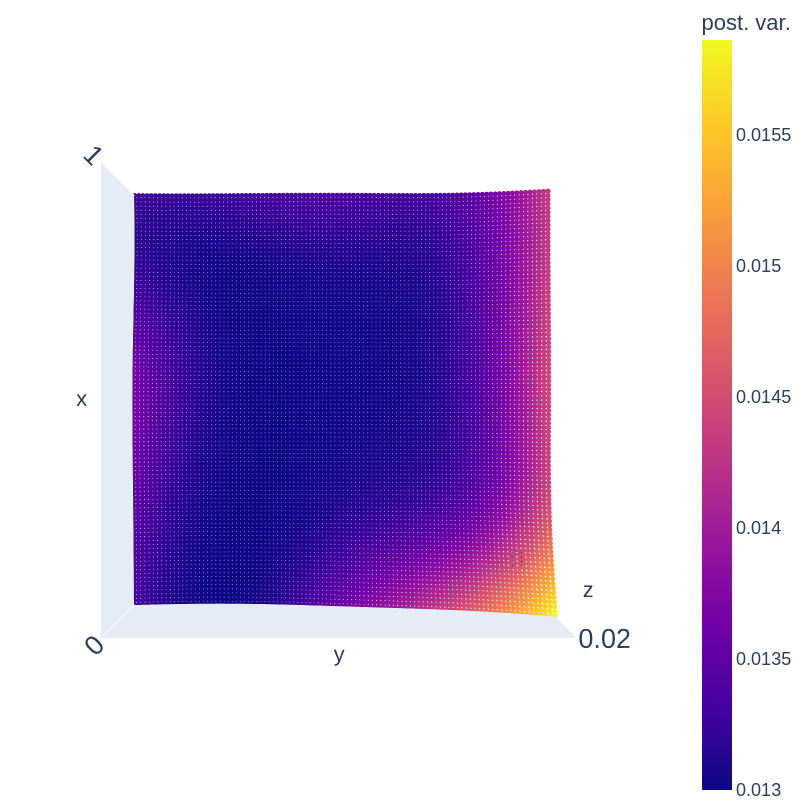}
     \end{subfigure}
     \put(-420,160){$k_{para}$}
     \put(-120,170){$RMSE=0.13$}
     \put(-120,160){$L=298.12$}
     \put(-120,150){$CRPS=-0.079$}
     \put(-120,140){$Time=830.04~s$}

     \begin{subfigure}[b]{0.45 \textwidth}
         \centering
         \includegraphics[trim={1cm 0 0cm 0cm},clip,width=6cm, height = 6cm]{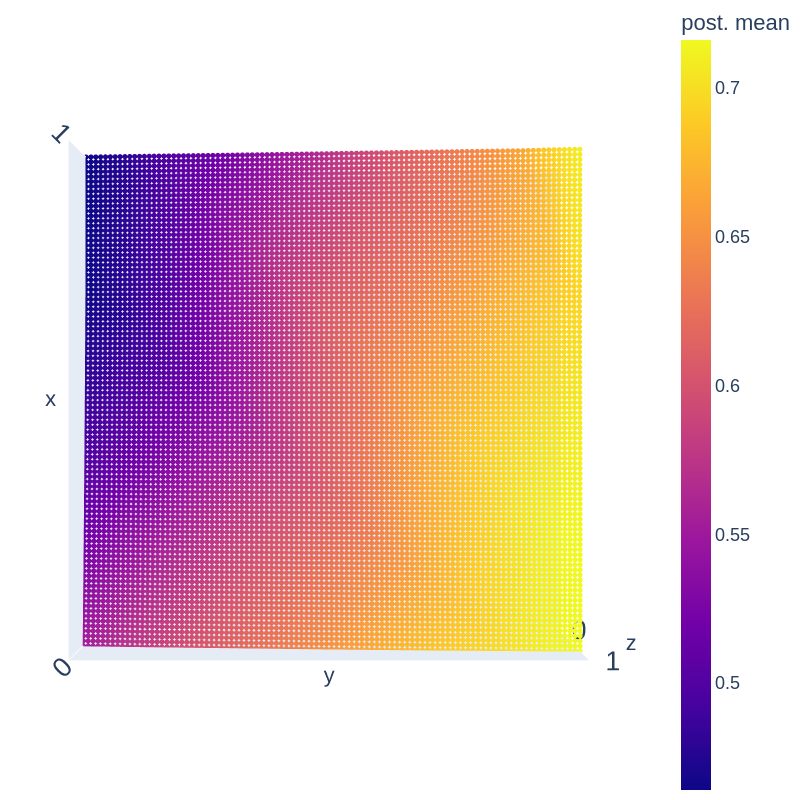}
     \end{subfigure}
     \hfill
     \begin{subfigure}[b]{0.45\textwidth}
         \centering
         \includegraphics[trim={1cm 0 0cm 0cm},clip,width=6cm, height = 6cm]{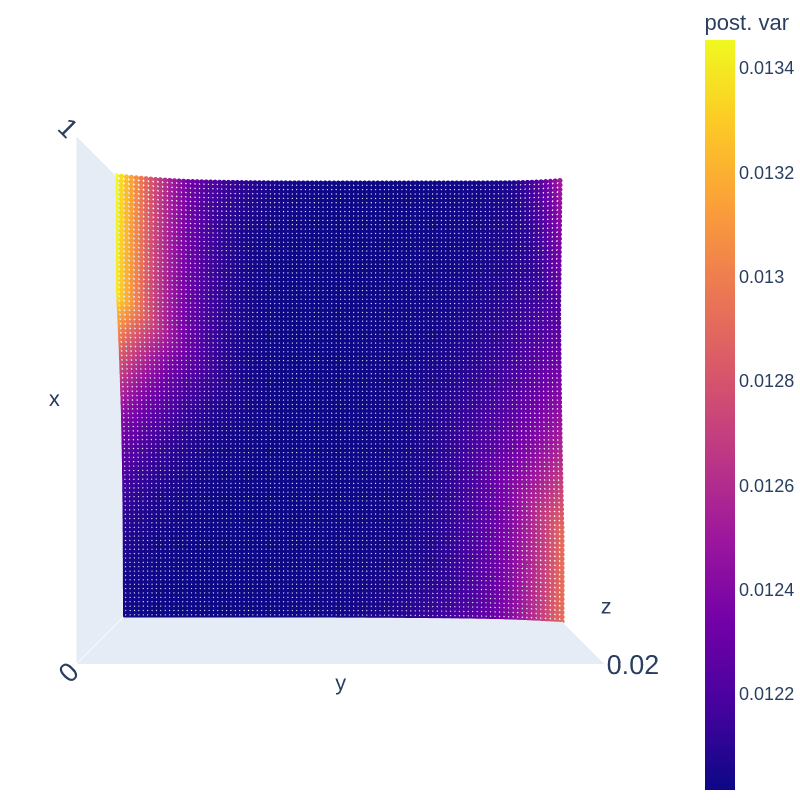}
     \end{subfigure}
     \put(-420,160){$k_{nn}$}
     \put(-120,170){$RMSE=0.132$}
     \put(-120,160){$L=288.48$}
     \put(-120,150){$CRPS=-0.076$}
     \put(-120,140){$Time=1789.63~s$}
     \caption{Performance overview for the climate model. From the top: $k_{stat}$, $k_{para}$, and $k_{nn}$. The posterior mean is displayed in the left column. The posterior variance is on the right. The model function is defined over $[0,1]^2 \times \{0.5\}$. This model is obtained after the convergence of a global evolutionary optimizer. Notable is the fast computing time of the stationary kernel due to the fact that only three hyperparameters have to be found: a signal variance, one isotropic length scale, and the nugget (i.i.d. noise). This parametric non-stationary kernel led to a total number of 58 hyperparameters, which increases the computing time significantly compared to the stationary kernel. The accuracy, in this case, is slightly higher. It is debatable and highly application-driven whether the parametric non-stationary kernel should be used for this dataset. If time is not an issue, the increased accuracy can pay off in downstream operations. The deep kernel contains 186 hyperparameters which are identified robustly. The accuracy of the prediction and the UQ is similar to the parametric non-stationary kernel, with a slight advantage in the CRPS but approximately twice the compute time. As with all our test runs, this run was repeated several times with the same result.}
     \label{fig:3dTempALLKernels}
\end{figure}

\begin{figure}[H]
    \centering
     \begin{subfigure}[b]{0.8 \textwidth}
         \centering
         \includegraphics[width=\textwidth]{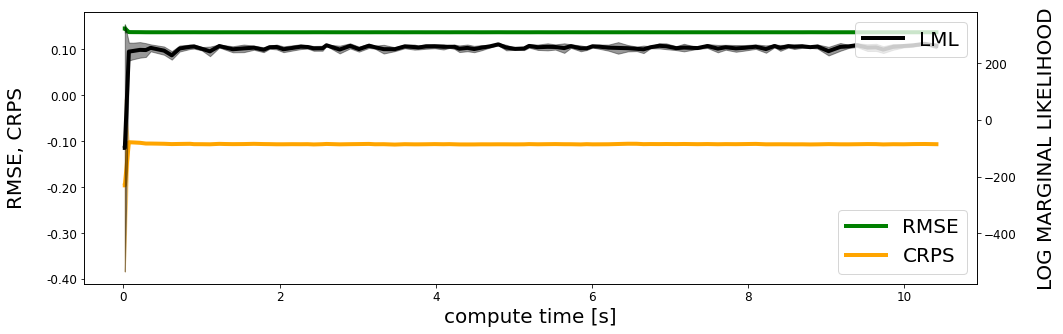}
     \end{subfigure}
     \put(-280,60){$k_{stat}$}

     \begin{subfigure}[b]{0.8 \textwidth}
         \centering
         \includegraphics[width=\textwidth]{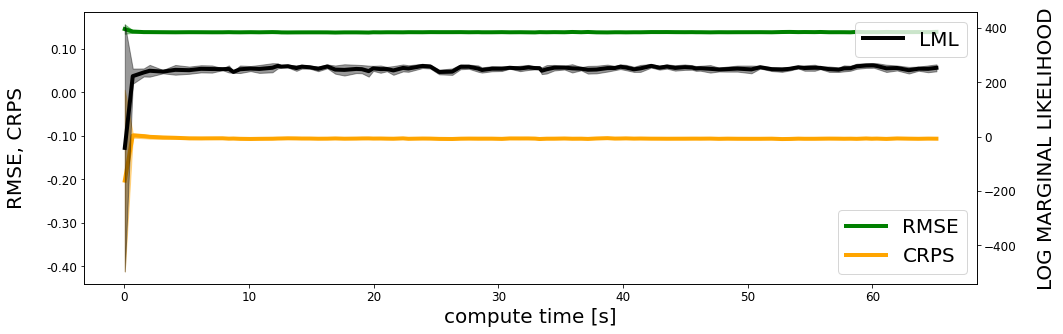}
     \end{subfigure}
     \put(-280,40){$k_{para}$}

     \begin{subfigure}[b]{0.8 \textwidth}
         \centering
         \includegraphics[width=\textwidth]{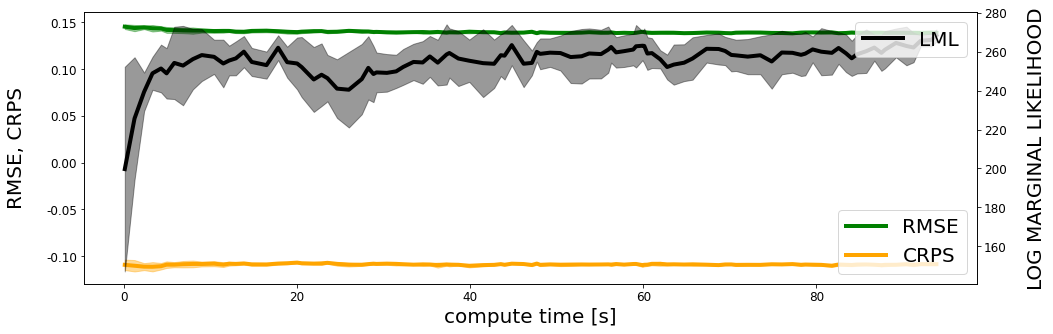}
     \end{subfigure}
     \put(-280,60){$k_{nn}$}
     \caption{MCMC sampling convergence for the climate model and all kernels. From the top: $k_{stat}$, $k_{para}$, and $k_{nn}$. The MCMC is converging robustly for the stationary and parametric non-stationary kernels; the deep kernel is by far the slowest to converge and the error metrics are barely improving. This might be due to the weak non-stationarity in this dataset (see Figure \ref{fig:climateData}). All runs were repeated five times and displayed are the mean and confidence bounds (one standard deviation).}
     \label{fig:3dTempMCMCALLKernels}
\end{figure}

\begin{figure}[H]
    \centering
     \begin{subfigure}[b]{0.45 \textwidth}
         \centering
         \includegraphics[trim={2cm 0 0cm 0cm},clip,width=8cm, height = 8cm]{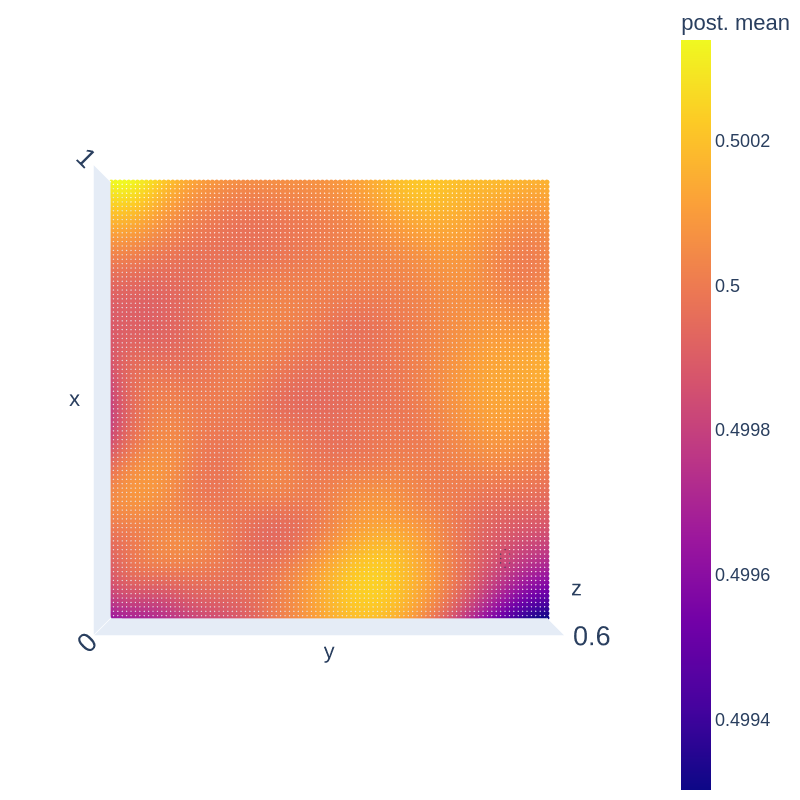}
     \end{subfigure}
     \hfill
     \begin{subfigure}[b]{0.45\textwidth}
         \centering
         \includegraphics[trim={2cm 0 0cm 0cm},clip,width=8cm, height = 8cm]{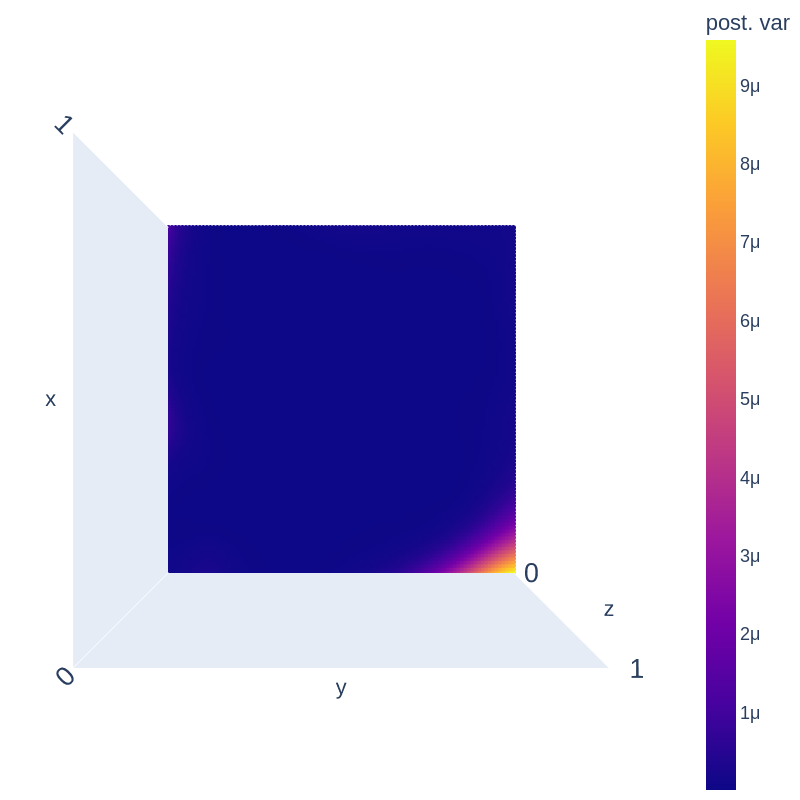}
     \end{subfigure}
     \put(-120,200){$RMSE=0.33$}
     \put(-120,180){$CRPS=-0.27$}
     \put(-120,170){$Time^*=53.7~min$}    
     \caption{Performance overview for the climate model, computed with the BDGP. The model function --- posterior mean (left) and variance (right) --- is defined over $[0,1]^2 \times \{0.5\}$. This result is only presented for completeness. Clearly, our BDGP setup is not competitive compared to the tested kernels. It is presented in Appendix \ref{sec:appB}.}
     \label{fig:3dTempDGP}
\end{figure}

%%%%%%%%%%%%%%%%%%%%%%%%%%%%%%%%%
Our three-dimensional climate dataset was introduced in Section \ref{sec:datasets}.
The stationary reference kernel ($k_{stat}$) is a Mat\'ern kernel with $\nu = 3/2$ (Equation \ref{eq:kstat1d}). 
In all cases, stationary and non-stationary, we added to the kernel matrix the noise covariance matrix $\mathbf{V}=\sigma^2_n \mathbf{I}$, where $\sigma^2_n$ is the nugget variance, treated as an additional hyperparameter. The parametric non-stationary kernel is similar to Equation \eqref{eq:kpara1d}; however, we place radial basis functions \eqref{eq:rbf} at $\{0,0.5,1\}^3$ and added a nugget variance, leading to a total of 58 hyperparameters. The deep kernel $k_{nn}(\boldsymbol{\phi}(\mathbf{x}_i), \boldsymbol{\phi}(\mathbf{x}_j)),~\boldsymbol{\phi}:\mathbb{R}^3 \rightarrow \mathbb{R}^3$ has two hidden layers of width 10, but is otherwise identical to \eqref{eq:knn1d}, yielding 186 hyperparameters.
The results, two-dimensional slices through the three-dimensional input space, are presented in Figure \ref{fig:3dTempALLKernels}. 
For completeness, we included the deep GP result in Figure \ref{fig:3dTempDGP}, which, however, in our run was not competitive. Once again, the stationary kernel ($k_{stat}$) delivers fast and robust results; however, lacks accuracy compared to the parametric non-stationary kernel ($k_{para}$) and the deep kernel ($k_{nn}$). The performance of the two non-stationary kernels is on par with a slight advantage in CRPS for the deep kernel and a significant advantage in compute time for the parametric non-stationary kernel. The MCMC (Figure \ref{fig:3dTempMCMCALLKernels}) sample runs revealed stable convergence, however, at significantly different time scales.

%%%%%%%%%%%%%%%%%%%%%%%%%%%%%%%%%
\subsubsection{The X-Ray Scattering Model}\label{sec:xray}
\begin{figure}[H]
    \centering
     \begin{subfigure}[b]{0.45 \textwidth}
         \centering
         \includegraphics[trim={1.5cm 0 0cm 0cm},clip,width=6cm, height = 6cm]{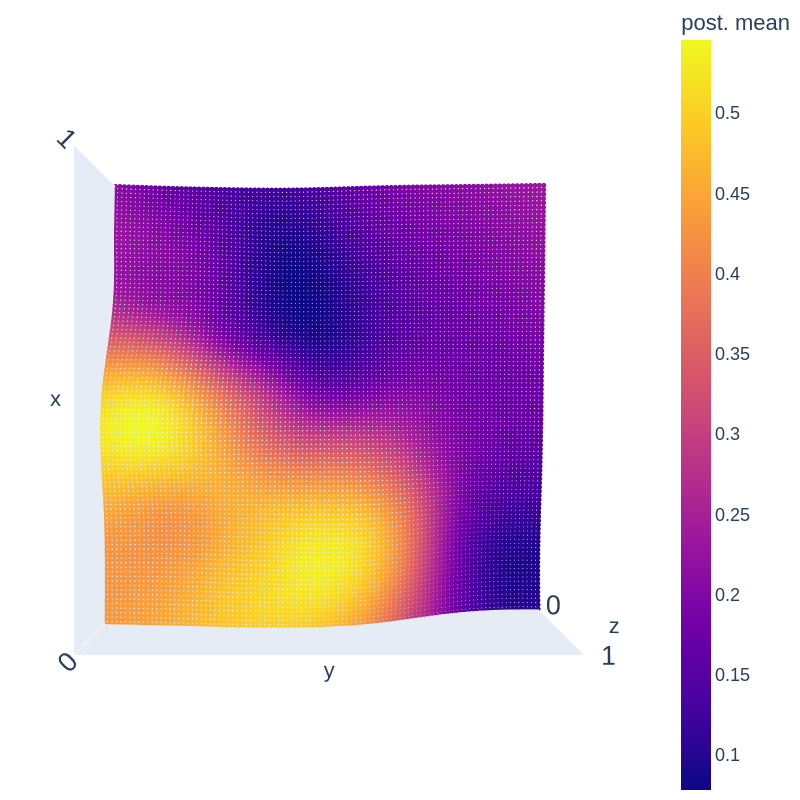}
     \end{subfigure}
     \hfill
     \begin{subfigure}[b]{0.45\textwidth}
         \centering
         \includegraphics[trim={1.5cm 0 0cm 0cm},clip,width=6cm, height = 6cm]{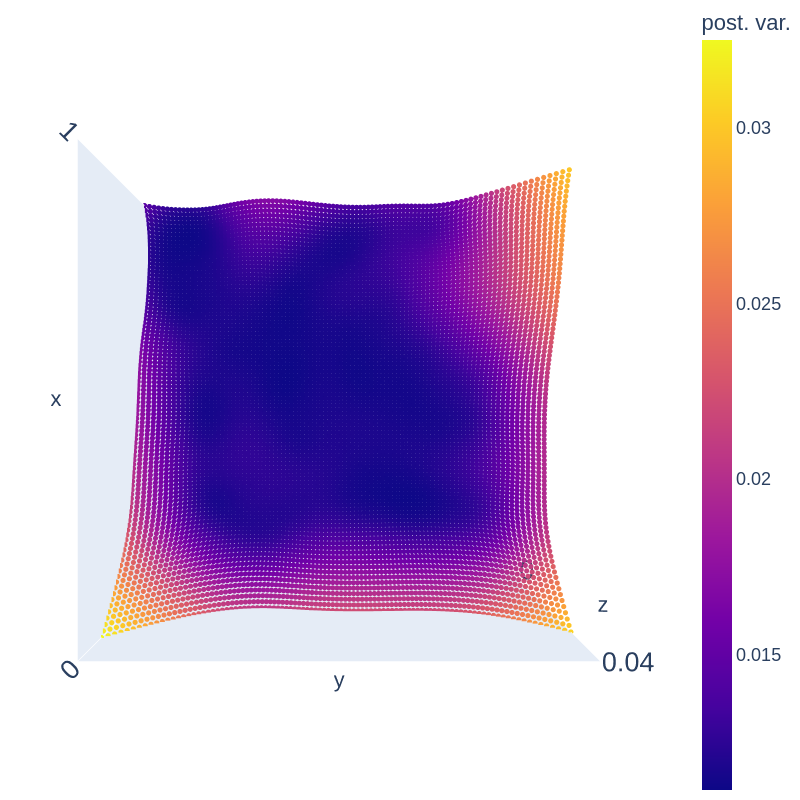}
     \end{subfigure}
     \put(-420,160){$k_{stat}$}
     \put(-120,170){$RMSE=0.096$}
     \put(-120,160){$L=425.17$}
     \put(-120,150){$CRPS=-0.051$}
     \put(-120,140){$Time=18.95~s$}   

     \begin{subfigure}[b]{0.45 \textwidth}
         \centering
         \includegraphics[trim={0cm 0 0cm 0cm},clip,width=6cm, height = 6cm]{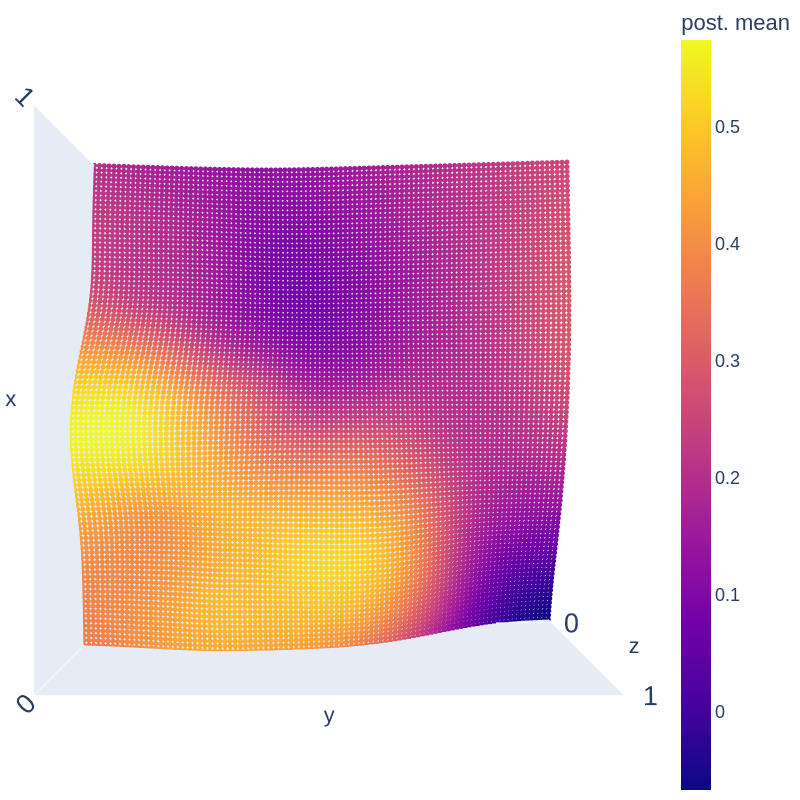}
     \end{subfigure}
     \hfill
     \begin{subfigure}[b]{0.45\textwidth}
         \centering
         \includegraphics[trim={0cm 0 0cm 0cm},clip,width=6cm, height = 6cm]{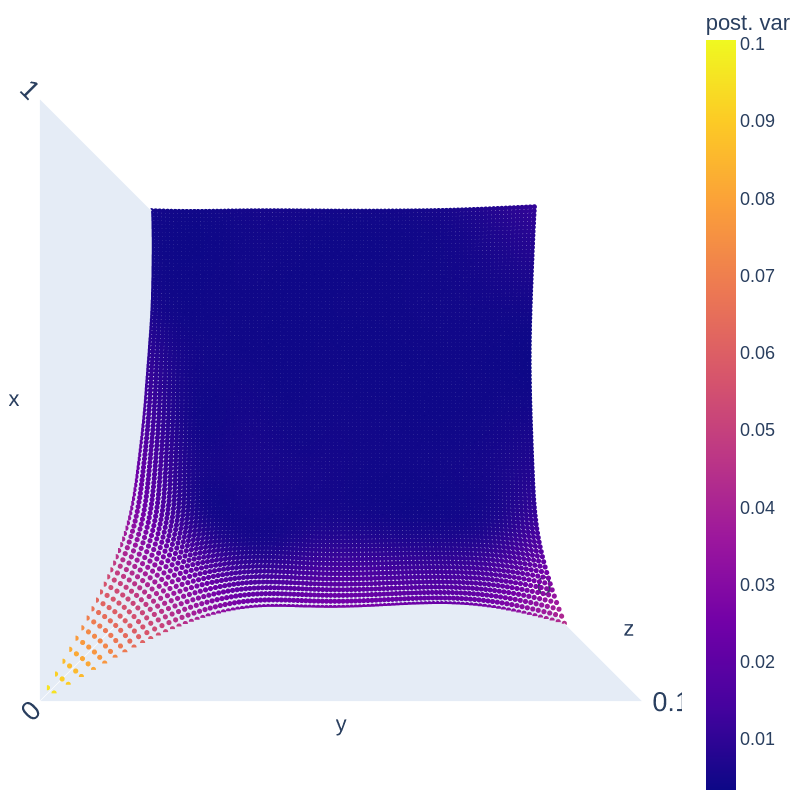}
     \end{subfigure}
     \put(-420,160){$k_{para}$}
     \put(-120,170){$RMSE= 0.097$}
     \put(-120,160){$L=445.05$}
     \put(-120,150){$CRPS=-0.049$}
     \put(-120,140){$Time=815.61~s$}    

     \begin{subfigure}[b]{0.45 \textwidth}
         \centering
         \includegraphics[trim={1.5cm 0 0cm 0cm},clip,width=6cm, height = 6cm]{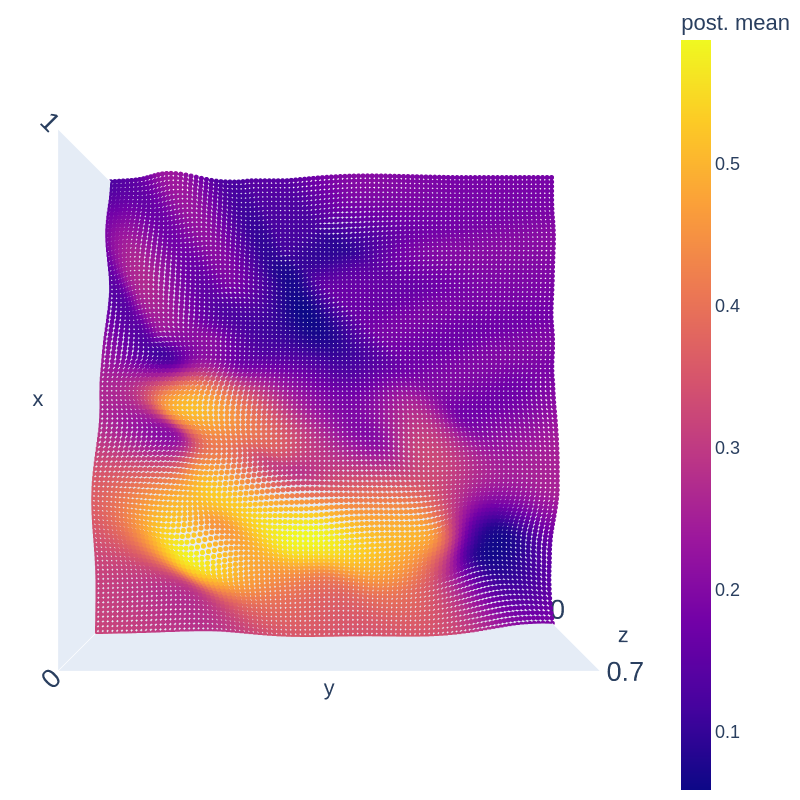}
     \end{subfigure}
     \hfill
     \begin{subfigure}[b]{0.45\textwidth}
         \centering
         \includegraphics[trim={1.5cm 0 0cm 0cm},clip,width=6cm, height = 6cm]{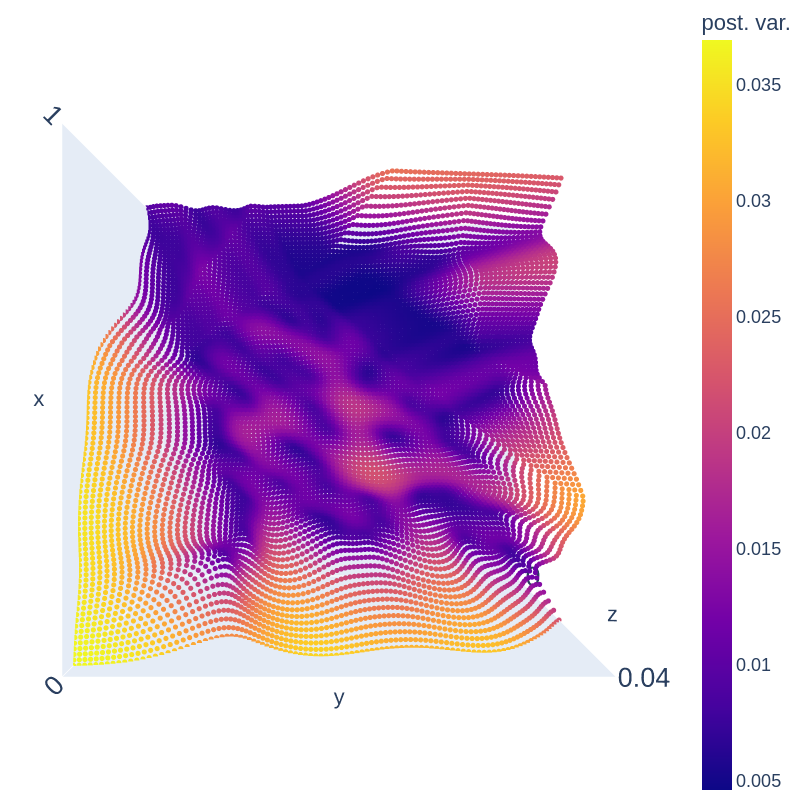}
     \end{subfigure}
     \put(-420,160){$k_{nn}$}
     \put(-120,170){$RMSE= 0.090$}
     \put(-120,160){$L=408.74$}
     \put(-120,150){$CRPS=-0.046$}
     \put(-120,140){$Time=2061.32~s$}    
     \caption{Performance overview for the X-ray model. From the top: $k_{stat}$, $k_{para}$, and $k_{nn}$. The posterior mean is displayed in the left column. The posterior variance is on the right. The model function is defined over $[0,1]^2 \times \{0.24\}$. This model is obtained after the convergence of a global evolutionary optimizer. Once again, the stationary kernel stands out due to fast computing speeds. The compute time increases significantly for the parametric non-stationary kernel, due to the 58 hyperparameters that have to be found. Since the CRPS is our most relevant performance metric, the approximation and the UQ are better than for the stationary kernel. The deep kernel, for this dataset, performs best with respect to the RMSE and CRPS while achieving lower log marginal likelihood. The training time increases again because 186 hyperparameters have to be found.
     }
     \label{fig:3dXrayALLKernels}
\end{figure}
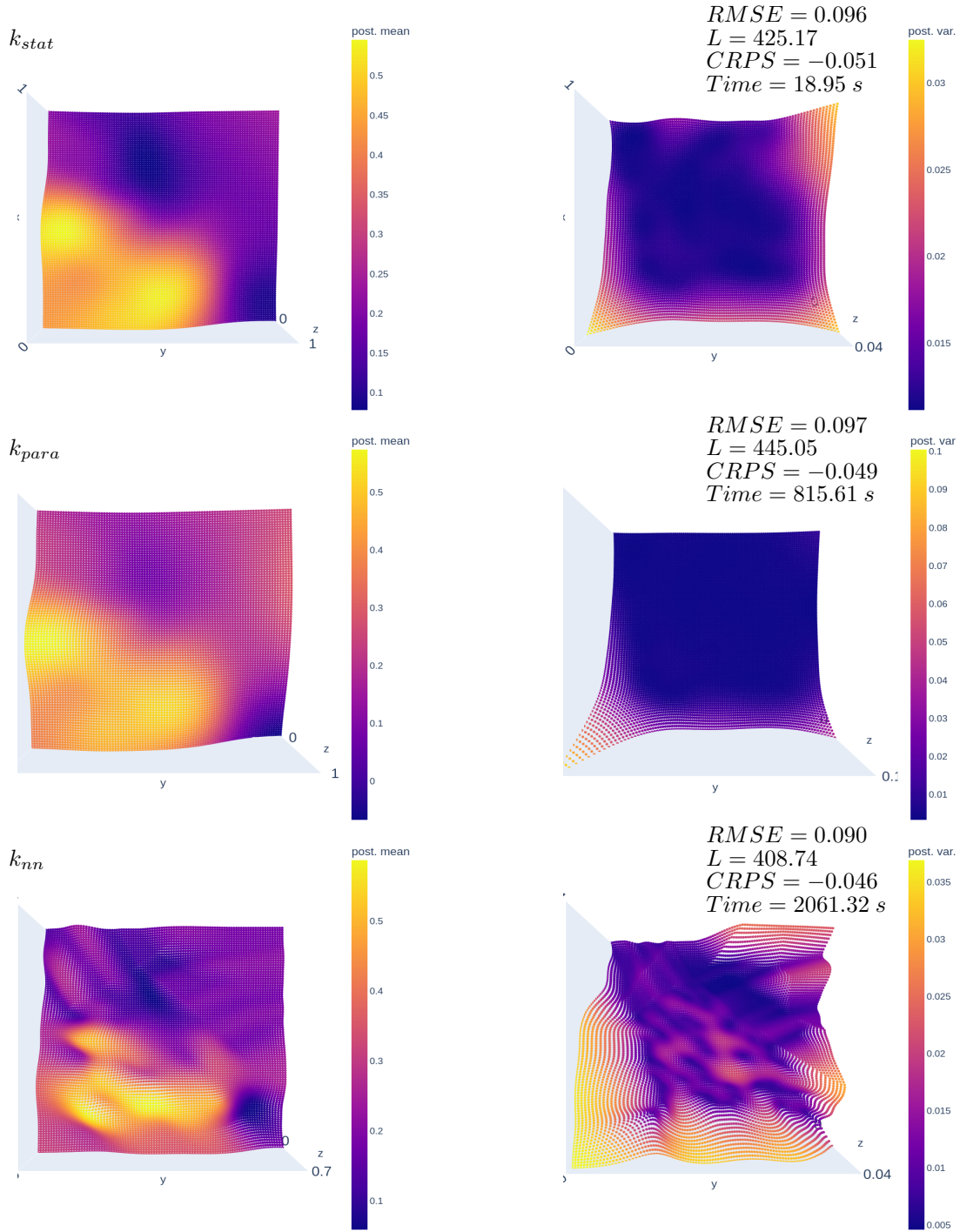

\begin{figure}[H]
    \centering
     \begin{subfigure}[b]{0.8 \textwidth}
         \centering
         \includegraphics[width=\textwidth]{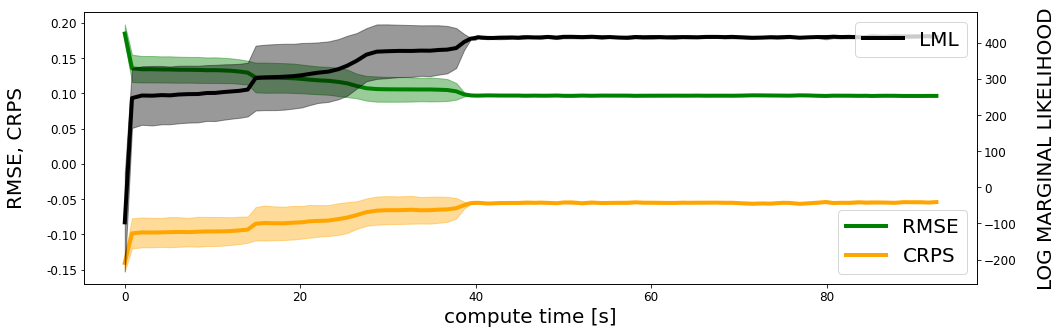}
     \end{subfigure}
     \put(-280,50){$k_{stat}$}

     \begin{subfigure}[b]{0.8 \textwidth}
         \centering
         \includegraphics[width=\textwidth]{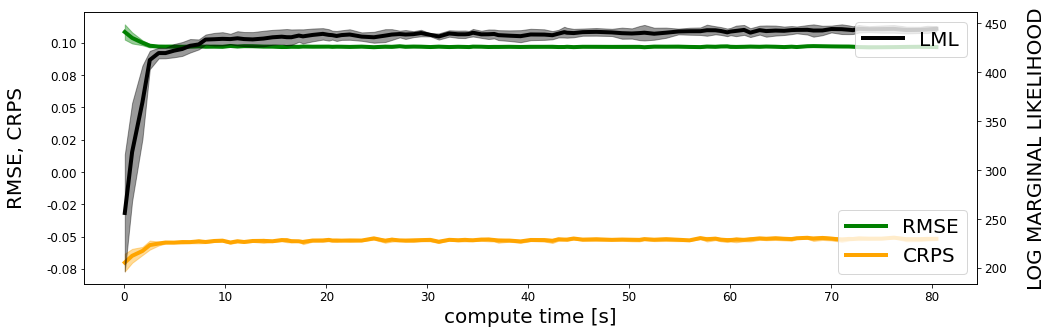}
     \end{subfigure}
     \put(-280,80){$k_{para}$}

     \begin{subfigure}[b]{0.8 \textwidth}
         \centering
         \includegraphics[width=\textwidth]{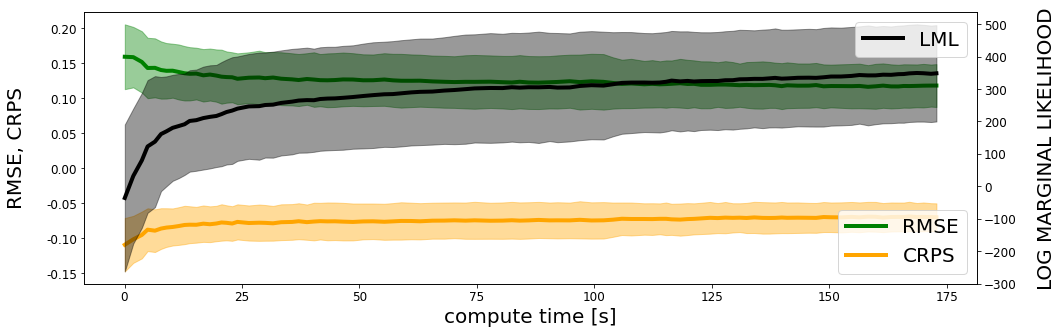}
     \end{subfigure}
     \put(-280,60){$k_{nn}$}
     \caption{MCMC sampling convergence for the X-ray model and all kernels. From the top: $k_{stat}$, $k_{para}$, and $k_{nn}$. The MCMC is converging robustly and, in this case, at similar time scales, making the case for the deep kernel that reached the best CRPS (in the optimization). All runs were repeated five times and displayed are the mean and confidence bounds (one standard deviation).}
     \label{fig:3dXrayMCMCALLKernels}
\end{figure}

\begin{figure}[H]
    \centering
     \begin{subfigure}[b]{0.45 \textwidth}
         \centering
         \includegraphics[trim={2cm 0 0cm 0cm},clip,width=8cm, height = 8cm]{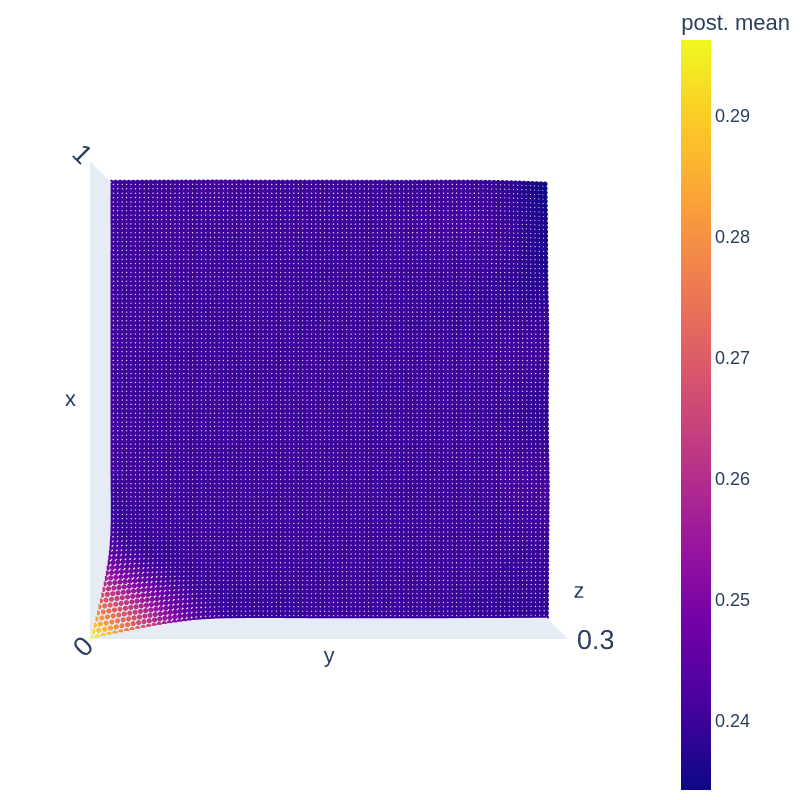}
     \end{subfigure}
     \hfill
     \begin{subfigure}[b]{0.45\textwidth}
         \centering
         \includegraphics[trim={2cm 0 0cm 0cm},clip,width=8cm, height = 8cm]{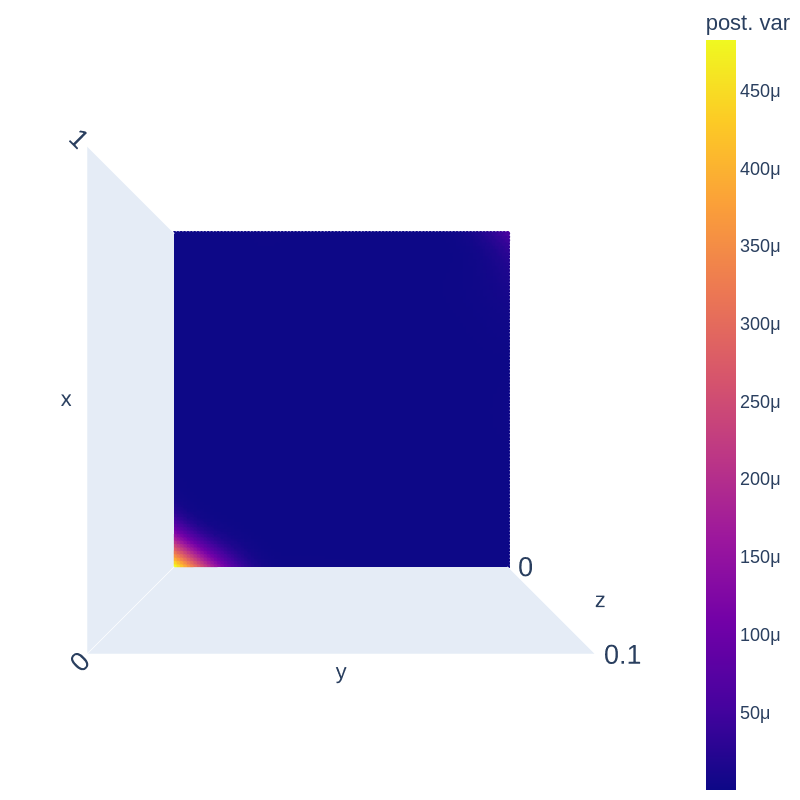}
     \end{subfigure}
     \put(-120,210){$RMSE= 0.44$}
     \put(-120,190){$CRPS=-0.35$}
     \put(-120,180){$Time^*=61.75~min$}  
     \caption{Performance overview for the X-ray scattering model, computed with the BDGP. The model function --- posterior mean (left) and variance (right) --- is defined over $[0,1]^2 \times \{0.24\}$. This result is, again, included for completeness; it is not competitive compared to the tested kernels. Our script is presented in Appendix \ref{sec:appB}.}
     \label{fig:2dXrayDGP}
\end{figure}

%%%%%%%%%%%%%%%%%%%%%%%%%%%%%%%%%
Our three-dimensional X-ray scattering dataset was introduced in Section \ref{sec:datasets}. The kernel setup is identical to our climate example (see Section \ref{sec:climate}). The results are presented in Figure \ref{fig:3dXrayALLKernels}. This experiment shows improving performance as kernel complexity increases, however, the improvements are moderate. As before, if accuracy is the priority, non-stationary kernels should be considered. Again, the deep kernel ($k_{nn}$) performed very competitively --- with a significant advantage in terms of RMSE and CRPS --- but did not reach the same log marginal likelihood as the competitors, which can be traced back to solving a high-dimensional optimization problem due to a large number of hyperparameters. This opens the door to an even better performance if more effort and time is spent in training the model. What stands out in Figure \ref{fig:3dXrayALLKernels} is the smaller and more detailed predicted uncertainties for the deep kernel which would affect decision-making in an online data acquisition context. Also, the posterior mean has more intricate details compared to the stationary and the parametric non-stationary kernel. Figure \ref{fig:3dXrayMCMCALLKernels} reveals stable convergence of the MCMC sampling for all kernels at similar time scales supporting the deep kernel as the superior choice for this model. We again included the BDGP, which however was not competitive (Figure \ref{fig:2dXrayDGP}).

%%%%%%%%%%%%%%%%%%%%%%%%%%%%%%
%%%%%%%%%%%%%%%%%%%%%%%%%%%%%%
%%%%%%%%%%%%%%%%%%%%%%%%%%%%%%
%%%%%%%%%%%%%%%%%%%%%%%%%%%%%%
\section{Discussion}
%%%%%%%%%%%%%%%%%%%%%%%%%%%%%%
%%%%%%%%%%%%%%%%%%%%%%%%%%%%%%
%%%%%%%%%%%%%%%%%%%%%%%%%%%%%%
%%%%%%%%%%%%%%%%%%%%%%%%%%%%%%
\subsection{Interpretations Test-by-Test}
The one-dimensional synthetic test function \eqref{eq:synthfunc} exhibits a complex behavior that is captured by our stationarity analysis through a broad distribution in the length scale and signal variance. The scattering patterns reveal a multifaceted behavior, with different regions of the data exhibiting distinct statistical properties. The linear correlation within one cluster (left, scatter plot, Figure \ref{fig:syntheticData}) and the dispersion in the other cluster captures the interplay between the sinusoidal and quadratic components of the function. The linear cluster indicates a linear correlation between these two hyperparameters within a specific region of input data. This pattern may correspond to the sinusoidal components of the function, where local oscillations exhibit a consistent relationship between the length scale and signal variance. The additional dispersed points in the scatter plot likely correspond to the regions influenced by the quadratic term in the function, where the statistical properties vary, leading to a broader distribution of the hyperparameters. The bimodal distributions observed in the violin plot for the length scale further corroborate the non-stationarity. The results provide insights into the intricate interplay between the sinusoidal and quadratic components of the function. Due to the strong non-stationarity in the data, non-stationary kernels performed well for this test function. Due to the low dimensionality, the number of hyperparameters is low in all cases, leading to robust training. It is in our opinion safe to conclude, that in one-dimensional cases, with moderate dataset sizes, and suspected non-stationarity in the data, non-stationary kernels are to be preferred. Both our parametric and deep non-stationary kernels performed well with a slight edge in accuracy for the parametric kernel (see Figure \ref{fig:1dsynth}). Our two tested deep GP setups (Figures \ref{fig:1dsynth} and \ref{fig:gpfluxdgp}) were not competitive given our reasonable-constraint effort, which in this case, was in the order of days.

\vspace{2mm}
\noindent
Moving on to the climate data example, Figure \ref{fig:climateData} shows a quite concentrated cluster near $(0,0)$ 
indicating weak non-stationarity.
This concentration suggests that the statistical properties are consistent across this region, possibly reflecting a dominant pattern or behavior in the data. The computational experiments (see Figures \ref{fig:3dXrayALLKernels}) reveal a trade-off between accuracy and compute time. One has to put much more effort --- number of hyperparameters and time --- into the computation for a relatively small gain in accuracy. Both the parametric non-stationary kernel and the deep kernel achieve higher accuracy than the stationary kernel but are costly. In time-sensitive situations, the stationary kernel is likely to be preferred in this situation; for best accuracy, the parametric non-stationary kernel or the deep kernel is the superior choice. Looking at the CRPS, the deep kernel has a slight edge in accurately estimating uncertainties over the parametric kernel; however, the parametric non-stationary kernel reaches the highest log marginal likelihood. For this dataset, we tested the BDGP without much success under our reasonable-effort constraint (see Figure \ref{fig:3dTempDGP}).

\vspace{2mm}
\noindent
Finally, for the X-ray scattering data, Figure \ref{fig:xrayData} shows a high level of dispersion in the scatter and violin plots, indicating strong non-stationarity, which is further corroborated in the violin plots. This strong non-stationarity --- particularly in the length scale --- clearly benefits the performance of the deep kernel (see Figure \ref{fig:3dXrayALLKernels}). The stationary kernel shows quite poor performance both in terms of RMSE and CRPS --- since stationary kernels only allow us to estimate uncertainties based on global properties of the data and local geometry it is expected that non-stationarity kernels estimate uncertainties more accurately which manifests itself in a higher CRPS. The parametric non-stationary kernel leads the field in log marginal likelihood. Surprisingly, among the three tested kernels, the deep kernel leads to by far the lowest log marginal likelihood, which, again, suggests a better optimizer might lead to a much-improved performance. Once again, our setup of the BDGP was not competitive (see Figure \ref{fig:2dXrayDGP}).

\subsection{Key Takeaways from the Computational Experiments}
While we included as much information in the computational experiments and the Appendix as possible to give the reader a chance to make up their own minds, here we summarize some key takeaways. 
\begin{enumerate}
    \item Stationary kernels proved to be surprisingly competitive in terms of the RMSE and are unbeatable when evaluating accuracy per time. It seems worth it in most cases to run a stationary GP first for comparison.
    \item Uncertainty is estimated more accurately by non-stationary kernels; if UQ-driven decision-making is the goal, non-stationary kernels are to be preferred. This is not a surprise since, given a constant (possibly anisotropic) length scale and a signal variance, the posterior variance only depends on data-point geometry.
    \item The parametric non-stationary kernel encodes a flexible non-stationarity while maintaining interpretability. The involved parametric functions over the input space can be visualized and interpreted. 
    \item Deep kernels are some of the most flexible kernels but interpretation is lost in all but the simplest cases, which can easily lead to model misspecifications (wrong model class and hyperparameters). Our models took a fair amount of experimenting before an acceptable performance was achieved. In online applications, in which testing beforehand is not possible, deep kernels should be used with caution.
    \item Non-stationarity in the covariance structure appears in signal variance and length scale; ideally a kernel addresses both (see Equation \ref{eq:newkernel}).
    \item While not included in the tests, experimenting with prior mean functions has shown that non-stationarity properties highly depend on the prior mean function of the GP. This is especially true for the non-stationary signal variance.
    \item Extrapolation and non-stationary kernels are difficult to combine. While the parametric non-stationary kernel can be set up for extrapolation, traditional neural networks are poorly equipped for that task.
    \item We should think of the number of hyperparameters conservatively; too many bear the risk of model misspecification (through local minima) and overfitting.
    \item The parametric non-stationary kernel achieved overall better RMSE; the deep kernel led to better uncertainty quantification as indicated by the CRPS.   
\end{enumerate}

\subsection{A Parametric Deep Kernel}
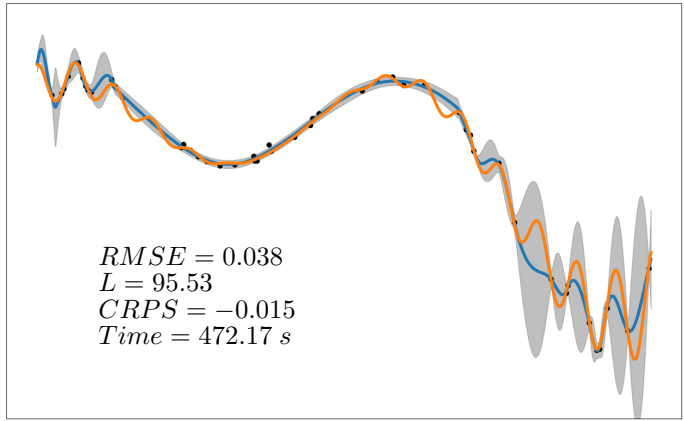
\begin{figure}[H]
    \centering
    \includegraphics[width = 0.6 \linewidth]{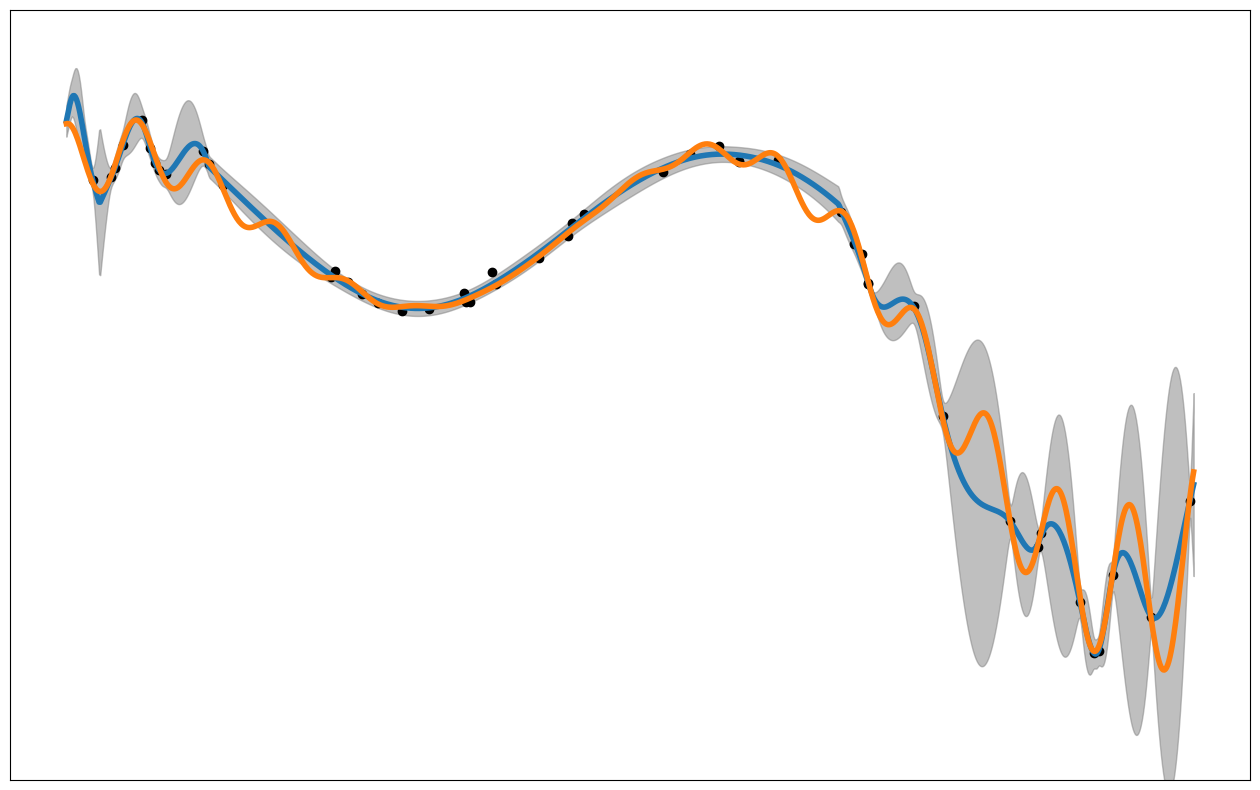}
     \put(-220,60){$RMSE=0.038$}
     \put(-220,50){$L=95.53$}
     \put(-220,40){$CRPS=-0.015$}
     \put(-220,30){$Time=472.17~s$}
    \caption{Approximation result for the one-dimensional synthetic dataset using the parametric deep non-stationary kernel \eqref{eq:newkernel}. The kernel combines non-stationarity in the signal variance and the length scale by leveraging both the parametric non-stationarity kernel design and a deep kernel. The approximation is on par with our previously tested methods but the reached likelihood is higher. This kernel might perform very well when strong non-stationarity is present in length scale and signal variance}
    \label{fig:newkernel}
\end{figure}

Given the observation that non-stationarity in the covariance structure of a dataset originates from a non-constant signal variance and length scales, one might argue that both should be addressed in the kernel design. Our parametric non-stationary kernel attempts to account for all non-stationary purely through the signal variance; it leaves the length scale constant --- however, implementations exist that allow non-constant and anisotropic length scales \cite{paciorek2006spatial, risser2015regression}, generally in the same flavor as the parametric non-stationary signal variance. The deep kernel, on the other hand, only acts on the length scale by warping the input space. It seems logical to ask what happens if we combine the two concepts.
The kernel
\begin{equation}\label{eq:newkernel}
    k(\mathbf{x}_i,\mathbf{x}_j)=\sum_{d=1}^2 g_d(\mathbf{x}_i) g_d(\mathbf{x}_j) k(\boldsymbol{\phi}(\mathbf{x}_i),\boldsymbol{\phi}(\mathbf{x}_j))
\end{equation}
achieves just that; it is a combination of our parametric non-stationary kernel and the deep kernel. Modeling our synthetic dataset, see Figure \ref{fig:newkernel}, we see a good performance with moderate improvements compared to our earlier tests (Figure \ref{fig:1dsynth}). The kernel might constitute somewhat of an overkill for such a simple problem but might lead to more significant gains in real applications. We encourage the reader to give this kernel a try in their next application.
%%%%%%%%%%%%%%%%%%%%%%%%%%%%%%
%%%%%%%%%%%%%%%%%%%%%%%%%%%%%%
%%%%%%%%%%%%%%%%%%%%%%%%%%%%%%
%%%%%%%%%%%%%%%%%%%%%%%%%%%%%%
\subsection{Connection between Multi-Task Learning and Non-Stationary Kernels}
%%%%%%%%%%%%%%%%%%%%%%%%%%%%%%
%%%%%%%%%%%%%%%%%%%%%%%%%%%%%%
%%%%%%%%%%%%%%%%%%%%%%%%%%%%%%
%%%%%%%%%%%%%%%%%%%%%%%%%%%%%%
Multi-task learning offers a powerful paradigm to leverage shared information across multiple related tasks, thereby enhancing the predictive performance of each individual task \cite{hrluo_2022a,gptune_web,zhu2022gptuneband}. This is particularly beneficial when data for some tasks are sparse, as information from data-rich tasks can be used to inform predictions for data-scarce tasks. Flexible non-stationary kernels offer an interesting benefit for multi-task learning: instead of employing specialized techniques, such as coregionalization, one can reformulate the problem to a single task problem and let a flexible non-stationary kernel learn intricate correlations between input ($\mathcal{X}_i$) and output ($\mathcal{X}_o$) space locations. By transforming the multi-task learning problem over $\mathcal{X}_i$ to a single-task learning problem over $\mathcal{X}_i \times \mathcal{X}_o$, no further changes to the core algorithm are required. This has been known for a long time and is referred to as problem-transformation methods \cite{borchani2015survey}. These methods were originally dismissed as not being able to capture intricate correlations between the tasks; however, this is only true if stationary, separable kernels are used. A flexible non-stationary kernel is able to flexibly encode covariances across the input and the output domain, independent of the indexing of the tasks. This makes it possible to transfer all complexities of multi-task learning to the kernel design and use the rest of the GP framework as-is, inheriting the superior robustness and interpretability properties of single-task GPs.

%%%%%%%%%%%%%%%%%%%%%%%%%%%%%%
%%%%%%%%%%%%%%%%%%%%%%%%%%%%%%
%%%%%%%%%%%%%%%%%%%%%%%%%%%%%%
%%%%%%%%%%%%%%%%%%%%%%%%%%%%%%
\section{Summary and Conclusion}
%%%%%%%%%%%%%%%%%%%%%%%%%%%%%%
%%%%%%%%%%%%%%%%%%%%%%%%%%%%%%
%%%%%%%%%%%%%%%%%%%%%%%%%%%%%%
%%%%%%%%%%%%%%%%%%%%%%%%%%%%%%
In this paper, we put on the hat of a machine learning practitioner trying to find the best kernel or methodology within the scope of a Gaussian process (GP) to address non-stationarity in various datasets. We introduced three different datasets --- one synthetic, one climate dataset, and one originating from an X-ray scatting experiment at the CMS beamline at NSLSII, Brookhaven National Laboratory. We introduced a non-stationarity measure and studied each dataset to be able to judge their non-stationarity properties quantitatively. We then presented four different methodologies to address the non-stationarity: Ignoring it by using a stationary kernel, a parametric non-stationary kernel, that uses a flexible non-constant signal variance, a deep kernel that uses a neural network to warp the input space, and a deep GP. We set all methodologies up under reasonable effort constraints to allow for a fair comparison; just like a a practitioner might encounter. In our case, that meant minutes of setup time for the stationary kernels and hours to days for the non-stationary kernels and the deep GPs. After the methodologies were set up, we ran our computational tests and presented the results unredacted and uncensored. This way, we hope, the reader gets the best value out of the comparisons. This is also to ensure that the reader has the chance to come up with conclusions different from ours. To further the readers' ability to double-check and learn, we are publishing all our codes online.

\vspace{2mm}
\noindent
Our tests have shown that even weak non-stationarity in a dataset motivates the use of non-stationary kernels if training time is not an issue of concern. If training time is very limited, stationary kernels are still the preferred choice. We have discovered that non-stationarity in the covariance comes in two flavors, in the signal variance and the length scale and, ideally, both should be addressed through novel kernel designs; we drew attention to one such kernel design. However, non-stationary kernels come with a great risk of model misspecification. If a new model should be trusted out of the box, a stationary kernel might be the preferred choice. 

\vspace{2mm}
\noindent
We hope that this paper motivates more practitioners to deploy and experiment with non-stationary kernels but to also be aware of some of the risks. 

\paragraph{Acknowledgements}
The work was supported by the Laboratory Directed Research and Development Program of Lawrence Berkeley National Laboratory under U.S. Department of Energy Contract No. DE-AC02-05CH11231. 
This work was further supported by the Regional and Global Model Analysis Program of the Office of Biological and Environmental Research in the Department of Energy Office of Science under contract number DE-AC02-05CH11231. This document was prepared as an account of work sponsored by the United States Government. While this document is believed to contain the correct information, neither the United States Government nor any agency thereof, nor the Regents of the University of California, nor any of their employees, makes any warranty, express or implied, or assumes any legal responsibility for the accuracy, completeness, or usefulness of any information, apparatus, product, or process disclosed, or represents that its use would not infringe privately owned rights. Reference herein to any specific commercial product, process, or service by its trade name, trademark, manufacturer, or otherwise, does not necessarily constitute or imply its endorsement, recommendation, or favoring by the United States Government or any agency thereof, or the Regents of the University of California. The views and opinions of the authors expressed herein do not necessarily state or reflect those of the United States Government or any agency thereof or the Regents of the University of California.

\vspace{2mm}
\noindent
We want to thank Kevin G. Yager from Brookhaven National Laboratory for providing the X-ray scattering dataset. This data collection used resources of the Center for Functional Nanomaterials (CFN), and the National Synchrotron Light Source II (NSLS-II), which are U.S. Department of Energy Office of Science User Facilities, at Brookhaven National Laboratory, funded under Contract No. DE-SC0012704.

\paragraph{Conflict of Interest}
The authors declare no conflict of interest.

\paragraph{Data Availability}
All data and the Jupyter notebook that runs all experiments can be found on \url{gpcam.lbl.gov/examples/non_stat_kernels}.

\paragraph{Code Availibility}
All experiments (except deep GPs) were run using the open-source Python package fvGP (\url{github.com/lbl-camera/fvGP}) which is available stand-alone and within gpCAM(\url{gpcam.lbl.gov}). The run scripts are available at \url{gpcam.lbl.gov/examples/non_stat_kernels} (available upon publication).

\paragraph{Author Contributions}
M.M.N. originally decided to write this paper, led the project, developed the test scripts and software with help from H.L. and M.R., and ran the computational experiments. H.L. suggested the non-stationarity measure and improved its practicality with help from M.M.N. and M.R. H.L. also did the majority of the work regarding deep GPs, both regarding the algorithms and the manuscript. M.R. oversaw all developments, especially regarding parametric non-stationary kernels, and further assisted with writing and editing the manuscript. All authors iteratively refined the core ideas, algorithms, and methods. All decisions regarding the work were made in agreement with all authors. All authors iteratively revised the manuscript. 
\newpage
\bibliographystyle{unsrtnat}
\bibliography{DGP_refs}

\newpage
\setcounter{section}{0}
\renewcommand{\thesection}{\Alph{section}}
\renewcommand{\theequation}{\thesection.\arabic{equation}}
\setcounter{equation}{0}
\section*{APPENDIX}
\section{Theoretical Motivation for Our Non-Stationarity Measure}\label{sec:AppNonStat}
The simple technique used in this paper is
to measure non-stationarity in the covariance structure of a dataset
based on the idea of model misspecification. Unlike traditional non-stationary
measures (e.g., histograms, periodograms), our non-stationary measure
is model-specific and can generate visual checks in wider scenarios. We followed \citet{white1982maximum} and \citet{schervish2012theory} in the following arguments.

Suppose $q(\mathbf{y}\mid\mathbf{X})$ of size $n$ has pairs of input
and responses generated independently and identically from a data-generating
mechanism $q$ (i.e., a possibly non-stationary GP or even completely
non-Gaussian models), and we can consider a parametric stationary
GP model $p(\mathbf{y}\mid\mathbf{X},(\sigma_{s},l))$ which may or
may not contain the ``true'' data generating mechanism $q$. The
MLE estimate of the stationary kernel parameters $(\sigma_{s},l)$ maximizes $\frac{1}{n}\sum_{i=1}^{n}\log p(\mathbf{y}_{(i)}\mid\mathbf{X}_{(i)},(\sigma_{s},l))$.
By the law of large numbers for the independent identically distributed
$\log p(\mathbf{y}_{(i)}\mid\mathbf{X}_{(i)},(\sigma_{s},l))$ summands,
this quantity converges in joint probability to the expectation of
$\mathbb{E}_{q}\log p(\mathbf{y}\mid\mathbf{X},(\sigma_{s},l))$ with
respect to the data-generating mechanism $q$.
When $q(\mathbf{y}\mid\mathbf{X})=p(\mathbf{y}\mid\mathbf{x},(\sigma_{s,0},l_{0}))$
for some kernel parameters $(\sigma_{s,0},l_{0})$, that is, the input-response
data is actually generated by the GP with parametric stationary kernel
with true parameter $(\sigma_{s,0},l_{0})$. The following argument
shows this equivalence between MLE and the KL-divergence minimizer.
\begin{align}
(\sigma_{s},l)^{*} & =\mathop{{\rm arg\,min}}\limits _{(\sigma_{s},l)}D_{KL}\left(\left.p(\mathbf{y}|\mathbf{X},(\sigma_{s,0},l_{0}))\right\Vert p(\mathbf{y}|\mathbf{X},(\sigma_{s},l))\right)\\
 & =\mathop{{\rm arg\,min}}\limits _{(\sigma_{s},l)}\mathbb{E}_{p(\mathbf{y}|\mathbf{x},(\sigma_{s,0},l_{0}))}\left[\log p(\mathbf{y}|\mathbf{X},(\sigma_{s,0},l_{0}))-\log p(\mathbf{y}|\mathbf{X},(\sigma_{s},l))\right]\\
 & =\mathop{{\rm arg\,max}}\limits _{(\sigma_{s},l)}\mathbb{E}_{p(\mathbf{y}|\mathbf{x},(\sigma_{s,0},l_{0}))}\log p(\mathbf{y}|\mathbf{X},(\sigma_{s},l))
\end{align}
and then $\mathbb{E}_{q}\log p(\mathbf{y}\mid\mathbf{X},(\sigma_{s},l))$
is maximized at $(\sigma_{s,0},l_{0})$. This explains the consistency
of the MLE and we will expect that the MLE $(\sigma_{s},l)^{*}$ of
kernel parameters $(\sigma_{s},l)$ will concentrate around the true
parameter value $(\sigma_{s,0},l_{0})$. With regard to our procedure,
if all the data are generated by a GP $p(\mathbf{y}|\mathbf{X},(\sigma_{s,0},l_{0}))$,
then each subsample needs to produce an MLE of $(\sigma_{s},l)$ that
is close to $(\sigma_{s,0},l_{0})$. Therefore, we will consider a
\emph{stably concentrated} $(\sigma_{s},l)$-plot as a strong evidence
that the data is from a stationary GP. 
When $q(\mathbf{y}\mid\mathbf{X})$ does not necessarily belong to
the GP with parametric stationary kernels, we may write similarly
\begin{align*}
\mathbb{E}_{q}\log p(\mathbf{y}|\mathbf{X},(\sigma_{s},l)) & =\mathbb{E}_{q}\log q(\mathbf{y}|\mathbf{X})-\ensuremath{\mathbb{E}}_{q}\log\frac{q(\mathbf{y}|\mathbf{X})}{p(\mathbf{y}|\mathbf{X},(\sigma_{s},l))}\\
 & =\mathbb{E}_{q}\log q(\mathbf{y}|\mathbf{X})-D_{KL}\left(\left.q(\mathbf{y}|\mathbf{X})\right\Vert p(\mathbf{y}|\mathbf{X},(\sigma_{s},l))\right)
\end{align*}
Since the term $\mathbb{E}_{q}\log q(\mathbf{y}|\mathbf{X})$ does
not involve the kernel parameter $(\sigma_{s},l)$, the value of $(\sigma_{s},l)^{*}$
obtained by maximizing $\mathbb{E}_{q}\log p(\mathbf{y}|\mathbf{X},(\sigma_{s},l))$
under a misspecified data-generating mechanism $q$ is the minimizer
of $D_{KL}\left(\left.q(\mathbf{y}|\mathbf{X})\right\Vert p(\mathbf{y}|\mathbf{X},(\sigma_{s},l))\right)$,
often referred to as pseudo-true parameters
which is observed to be sensitive to the selection of
the subsample. In other words, the minimizer of $D_{KL}\left(\left.q(\mathbf{y}|\mathbf{X})\right\Vert p(\mathbf{y}|\mathbf{X},(\sigma_{s},l))\right)$
exhibits higher sensitivity than $D_{KL}\left(\left.p(\mathbf{y}|\mathbf{X},(\sigma_{s,0},l_{0}))\right\Vert p(\mathbf{y}|\mathbf{X},(\sigma_{s},l))\right)$.
If we assume the joint distribution of $p(\mathbf{y}|\mathbf{X},(\sigma_{s,0},l_{0}))$
has mean $\bm{\mu}_{(\sigma_{s,0},l_{0})}$ and covariance $\bm{\Sigma}_{(\sigma_{s,0},l_{0})}$;
the joint distribution of $p(\mathbf{y}|\mathbf{X},(\sigma_{s},l))$
has mean $\bm{\mu}_{(\sigma_{s},l)}$ and covariance $\bm{\Sigma}_{(\sigma_{s},l)}$.
\begin{align*}
D_{KL}\left(\left.p(\mathbf{y}|\mathbf{X},\bm{\Sigma}_{(\sigma_{s,0},l_{0})})\right\Vert p(\mathbf{y}|\mathbf{X},\bm{\Sigma}_{(\sigma_{s},l)})\right) & \\ =\frac{1}{2}\left[\log\frac{|\bm{\Sigma}_{(\sigma_{s},l)}|}{|\bm{\Sigma}_{(\sigma_{s,0},l_{0})}|}-k+\left(\bm{\mu}_{(\sigma_{s,0},l_{0})}-\bm{\mu}_{(\sigma_{s},l)}\right)^{T}\bm{\Sigma}_{(\sigma_{s},l)}^{-1}\left(\bm{\mu}_{(\sigma_{s,0},l_{0})}-\bm{\mu}_{(\sigma_{s},l)}\right)+tr\left\{ \bm{\Sigma}_{(\sigma_{s},l)}^{-1}\bm{\Sigma}_{(\sigma_{s,0},l_{0})}\right\} \right],\\
\end{align*}
We can discover that subsampling when both $p$ and $q$ are from a GP, will only affect this quantity at
$O(1/\sqrt{n})$ by convergence rates of sample means and variances, while if $q$ is not a stationary GP model the magnitude
could be large. We have to point out that this observation --- that the KL-divergence becomes sensitive to subsampling when the data-generating mechanism and the model do not belong to the same family --- is empirical. However, it motivates our algorithm and works well in practice.  
Although the MLE kernel parameters $(\sigma_{s},l)^{*}$ under misspecification
will converge to the pseudo-true parameters rather than the true parameters
\cite{grunwald2017inconsistency,kleijn2006misspecification}, these
pseudo-true parameters are sensitive to subsampling. This creates
a much more scattered $(\sigma_{s},l)$-plot without specific concentration
pattern. By looking into the behavior of MLE parameters of stationary kernels
we can quantify how far it is from the ``stationarity'', which is
the rationale of our technique. As aforementioned, a stationary kernel
will assume that the pairwise covariance does not depend on the location
in space, but only on the distance of the input points. This leads
us (and others) to deduce that non-stationarity will manifest itself
by varying hyperparameters of a stationary kernel --- one isotropic
length scale and a constant signal variance --- when they are trained
on localized subsets of the data. In our visualization $(\sigma_{s},l)$-plot,
we can observe whether the length scale $l$ or variance parameter
$\sigma_{s}$ is the source of non-stationarity.

\section{gpflux DGP Script  }\label{sec:appA}
This script is written in $\mathtt{Python}$ with the support of cited packages.
{\scriptsize{}
\begin{lstlisting}[language=Python]
import numpy as np
import gpflow, gpflux
import tensorflow as tf
X = x_data1.reshape(-1,1)
Y = y_data1.reshape(-1,1)
# Layer 1
Z = np.linspace(X.min(), X.max(), X.shape[0] // 1).reshape(-1, 1)
kernel1 = gpflow.kernels.RBF()
inducing_variable = gpflow.inducing_variables.InducingPoints(Z.copy())
gp_layer1 = gpflux.layers.GPLayer(
    kernel1, 
    inducing_variable1, 
    num_data=X.shape[0], 
    num_latent_gps=X.shape[1]
)

# Layer 2
kernel2 = gpflow.kernels.RBF()
inducing_variable2 = gpflow.inducing_variables.InducingPoints(Z.copy())
gp_layer2 = gpflux.layers.GPLayer(
   kernel2,
   inducing_variable2,
   num_data=X.shape[0],
   num_latent_gps=X.shape[1],
   mean_function=gpflow.mean_functions.Zero(),
)

# Initialise likelihood and build model
likelihood_layer = gpflux.layers.LikelihoodLayer\
(gpflow.likelihoods.Gaussian(1e-4))
two_layer_dgp = gpflux.models.DeepGP([gp_layer1, gp_layer2], likelihood_layer)

# Compile and fit
model = two_layer_dgp.as_training_model()
model.compile(tf.optimizers.Adam(0.001))
history = model.fit({"inputs": X, "targets": Y}, epochs=int(1e4), verbose=0)


modelp = two_layer_dgp.as_prediction_model()
out = modelp(x_pred1D.reshape(-1,1))

mu = out.f_mean.numpy().squeeze()
var = out.f_var.numpy().squeeze()
lower = mu - 3 * np.sqrt(var)
upper = mu + 3 * np.sqrt(var)
meanDGP = mu
varDGP = var


plt.figure(figsize = (16,10))
plt.plot(x_pred1D,meanDGP, label = "posterior mean", linewidth = 4)
plt.plot(x_pred1D,f_test1(x_pred1D), label = "latent function", linewidth = 4)
plt.fill_between(x_pred1D, meanDGP - 3. * np.sqrt(varDGP), meanDGP + 3. * \
np.sqrt(varDGP), alpha = 0.5, color = "grey", label = "var")
plt.scatter(x_data1,y_data1, color = 'black')
plt.xticks([])
plt.yticks([])
#plt.legend()
# #rmse = torch.mean(torch.pow(predictive_means.mean(0) - test_y, 2)).sqrt()
# #print(f"RMSE: {rmse.item()}, NLL: {-test_lls.mean().item()}")
plt.ylim([-0.2,1.2])
print("error: ", mse(f_test1(x_pred1D),meanDGP))
print("average crps: ", crps(f_test1(x_pred1D),meanDGP,np.sqrt(varDGP)))
\end{lstlisting}
}

\section{1d Bayesian Deep GP Script }\label{sec:appB}
This script is written in $\mathtt{R}$. The same script was used in the three-dimensional test with changes being made to the data reading commands and the number of MCMC iterations.
{\scriptsize{}
\begin{lstlisting}[language=R]
library(deepgp)
library(rbenchmark)

N_MCMC <- 1000
N_replicate <- 1 #Number of replicates when doing benchmark
# Training data
data <- read.csv("data.csv")
# Extract x and y from the data
x <- as.matrix(data[,1])
y <- as.vector(data[,2])


# Testing data
np <- 1000
xp <- seq(0, 1, length = np)
yp <- ff(xp)

plot(xp, yp, type = "l", col = 4, xlab = "X",\
ylab = "Y", main = "ground truth function")
points(x, y)

#benchmark_result <- benchmark(fit <- \
fit_two_layer(x, y, nmcmc = N_MCMC, true_g = 1e-2), replications = N_replicate)
benchmark_result <- benchmark(fit <- \
fit_two_layer(x, y, nmcmc = N_MCMC), replications = N_replicate)
write.table(benchmark_result, file = "benchmark_results_1d.txt")
plot(fit)
fit <- trim(fit, 500, 2)
fit <- predict(fit, xp, cores = 1)
plot(fit)
write.csv(fit$mean, file = "mean_2layer_full_fixed.csv", row.names = FALSE)
write.csv(fit$s2, file = "var_2layer_full_fixed.csv", row.names = FALSE)
plot(fit)
\end{lstlisting}
}

\section{The Deep GP Algorithm}\label{app:deepGP}
DGPs provide a fusion of the non-parametric flexibility of GPs with the hierarchical structure typical of deep neural networks. In essence, DGPs stack multiple layers of Gaussian processes on top of each other, with the output of one layer serving as the input for the next, much like the architecture of deep neural networks. 
Fitting a DGPs is a nuanced procedure that marries the principles of GP with the hierarchical structure of deep networks. Through techniques like variational inference, the challenges posed by the depth and complexity of the model are addressed, paving the way for powerful, flexible, and interpretable modeling of data.
When fitting a DGP, the primary objective is to optimize the marginal likelihood of the observed data, given the parameters of the Gaussian processes in each layer and the likelihood function. The marginal likelihood, in the context of GPs, is the likelihood of the observations after integrating out the latent functions. Mathematically, this is expressed as
\[
p(Y | X) = \int p(Y | F) p(F | X) \, dF
\]
Here, \( Y \) are the observations, \( X \) are the inputs, \( F \) represents the latent functions (outputs of the GP layers), \( p(Y | F) \) is the likelihood, and \( p(F | X) \) is the Gaussian process prior. However, in DGPs, the optimization of this marginal likelihood becomes more intricate due to the cascading nature of layers. Each layer's latent functions become the inputs for the subsequent layer, resulting in a composite model where the inference is not straightforward. The depth introduces additional non-linearity and complexity. From a frequentist viewpoint, the intricacies of DGP are typically managed through variational inference. This method centers on approximating the authentic posterior using a variational distribution, characterized by specific variational parameters. The primary goal becomes the maximization of the evidence lower bound (ELBO) in Algorithm \ref{alg:DGPFit}, effectively providing a lower threshold on the log marginal likelihood. This transformation changes a previously intractable inference conundrum into a solvable optimization challenge, wherein the variational parameters are systematically tuned to align the variational distribution more closely with the authentic posterior. On the other hand, from a Bayesian perspective, \cite{sauer2023non} offers a profound exploration into the nuances of DGPs. This comparative analysis on an array of DGP models sheds light on their distinctive performance metrics and characteristics. A notable highlight from the paper is the unveiling of an innovative Bayesian approach to classification problems using DGPs, emphasizing their prowess in adeptly managing multifaceted structured data and in discerning intricate patterns and associations.

\begin{algorithm}
\caption{Detailed Deep Gaussian Process Model Training}
\begin{algorithmic}[1]
\Require $x_{data}$, $y_{data}$, $x_{pred}$
%\Ensure Predicted mean and variance for $x\_pred1D$

%\State Reshape data $X$ and $Y$ into single columns
\State Create inducing points $Z$ spanning the range of $X$
\State Define two GP layers:
    \begin{itemize}
        \item Each layer is defined with an RBF kernel
        \item Using a random collection of inputs $Z$ as inducing points
        \item The last layer has a zero mean function
    \end{itemize}
\State Define a Gaussian likelihood per layer with noise variance $1e-4$
\State Construct a 2-layer Deep GP model with the GP layers and likelihood
\State Compile the model with the stochastic Adam optimizer \cite{dutordoir2021gpflux,salimbeni2017doubly} (Alternatively, one can perform Bayesian variational inference \cite{sauer2022vecchia}.)
\State Train the model for $10^4$ epochs:
    \begin{itemize}
        \item Optimize the parameters to maximize the marginal likelihood $p(Y | X)$
        \item Where $p(Y | X) = \int p(Y | F) p(F | X) \, dF$
    \end{itemize}
\State Predict on $x_{pred}$ and extract mean and variance
%\State Plot results and print metrics

\end{algorithmic}\label{alg:DGPFit}
\end{algorithm}

\section{Available deep GP implementations}
Table \ref{tab:impl} offers a comprehensive overview of various implementations of Deep Gaussian Processes (DGPs), a powerful tool in machine learning. It's fascinating to see the diversity of implementations across different programming languages and frameworks, such as R, TensorFlow, and PyTorch. This wide variety reflects the growing interest and applicability of DGPs in various domains, from scientific research to industrial applications.

One of the key aspects of the table is the different fitting methods used, such as MCMC (Markov Chain Monte Carlo), SVI (Stochastic Variational Inference), and MLE (Maximum Likelihood Estimation). The choice of fitting method can significantly impact the efficiency and accuracy of the model. It's noteworthy that some packages may support more fitting methods than those listed, indicating flexibility and adaptability in the approach to modeling with DGPs.

Recent developments in the field are evident from the inclusion of implementations associated with recent publications. This ongoing research and development signal a vibrant and evolving field, where new methodologies and techniques are continually being explored and refined.

Accessibility is another important aspect highlighted in the table. Many of the implementations are openly accessible through platforms like GitHub. This open-source approach fosters collaboration and innovation, allowing researchers and practitioners to not only utilize these resources but also contribute to their development.

The table's caption mentions that it extends and compares information from other sources, suggesting a comprehensive effort to collate and compare various DGP implementations. This comparative approach adds value to the table by placing the different implementations in context with one another, providing a broader perspective on the landscape of DGPs.

While the table provides a valuable summary, it also hints at potential challenges. Users may need to delve into the individual papers or repositories to understand the specific features, advantages, and limitations of each implementation. This deeper exploration may reveal nuances and details that are beyond the scope of the table but are essential for a complete understanding of each DGP implementation.

\begin{table}[ht!]
\begin{adjustbox}{center}

{\scriptsize{}}%
\begin{tabular}{|c|c|c|c|}
\hline 
{\scriptsize{}Repository/Package} & {\scriptsize{}Language} & {\scriptsize{}Reference}\tablefootnote{{\scriptsize{}We sort the table by the publication date of the associated
paper.}} & {\scriptsize{}Fitting}\tablefootnote{{\scriptsize{}We follow the main fitting method featured by the package,
some packages may support more fitting methods than listed here.}}\tabularnewline
\hline 
\hline 
{\scriptsize{}\href{https://cran.r-project.org/web/packages/deepgp/index.html}{deepGP}} & {\scriptsize{}R} & {\scriptsize{}\citet{sauer2023non}} & {\scriptsize{}MCMC (elliptic sampler)}\tabularnewline
\hline 
{\scriptsize{}SDSP} & {\scriptsize{}R/tensorflow} & {\scriptsize{}\citet{zammit2022deep}} & {\scriptsize{}SVI}\tabularnewline
\hline 
{\scriptsize{}SDSP-MCMC} & {\scriptsize{}R/stan} & {\scriptsize{}\citet{zammit2022deep}} & {\scriptsize{}MCMC}\tabularnewline
\hline 
{\scriptsize{}SIWGP} & {\scriptsize{}R/tensorflow} & {\scriptsize{}\citet{zammit2022deep}} & {\scriptsize{}MLE}\tabularnewline
\hline 
{\scriptsize{}\href{https://github.com/secondmind-labs/GPflux}{GPFlux}} & {\scriptsize{}tensorflow} & {\scriptsize{}\citet{dutordoir2021gpflux}} & {\scriptsize{}MLE/SVI (variational)}\tabularnewline
\hline 
{\scriptsize{}\href{https://github.com/kekeblom/DeepCGP}{DeepCGP}} & {\scriptsize{}tensorflow} & {\scriptsize{}\citet{blomqvist2020deep}} & {\scriptsize{}SVI (doubly stochastic)}\tabularnewline
\hline 
{\scriptsize{}\href{https://github.com/hughsalimbeni/DGPs_with_IWVI}{DGPs\_with\_IWVI}} & {\scriptsize{}tensorflow} & {\scriptsize{}\citet{salimbeni2019deep}} & {\scriptsize{}SVI}\tabularnewline
\hline 
{\scriptsize{}\href{https://github.com/cambridge-mlg/sghmc_dgp}{sghmc\_dgp}} & {\scriptsize{}tensorflow} & {\scriptsize{}\citet{havasi2018inference}} & {\scriptsize{}MCMC (HMC)}\tabularnewline
\hline 
{\scriptsize{}\href{https://github.com/FelixOpolka/Deep-Gaussian-Process}{Deep-Gaussian-Process}} & {\scriptsize{}tensorflow2} & {\scriptsize{}\citet{salimbeni2017doubly}} & {\scriptsize{}SVI}\tablefootnote{{\scriptsize{}stochastic variational inference}}\tabularnewline
\hline 
{\scriptsize{}\href{https://github.com/cornellius-gp/gpytorch/tree/master/examples/05_Deep_Gaussian_Processes}{GPyTorch/DeepGP}} & {\scriptsize{}pytorch} & {\scriptsize{}\citet{salimbeni2017doubly,gardner2018gpytorch}} & {\scriptsize{}SVI (doubly stochastic)}\tabularnewline
\hline 
{\scriptsize{}\href{https://github.com/thangbui/deepGP_approxEP/tree/master}{DGPRFF}} & {\scriptsize{}pytorch} & {\scriptsize{}\citet{cutajar2017random}} & {\scriptsize{}SVI}\tabularnewline
\hline 
{\scriptsize{}\href{https://github.com/UCL-SML/Doubly-Stochastic-DGP}{Doubly-Stochastic-DGP}} & {\scriptsize{}pytorch} & {\scriptsize{}\citet{salimbeni2017doubly}} & {\scriptsize{}SVI (doubly stochastic)}\tabularnewline
\hline 
{\scriptsize{}\href{https://github.com/SheffieldML/PyDeepGP}{PyDeepGP}} & {\scriptsize{}python} & {\scriptsize{}\citet{dai2015variational,damianou2013deep}} & {\scriptsize{}MLE/SVI (variational)}\tabularnewline
\hline 
{\scriptsize{}DGPsparse} & {\scriptsize{}tensorflow} & {\scriptsize{}\citet{damianou2013deep}} & {\scriptsize{}MLE}\tabularnewline
\hline 
{\scriptsize{}DGPfull} & {\scriptsize{}python} & {\scriptsize{}\citet{schmidt2003bayesian}} & {\scriptsize{}MCMC (elliptic sampler)}\tabularnewline
\hline 
\end{tabular}{\scriptsize\par}

\end{adjustbox}{\footnotesize{}\caption{\label{tab:impl}Comparison of different DGP implementations. This table extends
Table 1 in \citet{zammit2022deep}; and tables in \citet{dutordoir2021gpflux}.}
}{\footnotesize\par}
\end{table}

\end{document}